%% file: main_arxiv.tex
\documentclass[sigconf,screen]{acmart}

\setcopyright{none}
\copyrightyear{2024}
\acmYear{2024}

\acmConference[]{ACM}{Survey}{Instructional Editing}

\usepackage{booktabs} %
\usepackage{mathtools}
\makeatletter \newif\if@restonecol \makeatother  
\usepackage[linesnumbered, ruled, vlined]{algorithm2e}
\usepackage{algpseudocode}

\usepackage[bottom]{footmisc}

\usepackage{epstopdf}
\usepackage{graphics}
\DeclareGraphicsExtensions{.eps,.gz,.eps,.pdf,.png,.jpg}

\graphicspath{{./}{./figures/}{../}}

\usepackage{xurl}
\usepackage{hyperref}

\newcommand\rurl[1]{%
  \href{https://#1}{\nolinkurl{#1}}%
}

\usepackage{makecell}
\usepackage{listings}
\usepackage{graphicx}
\usepackage{amsmath}
\usepackage{amsfonts}
\usepackage{amsthm}
\usepackage{dsfont}
\usepackage{wrapfig}
\usepackage{enumerate}
\usepackage{paralist}
\usepackage{multirow}
\usepackage{verbatim}
\usepackage{lipsum}
\usepackage{mathrsfs}
\usepackage{threeparttable}
\usepackage{tabularx}
\usepackage{subcaption}
\usepackage{adjustbox}

\theoremstyle{definition}

\theoremstyle{plain}

\newcommand{\sstitle}[1]{\smallskip\noindent\textbf{#1.\/}}

\def\Snospace~{\S{}}

\makeatletter
\newcommand{\removelatexerror}{\let\@latex@error\@gobble}
\makeatother

\hypersetup{
  pdftoolbar=true,        %
  pdfmenubar=true,        %
  pdffitwindow=false,     %
  pdfstartview={FitH},    %
  pdftitle={Complex Event Processing under Constrained Resources},    %
  pdfauthor={},     %
  pdfsubject={},   %
  pdfcreator={},   %
  pdfproducer={}, %
  pdfkeywords={}, %
  pdfnewwindow=true,      %
  colorlinks=true,       %
  linkcolor=Brown,          %
  citecolor=OliveGreen,        %
  filecolor=magenta,      %
  urlcolor=NavyBlue           %
}

 \newcommand\Mark[1]{\textsuperscript#1}

\begin{document}

\title[A Survey of Instructional Image Editing Controls]{Instruction-Guided Editing Controls for Images and Multimedia: A Survey in LLM era
}

\author{
Thanh Tam Nguyen\Mark{1}, %
Zhao Ren\Mark{2},
Trinh Pham\Mark{3},
Thanh Trung Huynh\Mark{4},
Phi Le Nguyen\Mark{5},\\
Hongzhi Yin\Mark{6}, %
Quoc Viet Hung Nguyen\Mark{1}%
}

\affiliation{%
  \institution{
  \Mark{1}Griffith University,
  \Mark{2}University of Bremen,
  \Mark{3}Ho Chi Minh City University of Technology,
  \Mark{4}\'{E}cole Polytechnique F\'{e}d\'{e}rale de Lausanne,
  \Mark{5}Hanoi University of Science and Technology,
  \Mark{6}The University of Queensland
  }
  \country{}
}

\renewcommand{\shortauthors}{Thanh Tam Nguyen, et al.}

\begin{abstract}

The rapid advancement of large language models (LLMs) and multimodal learning has transformed digital content creation and manipulation. Traditional visual editing tools require significant expertise, limiting accessibility. Recent strides in instruction-based editing have enabled intuitive interaction with visual content, using natural language as a bridge between user intent and complex editing operations. This survey provides an overview of these techniques, focusing on how LLMs and multimodal models empower users to achieve precise visual modifications without deep technical knowledge. By synthesizing over 100 publications, we explore methods from generative adversarial networks to diffusion models, examining multimodal integration for fine-grained content control. We discuss practical applications across domains such as fashion, 3D scene manipulation, and video synthesis, highlighting increased accessibility and alignment with human intuition. Our survey compares existing literature, emphasizing LLM-empowered editing, and identifies key challenges to stimulate further research. We aim to democratize powerful visual editing across various industries, from entertainment to education.
Interested readers are encouraged to access our repository at \url{https://github.com/tamlhp/awesome-instruction-editing}.

\end{abstract} 


\begin{teaserfigure}	
    \centering
    \includegraphics[width=0.77\linewidth]{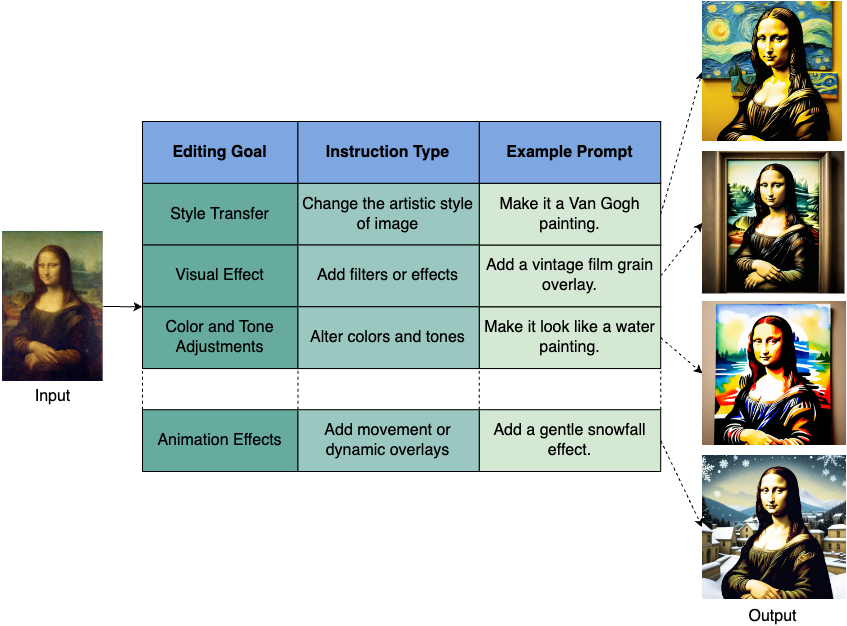}
    \vspace{-.5em}
    \caption{Instruction-guided image editing.}
    \label{fig:overview-image-editing}
 \end{teaserfigure}

\maketitle

\section{Introduction}
\label{sec:intro}

Visual design tools have become essential in various multimedia fields, although they often require prior knowledge to use effectively. Recent research has emphasised text-guided image editing as a way to make these tools more accessible and controllable~\cite{li2020manigan,patashnik2021styleclip,gal2022stylegan,crowson2022vqgan}, as in \autoref{fig:overview-image-editing}. Studies have shown the effectiveness of diffusion models in creating realistic images and their application in image editing through techniques like swapping latent cross-modal maps for visual manipulation~\cite{ho2020denoising,kim2022diffusionclip}. Additionally, specific region editing is made possible through guided masks~\cite{nichol2022glide,avrahami2022blended}. Moving away from complex descriptions and masks, instruction-based editing has gained traction for its straightforward approach, allowing users to directly command how and what aspects of an image to edit~\cite{hertz2023prompt,mokady2023null,kawar2023imagic}. This paradigm is noted for its practicality, aligning closely with human intuition~\cite{fu2024guiding,el2019tell,fu2020sscr}.

The latest text-to-image generative models offer impressive image quality and accuracy in reflecting the given captions, marking a significant leap in content generation technologies~\cite{alayrac2022flamingo,ramesh2022hierarchical,rombach2022high}. Among these advancements, instructional image editing has emerged as a particularly promising application~\cite{brooks2023instructpix2pix}. This method streamlines the editing process by eliminating the need for detailed before-and-after captions~\cite{avrahami2022blended,wallace2023edict}. Instead, users can provide simple, human-readable instructions, such as ``change the dog to a cat'', making the editing process more intuitive and aligned with how humans naturally approach image modification~\cite{zhang2024hive}.

In recent years, advancements in large language models (LLMs)~\cite{touvron2023llama,brown2020language} have dramatically reshaped the landscape of image and video manipulation. The convergence of these technologies has enabled more intuitive, flexible, and high-fidelity editing processes, largely driven by natural language instructions~\cite{wu2023visual,feng2024layoutgpt,chakrabarty2023spy}. These innovations span various applications, from fashion image editing and 3D scene manipulation to video-to-video synthesis and audio-driven editing, empowering users to achieve fine-grained control over visual content. Moreover, Multimodal large language models (MLLMs), building upon the foundational capabilities of traditional LLMs, have extended the boundaries of vision-language tasks~\cite{zhang2024llamaadapter}. By integrating latent visual knowledge and treating images as input, MLLMs enhance performance in tasks requiring both textual and visual reasoning. The emergence of diffusion models, such as LLaVA \cite{liu2024visual} and MiniGPT-4~\cite{zhu2024minigpt}, has further elevated the potential of these frameworks by improving image-text alignment through instruction tuning. These models, including GILL~\cite{koh2024generating} and SEED~\cite{ge2023making}, facilitate coherent image generation from textual input while preserving rich visual semantics, marking a pivotal evolution in instruction-based editing.

This review paper explores the evolution and diversity of techniques underpinning instruction-based image and video editing, synthesizing cutting-edge approaches that integrate human feedback, multimodal signals, and advanced neural architectures. The focus spans from early models leveraging generative adversarial networks (GANs)~\cite{patashnik2021styleclip} to the latest innovations using diffusion models, including frameworks like Pix2Pix~\cite{brooks2023instructpix2pix}, InstructBrush~\cite{zhao2024instructbrush}, and FlexEdit~\cite{nguyen2024flexedit}. Additionally, specialized models for audio- and video-driven editing, such as Noise2Music~\cite{huang2023noise2music} and Fairy~\cite{wu2023fairy}, are examined, demonstrating the versatility and creativity unlocked by these methods.

By analyzing over 100 recent key publications, this review delves into key technological breakthroughs, evaluates their effectiveness, and considers potential avenues for further innovation. From 3D image editing~\cite{sabat2024nerf} to fashion editing~\cite{wang2024texfit}, this paper highlights how these models are reshaping industries ranging from entertainment and fashion to education and remote sensing~\cite{han2024exploring}. Through this comprehensive overview, we aim to identify emerging trends, challenges, and opportunities in the growing field of text-driven, instruction-guided image and video editing.

\sstitle{Differences with Existing Surveys}
Our survey differs from existing surveys in its specific focus on instruction-based image and video editing empowered by LLMs. While Li et al.~\cite{li2024survey} focus on the integration of various modalities for retrieval tasks, our paper highlights the use of instructions for precise visual editing. Qin et al.~\cite{qin2024infobench} evaluate instruction-following abilities in LLMs but does not address their application in visual manipulation, which is a key focus of our review. Similarly, Yin et al.~\cite{yin2023llm} address instruction-following in language models with a broader emphasis on ethical concerns, whereas our review emphasizes the technical advancements in using these capabilities for visual content generation and editing across various domains, including image, video, and 3D manipulation. Closest to our review is~\cite{zhan2023multimodal}, which explores generative AI techniques but lacks the detailed exploration of instruction-following in visual editing contexts, as seen in our paper. Especially, we consider caption-based image editing~\cite{chen2018language,couairon2022diffedit,lin2023text} is a part of instruction-based image editing but we do not fully focus on the former. Rather, we are interested in user-friendly instructions that have practical implications for broad audience when editing images. \autoref{tab:surveys} summarises the difference between our surveys and existing ones.

\begin{table*}[t]
\centering
\footnotesize
\caption{A comparison between existing surveys}
\label{tab:surveys}
\begin{tabular}{c|c|c|c}
\toprule
\textbf{Survey} & \textbf{Focused Task} & \textbf{Focused Modality} & \textbf{Key Contents}\\
\midrule
\cite{qin2024infobench} & Editing & Text & Instruction development, Evaluation concerns \\
\cite{yin2023llm} & Editing & Text & LLM-empowered instructions, Instruction tuning \\
\cite{li2024survey} & Retrieval & Image, Video, Audio & Image-text composite retrieval, Multimodal composite retrieval \\
\cite{zhan2023multimodal} & Generation & Image & Text guidance, Audio guidance, Sketch guidance, etc. \\
Ours & Editing & Image, Video, Audio & Instruction mechanisms, Augmentations, Learning stragies, Model designs, Loss functions \\
\bottomrule
\end{tabular}
\end{table*}

\sstitle{Paper Collection Methodology}
To map the research landscape on this subject, we used a range of keyword searches and combinations such as ``image editing'', ``image manipulation'', ``text-guided'', ``instruction-followed'', and ``instruction-guided''. Initially, we relied on platforms like Google Scholar, Semantic Scholar, and the AI-enhanced tool Scite.ai to compile an initial set of studies. We then expanded this collection by conducting backward searches, reviewing the references in the selected papers, and forward searches to identify works that cited them. To ensure accuracy, we manually evaluated the relevance of each study, given that some focused on related areas like image generation or retrieval but employed similar techniques. This thorough process ultimately resulted in the identification of over 100 pivotal papers relevant to the field.

\begin{figure*}
    \centering
    \includegraphics[width=0.95\linewidth]{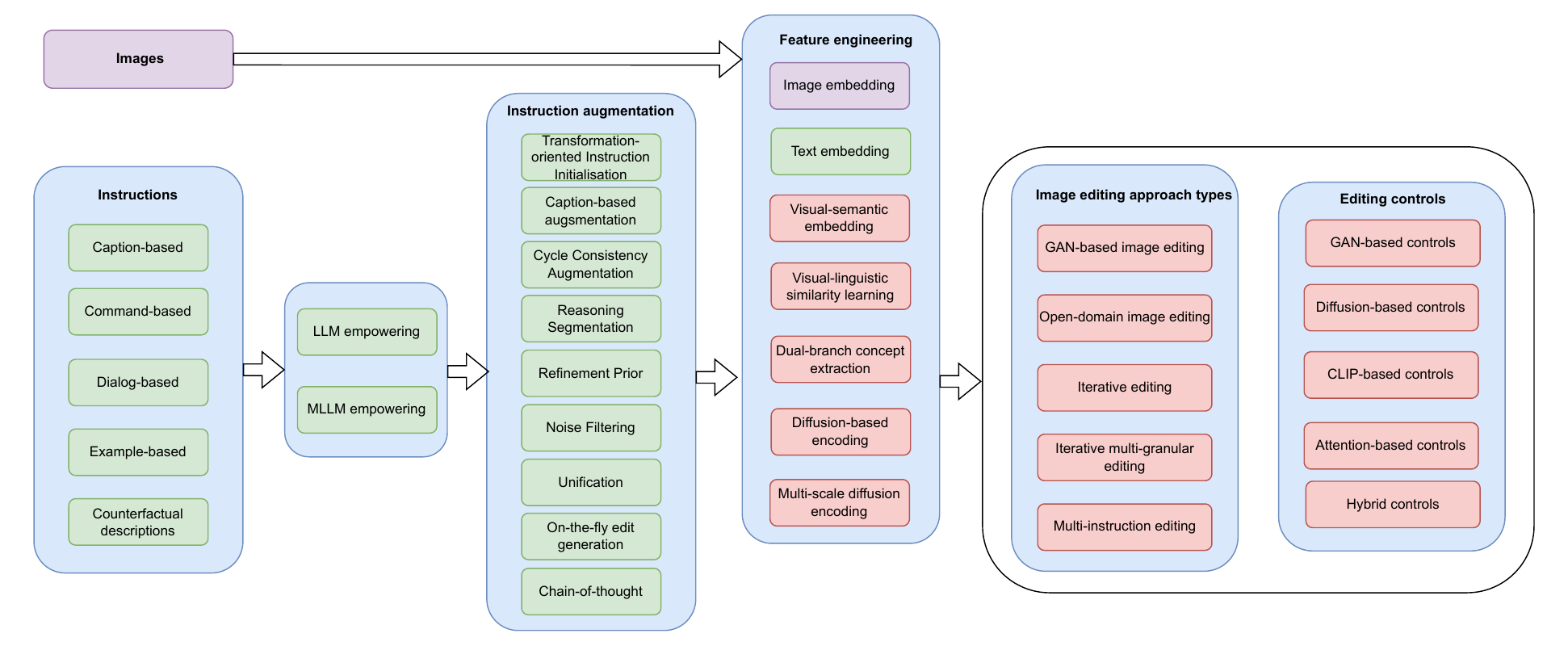}
    \caption{Process-based taxonomy of instruction-guided image editing.}
    \label{fig:taxonomy}
\end{figure*}

\sstitle{Contributions}
The main contributions of this survey are:

\begin{itemize}
    \item \emph{Comprehensive Review:} This study provides a comprehensive review of LLM-empowered image and media editing. We have gathered and summarised an extensive body of literature, including both published works and pre-prints up to October 2024.
    \item \emph{Process-based Taxonomy:} We have organised the literature according to the developmental stages of an image editing framework. \autoref{fig:taxonomy} presents the taxonomy we developed to structure the existing works in the field.
    \item \emph{Optimisation Tools:} We have curated a set of optimisation tools for developing end-to-end image editing frameworks, covering model designs, learning strategies, instruction mechanisms, data augmentations, and loss functions.
    \item \emph{Practical Applications:} We discuss various practical applications across multiple domains, including style, fashion, face editing, scene manipulation, charts, remote sensing, 3D, speech, music, and video editing.
    \item \emph{Challenges and Future Directions:} Instruction-guided visual design remains an emerging area of research. Based on the surveyed literature, we identify several unresolved challenges and propose future research directions to explore more editing use cases and user-friendly editing controls.
    \item \emph{Sources, Datasets, and Metrics:} To support empirical research, we provide a comprehensive overview of available source codes, datasets, and evaluation metrics that have been utilised in the field.
    \item \emph{Online Updating Resource:} To support ongoing research in LLM-empowered visual design, we have created an open-source repository\footnote{\url{https://github.com/tamlhp/awesome-instruction-editing}}, which consolidates relevant studies, including links to papers and available code.
\end{itemize}

\section{Models and Formulations}
\label{sec:problem}

\subsection{Image Editing}

\begin{figure*}
    \centering
    \includegraphics[width=0.95\linewidth]{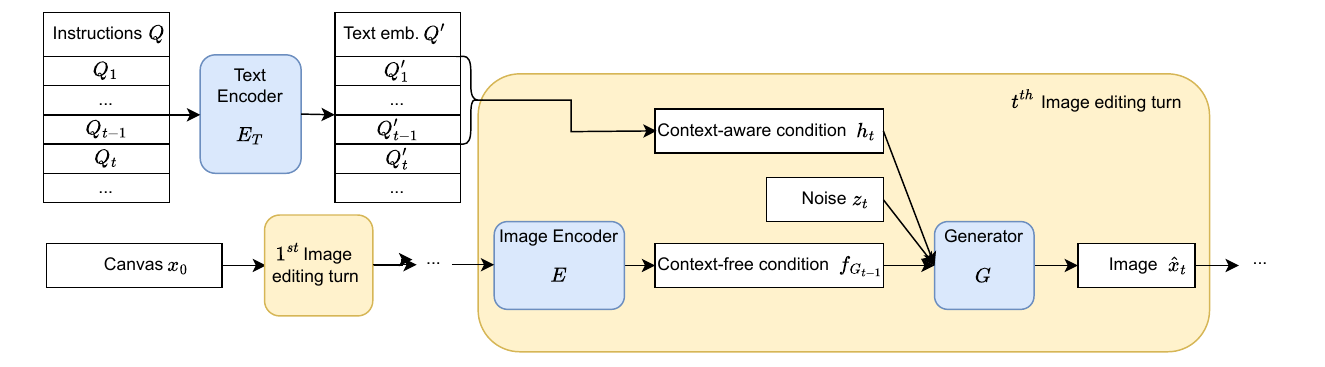}
    \caption{Basic procedure of instruction-guided image editing}
    \label{fig:gan-image-editing}
\end{figure*}

\sstitle{GAN-based Image Editing}
Image editing is an interactive process where a drawing canvas $x0$ and a sequence of instructions $Q$ are provided~\cite{el2019tell}. As shown in \autoref{fig:gan-image-editing}, each turn in the conversation results in a new image \(\tilde{x}_t\) generated by a conditioned generator G. The generator G takes as input a noise vector \(z_t\) from a standard normal distribution and is conditioned on variables \(h_t\) (a context-aware condition) and \(f_{Gt-1}\) (a context-free condition):
 \begin{equation}
 	\tilde{x}_t = G(z_t, h_t, f_{G_{t-1}})
 \end{equation}

The noise vector \(z_t\) has dimension \(N_z\). The context-free condition \(f_{Gt-1}\) is the encoding of the previous image \(\tilde{x}_{t-1}\), produced by encoder \(E_G\), which utilizes a Convolutional Neural Network (CNN) to generate low-resolution feature maps with dimensions \((K_g \times K_g \times N_g)\). The context-aware condition \(h_t\) incorporates the conversation history to facilitate better encoding of the instruction within the context of the conversation up to time \(t - 1\).

\sstitle{Open-domain Image Editing}
The objective of open-domain image editing is to edit images based on open-vocabulary instructions that delineate both the elements to be altered and those to be added, such as changing a ``red apple'' into a ``green apple''~\cite{liu2020open}. This process faces multiple challenges: first, generating plausible manipulation instructions for each training image is difficult, and collecting a large dataset of accurately manipulated images for fully supervised training is unfeasible. Secondly, open-domain images exhibit a much higher degree of variability than the single-domain images like flowers or birds used in previous studies~\cite{}. Lastly, the resulting manipulated images might not preserve all original details. Existing approaches in language-guided image editing, which rely on captions from similar-domain images for training, falter in this broader context as they cannot appropriately handle the diversity of open-domain images, where captions applicable to one type of image are unsuitable for another~\cite{liu2020open}.

\sstitle{Iterative Editing}
Fu et al.~\cite{fu2020sscr} describes a task where an editor function  modifies an image \( V_{t-1} \) based on an instruction \( I_t \) at each turn \( t \), resulting in a new image \( V_t \):
\begin{equation}
	V_t = Editor(V_{t-1}, I_t)
\end{equation}
After the final turn \( T \), the model's performance is evaluated by comparing the predicted image \( V_T \) with the ground truth \( O_T \) using a function \( Compare(V_T, O_T) \). This process aims for precise pixel-level image generation.

\sstitle{Iterative Multi-granular Editing}
Joseph et al.~\cite{joseph2024iterative} introduce Iterative Multi-granular Image Editing (IMIE), which allows users to iteratively and precisely edit images. Unlike one-shot generation methods, IMIE offers control over both global and local edits.
Given an input image \( I_0 \), edit instructions \( E = \{y_1, \ldots, y_k\} \), and optional masks \( M = \{m_1, \ldots, m_k\} \), the image editor \( \mathcal{M}(I_i, y_i, m_i) \) applies semantic modifications iteratively. If a mask \( m_i \) is provided, edits are constrained to the masked area. The resulting images \( I_{\text{edits}} = \{I_1, \ldots, I_k\} \) are visually and semantically consistent with the instructions and masks.

\sstitle{Multi-instruction Editing}
Guo et al.~\cite{guo2023focus} propose a method involves editing an input image \( I \) based on a composite instruction \( T \) with \( k \) sub-instructions \( \{T_1, T_2, \ldots, T_k\} \). The aim is precise and harmonious execution of these instructions.

\autoref{tab:notations} summarizes some important notations in this paper.

\begin{table}[h!]
\centering
\footnotesize
\caption{Summary of Important Notations}
\label{tab:notations}
\begin{tabular}{|p{2cm}|p{6cm}|}
\hline
\textbf{Notation} & \textbf{Description} \\
\hline
$x_t$ & Image state at each editing step  \\
$G(z_t, h_t, f_{G_{t-1}})$ & Generator function with noise vector $z_t$, context-aware $h_t$, and previous image state $f_{G_{t-1}}$  \\
\hline
$L_{ins}$ & Instruction tuning loss (e.g., cross-entropy) \\
$L_{total}$ & Total loss function combining multiple components  \\
$L(\mathcal{E}(x + \delta), \mathcal{E}(x))$ & Loss to maximize embedding distance between original and altered images  \\
\hline
$\mathcal{E}(x)$ & Embedding representation of an image $x$  \\
$e_{edit} = e_{after} - e_{before}$ & Edit direction embedding, used for transformations \\
\hline
$\delta$ & Perturbation term optimized for adversarial settings  \\
$\sqrt{\alpha_t} x_0 + \sqrt{1 - \alpha_t} \epsilon$ & Diffusion process equation, with $\alpha_t$ as noise control parameter  \\
\hline
$\|\delta\|_p \leq \xi$ & Constraint on perturbation $\delta$ with maximum budget $\xi$  \\
$x_0, x_T$ & Initial and terminal images in diffusion processes  \\
\hline
$z_t$ & Noise vector at time $t$ used in GANs  \\
$f_{G_{t-1}}$ & Context-free condition encoding the prior image  \\
$h_t$ & Context-aware condition (e.g., instruction history)  \\
\hline
\end{tabular}
\end{table}

\subsection{Feature Engineering}

\sstitle{Image embedding}
The image encoder is often a multi-layer convolutional neural network (CNN) that takes an input source image of dimensions \( H \times W \) and outputs a spatial feature map of dimensions \( M \times N \)~\cite{chen2018language}. Each position on this feature map holds a \( D \)-dimensional feature vector, effectively creating a set of feature vectors \( V = \{v_i : i = 1, \ldots, M \times N\}, v_i \in \mathbb{R}^D \).

\sstitle{Text embedding}
The text encoder is often a recurrent Long Short-Term Memory (LSTM) network~\cite{chen2018language}. It processes a natural language expression of length \( L \), first embedding each word into a vector using a word embedding matrix. Then, the LSTM generates a contextual vector for each word that captures its contextual information, including word order and word-word dependencies. The output is a language feature map \( U = \{u_i : i = 1, \ldots, L\}, u_i \in \mathbb{R}^K \), where each \( u_i \) is a \( K \)-dimensional vector representing the embedded word.

\sstitle{Visual-semantic embedding}
Liu et al.~\cite{liu2020open} employs a visual-semantic embedding (VSE) space, learned from a large image-caption dataset with CNNs and LSTMs encoding the visual and textual information. Training of these encoders uses a triplet ranking loss \( \mathcal{L}(v, t) \), defined as~\cite{faghri2017vse}:
\begin{equation}
\mathcal{L}(v, t) = \max[m + \langle v, \hat{t} \rangle - \langle v, t \rangle, 0] + \max[m + \langle \hat{v}, t \rangle - \langle v, t \rangle, 0]
\end{equation}
Here, \( v \) and \( t \) are the feature vectors of a matching image-caption pair, while \( \hat{v} \) and \( \hat{t} \) are the feature vectors of the hardest non-matching pair in the batch. The term \( m \) is a margin that the correct pairs must exceed in similarity over the incorrect pairs, pushing the model to differentiate between them effectively. After training, the visual feature maps are also embedded into the VSE space.

\begin{figure}
    \centering
    \includegraphics[width=.95\linewidth]{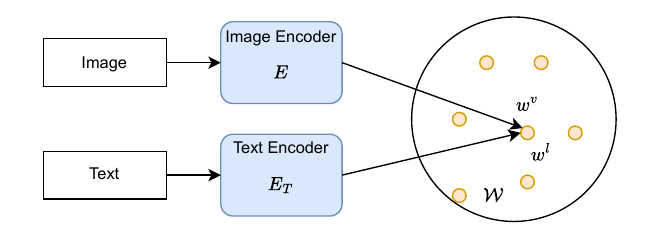}
    \caption{Visual-linguistic similarity learning.}
    \label{fig:visual-linguistic-sim-learning}
\end{figure}

\sstitle{Visual-linguistic similarity learning}
Xia et al.~\cite{xia2021tedigan} propose a visual-linguistic similarity learning to train a text encoder that aligns with an image encoder in the \(\mathcal{W}\) space (see \autoref{fig:visual-linguistic-sim-learning}). The encoder is trained to map both images and their descriptions into this space using a previously trained image encoder~\cite{dong2017semantic}. The alignment is achieved using a visual-linguistic similarity module that encodes the text and image into a shared embedding space~\cite{wang2020consensus}, formulated as:
\begin{equation}
\min_{\Theta_{E_l}} \mathcal{L}_{E_l} = \left\|\sum_{i=1}^L p_i\left(w_i^v - w_i^l\right)\right\|_2^2,
\end{equation}
where \(w_i^v\) and \(w_i^l\) are the projected codes of the image and text embeddings, respectively.

\begin{figure*}
    \includegraphics[width=0.9\linewidth]{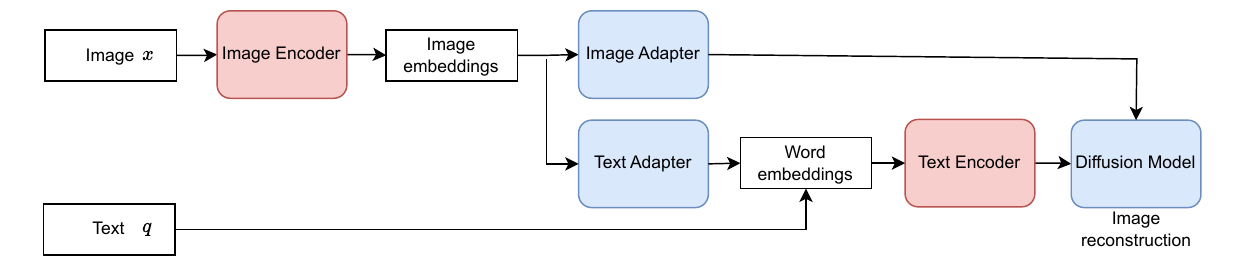}
    \caption{The dual-branch concept extraction.}
    \label{fig:dual-branch}
\end{figure*}

\sstitle{Dual-branch concept extraction}
Chen et al.~\cite{chen2023photoverse} propose a dual-branch concept extraction process for facial feature extraction and image generation (see \autoref{fig:dual-branch}). Initially, face detection identifies and localizes faces in an image \(x\), which are then expanded, cropped, and resized. A mask is applied to focus on facial attributes by removing non-facial elements. This preprocessed image is used for denoising during diffusion model training.
For textual conditions, Chen et al.~\cite{chen2023photoverse} embed the reference image into the textual word embedding space using pseudo-words \(S^*\)~\cite{gal2023an}. The CLIP image encoder extracts features, which are translated into multi-word embeddings \(S^*\) using a multi-adapter architecture, ensuring efficient and accurate transformation~\cite{wei2023elite}.
To enhance performance, the visual condition includes image space features as an auxiliary aid. The CLIP image encoder's features are mapped with an image adapter, capturing essential visual cues related to identity for accurate attribute representation during image generation.

\sstitle{Diffusion-based encoding}
Diffusion probabilistic models are generative models that invert a diffusion process~\cite{ho2020denoising}. The objective function for these models is:
\begin{equation}
L = \mathbb{E}_{x_0, t, \epsilon} \|x_t - \epsilon_{\theta}(x_t, t)\|_2^2,
\end{equation}
where \(\epsilon_\theta\) is the noise estimator. Starting from an input image \(x_0\), the process yields \(x_t\) based on \(\sqrt{\alpha_t}x_0 + \sqrt{1 - \alpha_t}\epsilon\), where \(\alpha_t\) controls the level of noise.
Denoising diffusion implicit model (DDIM), a deterministic variant of diffusion models, can be used to generate new images~\cite{song2021denoising}. Given an input \(x_T \sim \mathcal{N}(0, I)\), the update rule is:
\begin{equation}
x_{t-1} = \sqrt{\alpha_{t-1}}\left(\frac{x_t - \sqrt{1 - \alpha_t}\epsilon_\theta(x_t, t)}{\sqrt{\alpha_t}}\right) + \sqrt{1 - \alpha_{t-1}}\epsilon_\theta(x_t, t).
\end{equation}
The variable \(x\) is updated in small steps using the neural ordinary differential equation (ODE) \(u = x/\sqrt{\alpha}\) and \(\tau = \sqrt{1/\alpha - 1}\), with:
\begin{equation}
du = \epsilon_\theta\left(u/\sqrt{1 + \tau^2}, t\right)d\tau.
\end{equation}
DDIM sampling follows an Euler scheme to solve this ODE, which is parameterized over time \(t\)~\cite{song2021denoising}. Many works use DDIM encoding to transform \(x_0\) into a latent variable \(x_r\)~\cite{couairon2023diffedit}. This process preserves key image attributes, allowing subsequent decoding to reconstruct \(x_0\). This encoding is particularly useful for image editing.

\sstitle{Multi-scale diffusion encoding}
DDPM~\cite{ho2020denoising} uses a Markov chain to add and then remove noise from images, reconstructing the original image. SinDDM enhances this by using a multi-scale pyramid, starting denoising from the smallest scale and progressing through up-sampling.
The forward process for a given scale \( s \) is:
\[ q(x_t^s|x_{t-1}^s) = \mathcal{N}(x_t^s; \sqrt{1-\beta_t^s} x_{t-1}^s, \beta_t^s I) \]
with \( t \) as the noise addition steps, and \( \beta_t^s \) as the variance. The model predicts the noise \( \epsilon \) to denoise the image:
\[ p_\theta(x_{t-1}^s|x_t^s) = \mathcal{N}(x_{t-1}^s; \mu_\theta(x_t^s, t), \beta_t^s I) \]
The training loss function is:
\[ L_{t-1}^s = \mathbb{E}_{x^s, \epsilon} \left[ \| \epsilon - \epsilon_\theta(\sqrt{\bar{\alpha}_t} x^s + \sqrt{1-\bar{\alpha}_t} \epsilon, t, s) \|_2^2 \right] \]

\subsection{Control Mechanisms}
\begin{figure}
    \centering
    \includegraphics[width=.95\linewidth]{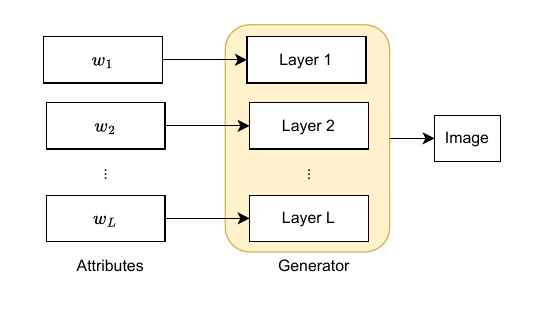}
    \caption{The control mechanism using StyleGAN. $L$ is the number of generator layers.}
    \label{fig:stylegan}
\end{figure}

Xia et al.~\cite{xia2021tedigan} propose a control mechanism based on style mixing in StyleGAN to manage attributes and objects within images. Layerwise analysis of StyleGAN~\cite{karras2019style} shows that different layers are responsible for different attributes, ranging from broad features like head pose to fine details such as freckles (see \autoref{fig:stylegan}). The mechanism mixes layers of a style code \(w^s\) with the content code \(w^c\), using either a random latent code for text-to-image generation or an image embedding for text-guided manipulation.
The control mechanism enables mapping texts and images into a common latent space, allowing for attribute-specific selection and synthesis.
Xia et al.~\cite{xia2021tedigan} also support multiple modalities, where the style code \(w^s\) and content code \(w^c\) can include sketches, labels, images, and noise. This enhances the accessibility, diversity, controllability, and accuracy, especially for multi-modal synthesis.

\section{Instruction Foundations}

\subsection{LLM Empowering}

\begin{figure*}
    \centering
    \includegraphics[width=0.8\linewidth]{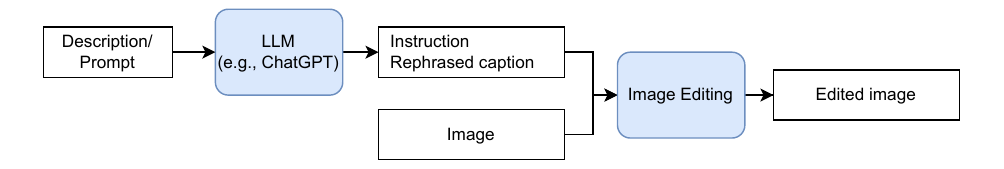}
    \caption{LLM-based instruction generation.}
    \label{fig:llm}
\end{figure*}

Brooks et al.~\cite{brooks2023instructpix2pix} approaches the instruction-based image editing as a supervised learning task. It involves two main steps: creating a paired training dataset with text instructions and corresponding before-and-after edit images, and then training an image editing diffusion model using this dataset (see \autoref{fig:llm}). The trained model is capable of generalizing from the synthetic training data to perform real image edits based on instructions written by humans.

The creation of a multi-modal training dataset involves two key processes using large-scale pretrained models~\cite{brown2020language,rombach2022high}. First, GPT-3~\cite{brown2020language} is fine-tuned to generate text edits by producing change instructions and modified captions from 700 manually annotated samples, yielding 454,445 examples. Second, a text-to-image model \cite{rombach2022high} converts these captions into images. To ensure image consistency, the Prompt-to-Prompt method~\cite{hertz2023prompt} uses shared attention weights during denoising. For varying degrees of change, 100 image pairs are generated per caption pair with random shared attention weight ratios~\cite{schuhmann2022laion} and filtered using a CLIP-based directional similarity metric~\cite{gal2022stylegan} to align image changes with captions. This process enhances dataset diversity and quality while mitigating model inconsistencies.

Similarly, Li et al.~\cite{li2023moecontroller} use high-quality images from lion-5b~\cite{schuhmann2022laion} to create a global transformation dataset. First, ChatGPT generates diverse target image captions from the original captions and reference examples. These captions are used to train the ControlNet model to produce pairwise global manipulation images. Six image condition extraction methods and various text scale coefficients are employed for batch image generation. Finally, the images are filtered by resolution, aesthetic scores, and CLIP scores to build a comprehensive dataset for model learning. To enhance instruction diversity, some works manually write dozens of instructions per task and used GPT-4 to rewrite and expand them, simulating user input~\cite{geng2023instructdiffusion}. During training, one instruction is randomly selected to incorporate diverse and intuitive instructions~\cite{wei2022finetuned}, significantly improving the model's multi-task fusion capabilities.

Gan et al.~\cite{gan2024instructcv} propose the creation of a multi-modal and multi-task instruction-tuning dataset by combining MS-COCO \cite{lin2014microsoft}, ADE20K \cite{zhou2019semantic}, Oxford-III-Pets \cite{parkhi2012cats}, and NYUv2~\cite{silberman2012indoor}, covering tasks such as semantic segmentation, object detection, monocular depth estimation, and image classification. The dataset \( D \) includes input images \( x_i \), task outputs \( y_i \), and task identifiers \( m_i \).
This dataset is converted into \( D_T \), where each entry includes \( x_i \), \( y_i \), and a natural language instruction \( I_i \). The task identifier \( m_i \) is expressed in \( I_i \), and \( y_i \) is visually represented as \( v(y_i) \).
Instruction generation uses fixed prompts, such as ``Segment \%category\%'', and a language model (LLM) generates diverse paraphrased instructions~\cite{raffel2020exploring}. The dataset includes both fixed prompt instructions \( D_{FP} \) and rephrased prompt instructions \( D_{RP} \).
Visual encoding methods vary by task: for semantic segmentation, \( v(y) \) is a binary mask; for object detection, \( v(y) \) includes bounding box coordinates; for depth estimation, \( v(y) \) converts depth values to RGB; and for image classification, \( v(y) \) uses color-coded prompts. This diverse dataset facilitates effective training and evaluation of models across multiple vision tasks.

\subsection{MLLM Empowering}
\begin{figure*}
    \centering
    \includegraphics[width=0.8\linewidth]{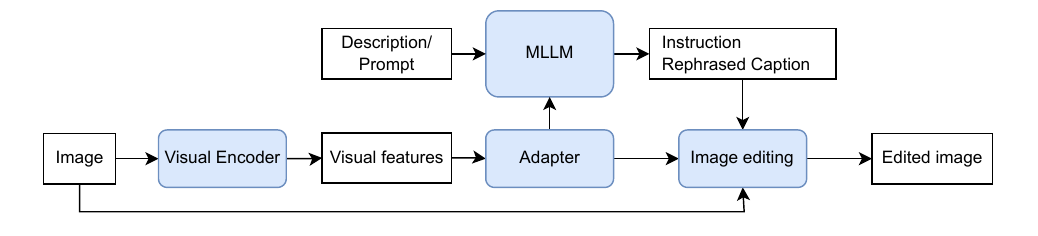}
    \caption{MLLM-empowered instruction generation.}
    \label{fig:mllm}
\end{figure*}

Multimodal LLMs (MLLMs), extensions of LLMs with additional sensory inputs like images, are designed understand and generate content aligned with images. These MLLMs are built upon a base LLM and enhanced with a visual encoder, such as CLIP-L~\cite{radford2021learning}, to extract visual features denoted by \( f \) from a visual data $V$, and an adapter \( W \) to integrate these features with the language model, as shown in \autoref{fig:mllm}. They are trained with a sequence of $l$ tokens \( C \) to predict the next token given the previous tokens and the visual context~\cite{liu2024visual}:
\begin{align}
\begin{split}
	& C = \{x_1, x_2, ..., x_l\}, \\
	& f = encode(V), \\
	& x_t = MLLM(\{x_1, ...x_{t-1}\} | W(f))
\end{split}
\end{align}
Here, \( C \) can be an image caption for feature alignment or data for following multimodal instructions. MLLMs undergo autoregressive training to predict the next token, enabling them to assist with visual tasks such as question answering and reasoning~\cite{fu2024guiding}. 

Fu et al.~\cite{fu2024guiding} proposes MLLM-Guided Image Editing (MGIE), a method to edit images using a multimodal large language model. 
MGIE works by generating an explicit instruction \( \mathcal{E} \) from the input prompt and image, simplified by a pre-trained summarizer. The instruction generation is formulated as:
\begin{align}
\begin{split}
 & \mathcal{E} = Summ(MLLM^*([\text{prompt}, \mathcal{X}] | W(f)))  = \{w_1, w_2, ..., w_l\}, \\
& w_t' = MLLM(\{w_1, ..., w_{t-1}\} | W(f)), 
\end{split}
\end{align}
where \( w_t' \) is the predicted word token, and the loss for instruction tuning is calculated using cross-entropy loss (CELoss) over the sequence length \( l \):
\begin{equation}
\label{eq:instruction_loss}
 \mathcal{L}_{ins} = \sum_{t=1}^{l} CELoss(w_t', w_t) 
\end{equation}
This training strategy improves the model's understanding of instructions and mitigates comprehension gaps, enabling it to generate a visual imagination for editing that is still confined to text. Visual tokens, denoted as \( [IMG] \), are added into the instruction \( \mathcal{E} \) to further enhance the model's visual understanding~\cite{koh2024generating}.

Another MLLM-empowered framework is SmartEdit~\cite{huang2023smartedit}, which initially struggled with perception and reasoning when using only conventional datasets like InstructPix2Pix~\cite{brooks2023instructpix2pix} and MagicBrush~\cite{zhang2024magicbrush}. The primary issues identified were a poor understanding of position and concepts due to the UNet in the diffusion model, and limited reasoning capabilities due to insufficient exposure to relevant data. To address these, the training was augmented with reasoning segmentation~\cite{lai2023lisa}, significantly improving perception abilities.

Zhang et al.~\cite{zhang2024effived} generate target video \( V_t \) from source video \( V_s \) and instruction \( c \) without fine-tuning. This involves using video triplets and applying random transformations (rotation, crop, translation, shearing) to create frames simulating camera movements.
The method leverages LLMs like InstructPix2Pix to create datasets by applying random transformations to pairs of original and edited images. This generates frames that simulate camera movements, enhancing training data diversity. Open-world videos are used with CoCa~\cite{yu2022coca} and VideoBLIP~\cite{zhang2024effived} to extract video captions, summarised by ChatGPT, to generate editing instructions and captions. CoDeF~\cite{ouyang2024codef}, enhanced with dynamic Neural Radiance Fields (NeRFs) \cite{park2021nerfies}, creates edited videos that preserve motion, improving video quality and consistency.

\subsection{Instruction Types}

\sstitle{Caption-based}
Many works generate target images from source images (such as sketches, grayscale, or natural images) using natural language caption~\cite{chen2018language}. Caption-based instruction has applications ranging from Computer-Aided Design (CAD) to Virtual Reality (VR). 
For instance, a fashion designer can modify a sketch of shoes based on a customer's verbal description, with the LBIE system generating the revised image. In another example, the system alters a virtual environment scene according to a user's textual description. Additionally, it can transform a sketch and grayscale image of a flower into a color image based on a descriptive text.
The editing system can handle tasks like segmentation and colorization, ensuring that generated images align with the given descriptions and common sense, providing a natural user interface for future VR systems~\cite{chen2018language}.

Mask-based caption allows users to specify precise areas for modification but requires laborious manual mask creation, limiting user experience~\cite{couairon2022diffedit}. In contrast, mask-free captions do not need masks and directly modify the image's appearance or texture, although they struggle with local modifications and detailed mask accuracy~\cite{lin2023text}. 

\sstitle{Command-based}
Command-based instruction involves guiding image changes through specific textual commands, such as ``red apple $\rightarrow$ green apple''. These instructions are encoded into a joint embedding space along with the image's visual features, enabling a unified processing approach. For example, Liu et al.~\cite{liu2020open} manipulates the visual feature maps of the source image based on these encoded instructions, and an image decoder generates the final manipulated image from these updated features. 

\sstitle{Counterfactual descriptions}
Fan et al.~\cite{fan2023target} propose generating both factual and counterfactual descriptions for an image, and then use a cross-modal interpreter to distinguish between them to ensure accuracy in the generated images. 
The factual description \( S^f \) replaces the [LOC] placeholder with location adverbs extracted from the instruction \( T \) using CoreNLP~\cite{manning2014stanford}. The [OBJ] placeholder is replaced with object labels \( o^r_f \) and \( o^t_f \) for reference and target images respectively, identified by comparing labels \( O_r \) and \( O_t \) through specific rules based on their cardinality differences. Counterfactual descriptions \( S^c \) are created by randomly replacing tokens in \( S^f \).
The reasoning module \( R \) is trained using T5 for sequence-to-sequence tasks. The first task predicts image-level labels \( O_t \) by observing \( O_r \), \( T \), and \( T_{\text{loc}} \). The second task generates instructions \( T' \) by combining \( O_r \), \( O_t \), and \( T \). The combined sequence \( \hat{T} = T_{\text{loc}} \oplus T^o_r \oplus T^o_t \) allows \( R \) to predict \( T' \) and maintain the semantic and visual integrity of the generated images.

\sstitle{Dialog-based Instruction}
Jiang et al.~\cite{jiang2021talk} propose a dataset, called CelebA-Dialog, specifically designed for dialog-based facial editing scenarios, which often require multiple rounds of edits to satisfy users. In such cases, the editing system must be capable of generating continuous and fine-grained facial editing results that translate intermediate states from source to target images. However, for most facial attributes, binary labels are insufficient to accurately capture the degrees of these attributes. Methods trained solely on binary labels often struggle with fine-grained editing, especially as attribute degrees become larger.

Motivated by these challenges, the CelebA-Dialog dataset offers a solution.
First, it includes rich fine-grained labels that classify each attribute into multiple degrees based on their semantic meaning. Second, it provides textual descriptions for each image, which contain captions describing the attributes and a user request template. These features enhance the dataset's usefulness in dialog-based facial editing. The dataset utilises the CelebA dataset~\cite{liu2015deep}, which contains 202,599 images with binary attribute annotations. For the CelebA-Dialog dataset, five attributes were selected for fine-grained labeling: bangs, eyeglasses, beard, smiling, and young (age). 
Each image has associated captions and a user request template, enriching the dataset and facilitating conditional requests.

Yue et al.~\cite{yue2023chatface} use a LLM to interpret and execute user editing requests. Editing queries are parsed to extract desired attributes, editing strength, and diffusion steps. The LLM then maps these attributes to semantic offsets and generates the edited image. This system allows users to perform precise image edits while maintaining consistency and identity. By understanding and decomposing user intentions into structured attributes, the LLM facilitates detailed and controlled image manipulation.

DialogPaint~\cite{wei2023dialogpaint} accepts user images and editing instructions via dialogue, clarifying vague instructions and providing feedback to complete image edits. The methodology involves creating a dialogue-based image editing dataset and model. To build the datasets, image captions are sourced from CUB-200-2011, Microsoft COCO~\cite{riemenschneider2014learning}, DeepFashion~\cite{liu2016deepfashion}, and Flickr-Faces-HQ (FFHQ)~\cite{karras2019style}. These captions are combined with prompt instructions using text-davinci-003 to generate the dialogue data~\cite{wang2023self}. For the image editing dataset, 10,000 image-text pairs are selected and processed to create editing instructions, resulting in 6468 verified pairs. DialogPaint consists of two main components: the Dialogue Model and the Image Editing Model~\cite{tao2023net,jiang2022text2human}. The Dialogue Model uses pre-trained Blender dialogue model~\cite{roller2021recipes} weights to generate high-quality, contextually relevant editing instructions. These instructions are crucial for the Image Editing Model, which performs edits based on the dialogue-derived instructions.

Cheng et al.~\cite{cheng2020sequential} present a task where users guide an agent to edit images through multi-turn textual commands. Each session involves the agent modifying an image based on user descriptions from previous turns. The main challenges are ensuring consistency between the image and text and maintaining visual coherence.

Fu et al.~\cite{fu2020sscr} discuss a task that involves following iterative instructions to edit images step by step. The main challenge is data scarcity, as it is difficult to collect large-scale examples of images before and after instruction-based changes. Humans can achieve these tasks by leveraging counterfactual thinking, which allows them to consider alternative events based on previous experiences.
The authors further emphasize the potential benefits of automating digital design tools like Illustrator or Photoshop using dialog-based instruction to enhance accessibility. Existing models like GeNeVA~\cite{el2019tell} neglect data scarcity issues and highlights the importance of counterfactual thinking, which allows humans to operate in data-scarce scenarios by considering alternative instructions from seen examples.

\sstitle{Example-based Instruction}
Some works argue the limitations of text-based image editing~\cite{kawar2023imagic,kim2022diffusionclip,nichol2022glide,ruiz2023dreambooth} and suggests images as a better alternative for conveying complex ideas~\cite{yang2023paint}. 
ImageBrush~\cite{yang2023imagebrush} formulates the problem of exemplar-based image manipulation, aiming to generate an edited image \( I' \) from a pair of examples \( \{E, E'\} \) and a query image \( I \). Some works also use a binary mask to identify editable regions~\cite{yang2023paint}. The goal is to produce an image that follows the instructions implied by the examples, which involves understanding their relationships and correlations~\cite{yang2023paint}.
The proposed solution, namely ``Progressive In-Painting'', uses cross-attention to integrate the examples and query image into a general context. To overcome challenges in capturing detailed information with cross-attention, the method employs self-attention for learning low-level context. This involves creating a grid-like image \( \{E, E', I, M\} \) with a blank image \( M \) and iteratively recovering \( \{E, E', I, I'\} \).

Yang et al.~\cite{yang2023paint} propose another solution involving understanding the reference image, synthesizing a transformed view that fits the source image, and inpainting the surrounding areas for a smooth transition, possibly requiring super-resolution. The proposed solution includes several principles: a content bottleneck to force the model to understand and regenerate the reference image content, strong augmentation techniques to mitigate train-test mismatch, mask shape augmentation to handle arbitrary-shaped masks, and control of the similarity degree between the edited area and the reference image using classifier-free guidance~\cite{ho2022classifier}.

\subsection{Instruction Augmentation}

\sstitle{Transformation-oriented Instruction Initialisation}
Zhao et al.~\cite{zhao2024instructbrush} propose transformation-oriented instruction initialization by using unique semantic phrases to initialize target concept learning, avoiding manual captioning~\cite{daras2022multiresolution,zhang2023prospect}. The similarity between phrases is calculated as:
\[ \text{sens} (\langle p_{x} \rangle) = \text{sim} (\langle p_{x}, \{x\} \rangle) - \text{sim} (\langle p_{x}, \{y\} \rangle) \]
with the condition for including a unique phrase:
\[ \langle p_{x} \rangle = \begin{cases} 
\langle p_{x} \rangle & \text{if } \text{sens} (\langle p_{x} \rangle) \geq \eta \\
\emptyset & \text{otherwise}
\end{cases} \]
These unique phrases help capture transformation details more effectively, improving the model's ability to guide new image generation~\cite{wen2024hard}.

\sstitle{Caption-based Augmentation}
Zhang et al.~\cite{zhang2024hive} extends the LLM-empowered method~\cite{brooks2023instructpix2pix} to generate instructions, starting with a dataset of 1,000 images with their captions. Three annotators provided three sets of instructions and corresponding edited captions for each, resulting in 9,000 data points consisting of the original caption, the instruction, and the edited caption. This was used to fine-tune GPT-3 with OpenAI API v0.25.0~\cite{brown2020language}. Additionally, GPT-3 was employed to create five sets of instructions and edited captions for each image-caption pair from the Laion-Aesthetics V2 dataset~\cite{schuhmann2021laion}, which were then augmented with more descriptive captions using BLIP due to Laion's captions lacking visual detail~\cite{li2022blip}. For image generation, the Prompt-to-Prompt method~\cite{hertz2023prompt} based on stable diffusion was used. A cycle-consistent augmentation was also designed to expand the dataset. In total, 1.17 million training triplets were generated and combined with an existing 281,000 from another source~\cite{brooks2023instructpix2pix}, resulting in a dataset of 1.45 million training image pairs with instructions.

\sstitle{Cycle Consistency Augmentation}
Zhang et al.~\cite{zhang2024hive} applies the cycle consistency augmentation for image editing, a concept borrowed from image-to-image generation~\cite{isola2017image,zhu2017unpaired}. Cycle consistency involves bi-directional mappings \( F \) and \( G \) such that \( F(G(X)) \approx X \) and \( G(F(Y)) \approx Y \), enhancing the generative model's capabilities. For instructional image editing, the process incorporates a forward pass \( F: x \xrightarrow{\text{inst}} \hat{x} \) and a reverse pass \( R: \hat{x} \xrightarrow{\text{inst}^{-1}} x \) to establish a loop that ensures edits are reversible -- for example, ``add a dog'' is reversed by ``remove the dog''. To implement this, Zhang et al.~\cite{zhang2024hive} distinguishes between invertible and non-invertible instructions within the dataset, using natural language processing techniques like speech tagging and template matching with the Natural Language Toolkit (NLTK). By analyzing and summarizing invertible instructions into templates, the method identifies 29.1\% of the instructions as invertible. This data is then augmented to support cycle consistency in the training set, creating a more robust training environment for the model.

\sstitle{Reasoning Segmentation}
Existing approaches face two main challenges: poor perception of position and concept, and limited reasoning capability despite having MLLM~\cite{huang2023smartedit}. These issues arise from the UNet's lack of understanding and limited exposure to reasoning-based data. To address this, segmentation data was added to improve perception, and reasoning segmentation data was incorporated to enhance reasoning abilities~\cite{lai2023lisa}. A data production pipeline generated 476 paired samples, including scenarios with complex understanding and reasoning requirements~\cite{huang2023smartedit}. For complex understanding scenarios, it starts with two internet images, x1 and x2, identifying specific animals using the SAM algorithm~\cite{kirillov2023segment}. In x1, a cat (mask1) is replaced with a rabbit (mask2) from x2. The inpainting algorithm MAT is applied to remove the cat from x1, creating y1. The rabbit is resized and filtered, then merged with y1 to form y3. Harmonization algorithm PIH~\cite{wang2023semi} adjusts saturation and contrast to create y4. This process produces two training pairs: (y1, x1) with instruction ``Add a cat to the right of the cat'' and (x1, y4) with instruction ``Replace the smaller cat with a rabbit.'' For reasoning scenarios, object masks are generated using SAM, and stable diffusion~\cite{rombach2022high} is used for inpainting based on instructions. Manual filtering removes unsatisfactory results.

\sstitle{Refinement Prior}
An MLP projector can be used to map the output of a multi-modal encoder to the embedding space of a LLM~\cite{li2023instructany2pix}. The goal is to align these spaces using many image-text and audio-text pairs, with all model parts except the projector frozen. 
To address low-quality images, Li et al.~\cite{li2023instructany2pix} develops a refinement module, incorporating a transformer, improves image quality in the embedding space by predicting the ground-truth image embedding and applying L2 regression loss.

\sstitle{Noise Filtering}
Chakrabarty et al.~\cite{chakrabarty2023learning} propose a method for curating a high-quality training dataset, focusing on underspecification, grounding, and faithfulness. It handles noisy edit instructions by leveraging large language models and Chain-of-Thought prompting~\cite{wei2022chain} to predict the appropriate context and generate the necessary entity or region for the edit.
For grounding, it uses object detection and segmentation. ChatGPT identifies the edit entity, GroundingDINO~\cite{liu2023grounding} locates it, and the SAM model~\cite{kirillov2023segment} segments it, ensuring precise and contextually appropriate edits.
Stable Diffusion Inpainting generates images based on text inputs and segmentation masks, using "edited captions" from the dataset. Multiple images are created and re-ranked for faithfulness.
Faithfulness is ensured through Visual Question Answering (VQA) techniques. Questions are generated about unmodified elements in the image using the Vicuna-13B model~\cite{touvron2023llama}. For instance, with the caption ``Buttermere Lake District with Aurora Borealis'' and the instruction ``Remove Aurora Borealis'', questions like ``Is there a lake district in the picture?'' ensure the edit's success. The edited images are evaluated using the BLIP-2 VQA model~\cite{li2023blip}, selecting the image with the most correct answers.

\sstitle{Unification}
Meng et al.~\cite{meng2024instructgie} propose a method called language instruction unification to address the instability of language instructions~\cite{yang2023imagebrush}. They process 50\% of the training data's instructions \( l \) through a frozen lightweight language model (LLM), Open Llama 3b V2 Quant 4 (\( U \)), to create unified instructions \( l' \):
$ l' = U(l)$.
These unified instructions augment the training data. During inference, instructions \( l \) are unified by \( U \) before processing. This method ensures consistent learning from diverse instructions, improving the model's ability to generalise to new tasks.

\sstitle{On-the-fly edit generation}
The process of obtaining the edit direction embedding involves using CLIP within the image generator diffusion model. Two textual captions are required: \( c_{\text{before}} \) (before editing) and \( c_{\text{after}} \) (after editing). The embeddings for these captions, \( e_{\text{before}} \) and \( e_{\text{after}} \), are vectors whose difference provides the edit direction embedding \( e_{\text{edit}} \):
\[ e_{\text{before}} = \text{CLIP}(c_{\text{before}}) \]
\[ e_{\text{after}} = \text{CLIP}(c_{\text{after}}) \]
\[ e_{\text{edit}} = e_{\text{after}} - e_{\text{before}} \]
A more refined approach~\cite{santos2024pix2pix} uses multiple before-edit and after-edit captions, averaging the embeddings for each set and then calculating their difference.
Parmar et al.~\cite{parmar2023zero} used GPT-3 to generate numerous before-edit and after-edit captions, but this method is labor-intensive. Santos et al.~\cite{santos2024pix2pix} propose a new method generating these edit directions dynamically based on user input using the Phi-2 LLM~\cite{gunasekar2023textbooks}, known for its compact size and high performance. A specific prompt template guides the generation, allowing for the creation of before-edit and after-edit captions based on the desired transformation, from which the edit direction embedding is then computed.

\sstitle{Chain-of-thought}
Zhang et al.~\cite{zhang2024tie} propose a data preparation process using Chain-of-Thought (CoT). The process begins with an editing instruction that specifies modifications to a source image. 
The GPT-4V model is utilized to generate CoT responses based on three prompt templates: instruction decomposition, region localization, and detailed description. For instance, the instruction is decomposed into specific tasks: placing a glass of soda, retaining only one vase of flowers, and adding a bottle of beer. The model then localizes the regions in the original image, identifying which vases need to be removed to align with the instruction of having just one vase. Detailed descriptions of the areas are generated, explaining the specific changes made to achieve the target image.

The Chain-of-Thought (CoT) pipeline for image editing involves three phases: instruction decomposition, region localization, and detailed description. In instruction decomposition, complex instructions are broken down into simpler sub-prompts for manageable editing. Region localization identifies specific areas in the image for each sub-prompt, ensuring precise edits like adding or removing objects. Detailed description provides thorough characterizations of target areas, enabling accurate and seamless modifications. Supported editing operations include ADD, REMOVE, and CHANGE (object, attribute, background). This method ensures sophisticated and contextually accurate image edits.

The responses from GPT-4V for each of these prompts are encapsulated into a sample set, represented as \( S_i = \{ X_i, I_i, M_i, T_i, R_i \} \), where \( X_i \) is the original image, \( I_i \) is the editing instruction, \( M_i \) is the masked image, \( T_i \) is the target image, and \( R_i \) is the set of responses. This sample is then used for fine-tuning the LISA model~\cite{yang2023improved}, enhancing its ability to generate precise masks and inpainting prompts based on detailed area descriptions. This approach ensures the edited image aligns accurately with the given instructions, improving the model's performance in image editing tasks.

\section{Editing Controls}
\label{sec:method}

\begin{table*}[!h]
\centering
\caption{Highlighted image editing papers and their available tools.}
\label{tab:tools}
\begin{adjustbox}{max width=\textwidth}
    \begin{threeparttable}
\begin{tabular}{lllll}
\toprule
\textbf{Model Designs} & \textbf{Learning Strategies} & \textbf{Instruction Mechanisms} & \textbf{Learning Augmentations} & \textbf{Loss Functions} \\
\midrule
Image encoder~\cite{chen2018language} & Adversarial training~\cite{cheng2020sequential,li2020manigan,fu2020sscr} & Command-based~\cite{liu2020open} & Cycle consistency augmentation~\cite{zhang2024hive,chandramouli2022ldedit} & Ranking loss~\cite{xia2021tedigan} 
\\
Language encoder~\cite{chen2018language} & Counterfactual reasoning~\cite{fu2020sscr} & Caption-based~\cite{chen2018language,fan2023target} & Reasoning segmentation~\cite{lai2023lisa,huang2023smartedit} &  Perceptual loss~\cite{liu2020open,xia2021tedigan,fan2023target,li2020lightweight,bodur2023iedit}
\\
Visual-semantic embedding~\cite{liu2020open} & Contrastive Pretraining~\cite{patashnik2021styleclip}  & Dialog-based~\cite{cheng2020sequential,jiang2021talk} & Perturbation regularisation~\cite{liu2020open} & Adversarial hinge loss~\cite{el2019tell,liu2020open,cheng2020sequential} 
\\
Bidirectional interaction~\cite{huang2023smartedit} & Instance-level preservation~\cite{xia2021tedigan}  &  MLLM-empowered~\cite{huang2023smartedit} & Cross-task consistency~\cite{fu2020sscr} & Multimodal matching loss~\cite{liu2020open,cheng2020sequential,li2020lightweight} 
\\
Recurrent attention~\cite{chen2018language} & Classifier-free guidance~\cite{brooks2023instructpix2pix,zhang2024hive,fu2024guiding,nichol2022glide} & Continuous~\cite{xia2021tedigan} & Gradient penalty regularisation~\cite{el2019tell} & LPIPS loss~\cite{liu2020open,xia2021tedigan} 
\\
Sequential attention~\cite{cheng2020sequential} & Supervised fine-tuning & Multi-tasks~\cite{tao2023net,sheynin2023emu} & Noise combination~\cite{kim2022diffusionclip} & Reconstruction loss~\cite{liu2020open,fan2023target,yue2023chatface} 
\\
Latent inversion~\cite{xia2021tedigan} & Interactive feedback~\cite{jiang2021talk} & Multi-objects~\cite{wang2022manitrans,yang2024objectaware} & CLIP regularisation~\cite{crowson2022vqgan} & Cycle-consistency loss~\cite{liu2020open,fu2020sscr,fan2023target}
\\
Affine combination~\cite{li2020manigan,li2020lightweight} & CLIP guidance~\cite{patashnik2021styleclip,nichol2022glide,kim2022diffusionclip,gal2022stylegan} & Multi-turns~\cite{sheynin2023emu} & Projective augmentation~\cite{avrahami2022blended} & Teacher-forcing loss~\cite{fu2020sscr} 
\\
Cross-modal alignment~\cite{fan2023target} & Weak Supervision~\cite{fan2023target,bodur2023iedit} & Multi-instructions~\cite{guo2023focus} & Detail correction~\cite{li2020manigan} &  CLIP-based losses~\cite{patashnik2021styleclip,kim2022diffusionclip,gal2022stylegan,kocasari2022stylemc,bodur2023iedit,yue2023chatface,wang2022manitrans}
\\
Latent space manipulation~\cite{patashnik2021styleclip} & Few-shot learning~\cite{sheynin2023emu} & Multi-modal~\cite{han2024stylebooth} & Word-level training~\cite{li2020lightweight} & Identity loss~\cite{patashnik2021styleclip,jiang2021talk,kocasari2022stylemc,yue2023chatface}
\\
Semantic field~\cite{jiang2021talk} & Instruction tuning~\cite{huang2023smartedit,gan2024instructcv,li2023moecontroller} & Multi-direction~\cite{dalva2024gantastic} & Conditional augmentation~\cite{zhang2017stackgan,li2020lightweight} & Mask loss~\cite{bodur2023iedit}
\\
Mask conditioning~\cite{couairon2023diffedit,wang2023imagen} & Negative prompting~\cite{tumanyan2023plug} & Mask-free~\cite{lin2024text,couairon2023diffedit} & Human alignment~\cite{geng2023instructdiffusion} & Scale loss~\cite{wang2022manitrans}
\\
Dual-branch concept~\cite{chen2023photoverse} & Tuning-free~\cite{cao2023masactrl,hertz2023prompt,tumanyan2023plug,kawar2023imagic} & Chain-of-thought~\cite{zhang2024tie} & Embedding interpolation~\cite{kawar2023imagic} & Leakage losses~\cite{yang2024dynamic} \\
Cross conditional attention~\cite{guo2023focus} & Pivotal tuning~\cite{roich2022pivotal,mokady2023null} & & Locality/Sparsity regularisation~\cite{roich2022pivotal,bar2022text2live} & Semantic losses~\cite{bar2022text2live} \\
Feature injection~\cite{tumanyan2023plug} & Inversion-free~\cite{xu2024inversion} & & Latent regularisation~\cite{pernuvs2023fice} & Alignment losses~\cite{dalva2024gantastic} \\
Cross-frame attention~\cite{wu2023fairy} & Equivariant finetuning~\cite{wu2023fairy} &  & Localised attention~\cite{li2024zone} & Directional loss~\cite{lin2024text} \\
Layer blending~\cite{li2024zone} & Decoupled classifier-free~\cite{zhang2024effived} & & Scale weighting~\cite{han2024stylebooth} & Structural loss~\cite{lin2024text}\\
Mask extraction/refinement~\cite{shagidanov2024grounded,li2024zone} & Zero-shot learning~\cite{santos2024pix2pix} & & Region generation/location~\cite{lin2024text,wang2024texfit} & Distortion loss~\cite{haque2023instruct,barron2022mip} \\
\bottomrule
\end{tabular}
    \end{threeparttable}
    \end{adjustbox}
\end{table*}

\subsection{GAN-based Controls}

\begin{figure}
    \centering
    \includegraphics[width=.95\linewidth]{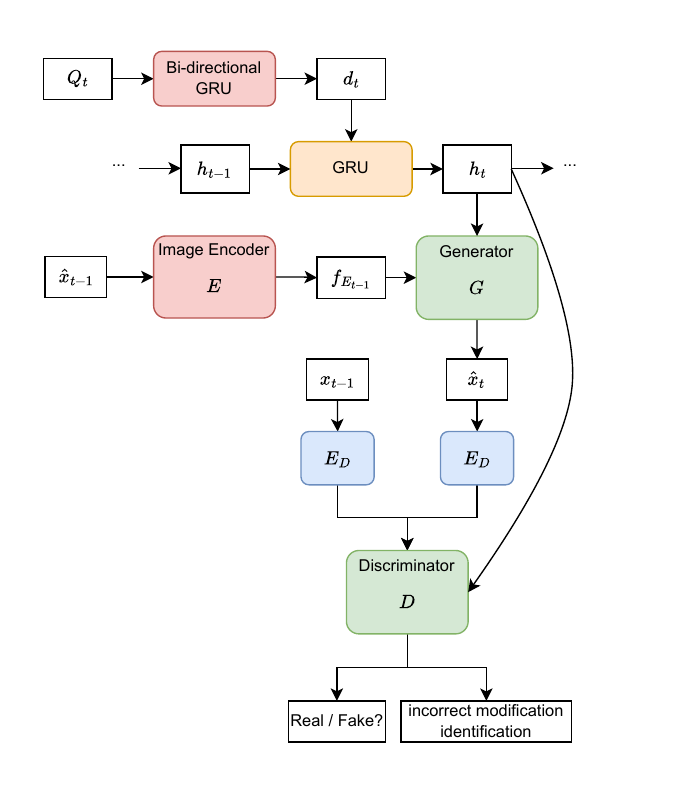}
    \caption{The framework of GAN-based controls.}
    \label{fig:gan-control}
\end{figure}

In \cite{el2019tell}, each instruction \( q_t \) is encoded using a bi-directional Gated Recurrent Unit (GRU) on top of GloVe word embeddings, resulting in the instruction encoding \( d_t \) (see \autoref{fig:gan-control}). The context-aware condition \( h_t \) is computed recursively by another GRU as \( h_t = R(d_t, h_{t-1}) \) and has dimension \( N_c \)~\cite{miyato2018cgans}.
The context-free condition \( f_{Gt-1} \) is the encoder output from the previous image, and the context-aware condition \( h_t \) describes the instruction modifications. These are concatenated with the noise vector \( z_t \) for generating the next image~\cite{zhang2017stackgan}, also utilising conditional batch normalisation \cite{miyato2018cgans}. \( f_{Gt-1} \) is concatenated with feature maps from the generator's intermediate layer \( L_{fG} \) with matching dimensions.
The discriminator \( D \) discriminates not only between real and generated images but also considers if the image is modified incorrectly or unmodified, using an image encoder \( E_D \) \cite{reed2016generative}. The output feature maps \( (K_d \times K_d \times N_d) \) from \( E_D \) are fused and passed through \( D \) \cite{odena2017conditional}.
The adversarial hinge loss is used for training \cite{zhang2019self}, where \( D \) minimises:
 \begin{equation}
 	L_D = L_{D_{real}} + \frac{1}{2}(L_{D_{fake}} + L_{D_{wrong}}) + \beta L_{aux},
 \end{equation}  
 with terms for real, fake, and wrongly instructed images, as well as an auxiliary object detection task. The generator \( G \) minimises:
  \begin{equation}
   L_G = -\mathbb{E}_{z_t \sim N, p_z, c_t \sim p_{data}}[D(G(z_t, \tilde{c_t}), c_t)] + \beta L_{aux}. 
     \end{equation}
  To improve training stability, a zero-centred gradient penalty regularisation \cite{mescheder2018training} is applied to \( D \) with a weighting factor.

\sstitle{Encoding feature manipulation}
Liu et al.~\cite{liu2020open} proposes a method for manipulating image features using textual instructions within a visual-semantic embedding (VSE) space, applying vector arithmetic. For changing attributes, the formula is:
\[ v_{i,j}^{m} = v_{i,j} - \alpha(v_{i,j}, t_1)t_1 + \alpha(v_{i,j}, t_1)t_2 \]
To remove concepts, the formula simplifies to:
\[ v_{i,j}^{m} = v_{i,j} - \alpha(v_{i,j}, t)t \]
For adjusting relative attributes, the method uses:
\[ v_{i,j}^{m} = v_{i,j} + \alpha(v_{i,j}, t)t \]
where \( v_{i,j} \) is the original visual feature at location \( (i,j) \), \( v_{i,j}^{m} \) is the manipulated feature, \( t_1 \) and \( t_2 \) are textual embeddings of source and target attributes respectively, \( t \) is the embedding of the concept to adjust, and \( \alpha \) is a hyper-parameter controlling the manipulation strength through the dot product of \( v_{i,j} \) and \( t \). 

\sstitle{Decoding feature manipulation}
After that, Liu et al.~\cite{liu2020open} develops a method for decoding manipulated feature maps \( V_m \in \mathbb{R}^{1024 \times 7 \times 7} \) into images that preserve structure, using an image decoder trained without paired data. The decoder's training leverages three losses:
(1) Adversarial loss (\( L_G \)) for the generator, and (\( L_D \)) for the discriminator, both based on hinge loss.
(2) Perceptual loss (\( \lambda_{VGG} \)) measuring VGG network feature distance between generated and original images.
(3) Discriminator feature matching loss (\( \lambda_{FM} \)), ensuring similarity in discriminator features between generated and input images.
To preserve image structure, edge constraints are introduced through spatially-adaptive normalization~\cite{he2019bi}, which is guided by an edge map \( \mathcal{E} \) and calculated as:
\begin{equation}
\gamma_{c,h,w}(\mathcal{E}) = \frac{f_{n,c,h,w} - \mu_c}{\sigma_c} + \beta_{c,h,w}(\mathcal{E})	
\end{equation}
Here, \( f_{n,c,h,w} \) is the feature map value at a given batch, channel, and location, \( \mu_c \) and \( \sigma_c \) are the mean and standard deviation of the feature map at channel \( c \), and \( \gamma \) and \( \beta \) predict spatially-adaptive scale and bias parameters. This ensures the decoder produces images that maintain the original's edges and structures.

\sstitle{Cycle-consistency constraint}
Liu et al.~\cite{liu2020open} proposes an additional step for fine-tuning image manipulation that ensures cycle-consistency to maintain image attributes and structure. It introduces two types of losses: cycle-consistency loss \( L_{cyc} \) and reconstruction loss \( L_{rec} \), which are combined with L1 and perceptual losses to optimize the decoder:
\begin{align}
 L_{cyc} = \| I_c - I \|_1 + \lambda \sum \| F_k(I_c) - F_k(I) \|_1 \\
 L_{rec} = \| I_r - I \|_1 + \lambda \sum \| F_k(I_r) - F_k(I) \|_1 
\end{align}
Here, \( \lambda \) is the weight for perceptual loss, \( F_k \) are the VGG features, and \( I \), \( I_m \), \( I_c \), and \( I_r \) represent the original, manipulated, cycled, and reconstructed images, respectively.
To avoid overfitting, perturbations \( P_1, ..., P_{n-1} \) are optimized within the decoder's layers, denoted by \( G_1, ..., G_n \), to address specific image manipulations:
\begin{equation}
	G'(V) = G_n(...(G_1(V) + P_1)...) + P_{n-1}
\end{equation}
These perturbations are regulated with a loss term to prevent extreme deviations from the original image's structure. The method balances manipulating images as instructed while preserving their high-frequency details and overall structure.

\sstitle{Recurrent-Attention}
Chen et al.~\cite{chen2018language} develops a framework for language-based image segmentation and colourization, consisting of a CNN image encoder that converts a \( H \times W \) image into a \( M \times N \) feature map with \( D \)-dimensional vectors (\( V = \{v_i\}_{i=1}^{M \times N}, v_i \in \mathbb{R}^D \)). It includes an LSTM for text encoding, a fusion module for merging image and text features. This module employs attention to iteratively refine a feature map \( S^t \) and generates a fusion map \( O \) using Convolutional Gated Recurrent Units (C-GRUs). It introduces termination gates for each image region to decide when to stop the fusion process probabilistically. For inference, the network performs sequential attention updates and state refinements guided by termination gates. The final image is generated by a deconvolutional image decoder. A discriminator evaluates the realism of generated images against text descriptions, using a mix of GAN and L1 losses for training. For colourization tasks, the framework produces color images from grayscale inputs, guided by natural language expressions and optimized with a GAN loss and an L1 loss between the predicted and actual color channels.

\sstitle{Sequential Attention}
Cheng et al.~\cite{cheng2020sequential} present a generative model called Sequential Attention Generative Adversarial Network (SeqAttnGAN), which creates a series of images (\(x_1, ..., x_T\)) from an initial image (\(x_0\)) and corresponding text descriptions (\(o_1, ..., o_T\)). It encodes \(x_0\) into a feature vector \(v_0\) using ResNet-101~\cite{he2016deep,deng2009imagenet} and processes each \(o_t\) with a bidirectional LSTM to obtain word and sentence feature vectors (\(e_t\) and \(\hat{e}_t\)). The model updates its hidden state \(h_t\) using a GRU~\cite{chung2014empirical}:
$ h_t = GRU(h_{t-1}, e_t)$.
 The Attention Module then combines \(h_{t-1}\) with \(e_t\), generating context-aware image features \(h'_t\), which are used by the Generator to produce the image \(x_t\)~\cite{wang2016generative,zhu2017your}:
 \begin{equation}
 h'_t = F_{attn}(e_t, F(h_{t-1})) 
 \end{equation}
The attention-based word-context vector for each sub-region of the image:
$ c_t^{(i)} = \sum_{j=0}^{L-1} \beta_{i,j}e_t^{(j)}$,
where \(\beta_{i,j}\) is the attention weight calculated as:
$ \beta_{i,j} = \frac{\exp(s_{i,j})}{\sum_{k=0}^{L-1} \exp(s_{i,k})}$.
 Training incorporates losses from the generator (\(L_G\)) and the discriminator (\(L_D\)), as well as an image-text matching loss (\(L_{DAMSM}\)) from the Deep
Attentional Multimodal Similarity model (DAMSM)~\cite{xu2018attngan}, balanced by a hyperparameter \(\lambda\):
 \begin{align}
 	&	  L_G = -\frac{1}{2} \mathbb{E}_{x_t \sim P_G}[\log D(x_t)] - \frac{1}{2} \mathbb{E}_{x_t \sim P_G}[\log(1 - D(x_t, e_t))] \\
&  L_D = -\frac{1}{2} \mathbb{E}_{x_t \sim P_d}[\log D(x_t)] - \frac{1}{2} \mathbb{E}_{x_t \sim P_G}[\log(1 - D(x_t))] \\
 & {L} = {L}_G + \lambda {L}_{DAMSM}
 \end{align}
The goal is to sequentially produce images aligned with the narrative of the text descriptions.

\sstitle{Cross-Task Consistency}
Fu et al.~\cite{fu2020sscr} propose a self-supervised counterfactual reasoning (SSCR) and cross-task consistency (CTC) method for image editing. An iterative editor uses a GAN with a conditional generator \( G \) and a discriminator \( D \)~\cite{el2019tell}. Instructions \( I_t \) are encoded into \( d_t \) by a bidirectional GRU, and the instruction history is captured in \( h_t \) with another GRU~\cite{chung2014empirical,xu2015show}:
$ h_t = GRU(d_t, h_{t-1})$.
The generator \( G \) creates a new image \( V_t \) based on the previous image's features and the instruction history~\cite{miyato2018cgans}:
\begin{equation}
	V_t = G(f_{t-1}, h_t)
\end{equation} 
The discriminator's training involves losses \( L_G \) and \( L_D \) to distinguish real images from generated ones according to the instruction history~\cite{reed2016generative}. Additionally, an iterative explainer \( E \) reconstructs editing instructions to provide explicit token-level training feedback \( L_E \):
\begin{equation}
	L_E = \sum_{i=1}^{L} CE_{loss}(\hat{w}_i, w_i)
\end{equation} 
This framework aims to generate images that match given instructions and enhance learning in a data-scarce environment.

\sstitle{Affine Combination}
Li et al.~\cite{li2020manigan} proposes a novel component called Affine Combination Module (ACM), which is tasked with merging text and image information to manipulate an image semantically according to the text input while preserving parts of the image that are irrelevant to the text~\cite{li2019controllable}. This module uses multiplication and addition operations to fuse hidden feature representations, \( h \in \mathbb{R}^{C \times H \times D} \), from the text with image features, \( v \in \mathbb{R}^{256 \times 17 \times 17} \), extracted from a pretrained Inception-v3 network~\cite{szegedy2016rethinking}. The combined features \( h' \) are obtained by:
\[
 h' = h \odot W(v) + b(v)	
\]
where \( W(v) \) and \( b(v) \) are learned weights and biases based on the image features \( v \), and \( \odot \) denotes the Hadamard element-wise product. The technique aims to precisely alter relevant parts of the image according to the text while maintaining the integrity of parts that the text does not specify~\cite{huang2017arbitrary,park2019semantic}. This approach is preferred over simple concatenation as it allows for more nuanced and accurate image manipulation~\cite{de2017modulating,nam2018text}.

\sstitle{Detail Correction}
Li et al.~\cite{li2020manigan} proposes a novel component called Detail Correction Module (DCM), is designed to refine image details and fill in missing content after initial manipulation by the Affine Combination Module (ACM). It uses three inputs: the last hidden features from ACM, word-level features from an RNN~\cite{xu2018attngan}, and visual features from a VGG-16 network~\cite{simonyan2014very}. By applying spatial and channel-wise attention mechanisms~\cite{li2019controllable}, the DCM enhances the alignment of fine-grained textual information with the image's intermediate feature maps. This process results in a final image that is a more accurate and detailed representation of the text description. The DCM helps ensure that the manipulated image retains high-quality visual attributes while also incorporating any necessary details that the initial processing steps may have missed.

\sstitle{Latent Inversion}
Xia et al.~\cite{xia2021tedigan} propose the TediGAN framework for training an image encoder to map real images to the latent space, ensuring alignment at both the pixel-level and the semantic-level. The framework uses the hierarchical characteristic of \( \mathcal{W} \) space to align text and image embeddings. It maintains identity during manipulation through instance-level optimization and regularization.
Key differences from conventional GAN inversion~\cite{zhu2020domain} include training with real images instead of synthesized images, and focusing on image space rather than latent space~\cite{bau2019seeing}. The training process involves minimizing the following objective functions~\cite{johnson2016perceptual,zhang2018unreasonable}:
\begin{multline}
\min_{\Theta_{E_v}} \mathcal{L}_{E_v} = \left\| x - G\left(E_v(x)\right)\right\|_2^2 + \lambda_1 \left\| F(x) - F\left(G\left(E_v(x)\right)\right)\right\|_2^2 \\ - \lambda_2 \mathbb{E}\left[D_v\left(G\left(E_v(x)\right)\right)\right]
\end{multline}
\begin{equation}
\min_{\Theta_{D_v}} \mathcal{L}_{D_v} = \mathbb{E}\left[D_v\left(G\left(E_v(x)\right)\right)\right] - \mathbb{E}\left[D_v(x)\right] + \frac{\lambda_3}{2}\mathbb{E}\left[\left\|\nabla_x D_v(x)\right\|_2^2\right]
\end{equation}
where \( E_v(\cdot) \) represents the visual encoder, and \( D_v(\cdot) \) is the discriminator. The parameters \( \Theta_{E_v} \) and \( \Theta_{D_v} \) are learnable, with \(\lambda_1\), \(\lambda_2\), and \(\lambda_3\) being hyperparameters. \(F(\cdot)\) denotes a VGG feature extraction model.
The inversion module aims to align the semantic domain of the StyleGAN generator~\cite{karras2019style} and utilizes cross-modal similarity between text and image pairs for training. The inversion process focuses on reconstructing meaningful and interpretable latent codes~\cite{shen2020interpreting,yang2021semantic}.

\sstitle{Semantic Field}
Jiang et al.~\cite{jiang2021talk} propose the use of a pre-trained GAN generator \(G\) to manipulate the attribute degrees of a facial image~\cite{shen2020interpreting,pan2021do}. Given an input image \(I \in \mathbb{R}^{3 \times H \times W}\) and a latent code \(z \in \mathbb{R}^d\), the goal is to find a vector \(f_z \in \mathbb{R}^d\) that can change the attribute degree~\cite{shen2020interfacegan,zhuang2021enjoy}. The paper introduces the concept of a ``semantic field'' \(F\), which is a vector field that assigns a vector to each latent code, defined as \(F = \nabla S\), where \(S\) is the attribute score. The direction of the semantic field vector \(f_z\) represents the direction in which the attribute score \(s\) increases the fastest. 
To shift the latent code \(z\) by a certain amount, Jiang et al.~\cite{jiang2021talk} suggest an equation \(z' = z + \alpha f_z = z + \alpha F(z)\), where \(\alpha\) is the step size, set to 1. The equation \(\int_{z_a}^{z_b} f_z \cdot dz = s_b - s_a\) describes how the attribute score can change along a semantic field line from \(z_a\) to \(z_b\).
The system discretizes the semantic field by approximating the continuous field into a discrete version using the equation \(s_a + \sum_{i=1}^N f_{z_i} \cdot \Delta z_i = s_b\). The system uses a mapping network \(F\) to implement the semantic field and trains a pre-trained fine-grained attribute predictor \(P\) to supervise its learning. The semantic field learning uses three loss functions: \(L_{pred}\), \(L_{id}\), and \(L_{disc}\). The identity keeping loss \(L_{id}\) ensures that face identity remains consistent when shifting latent codes along the semantic field, while the discriminator loss \(L_{disc}\) helps prevent unrealistic artifacts in edited images. The final loss function for learning the semantic field is \(L_{total} = \lambda_{pred} L_{pred} + \lambda_{id} L_{id} + \lambda_{disc} L_{disc}\).

\sstitle{Pivotal tuning}
Roich et al.~\cite{roich2022pivotal} leverage StyleGAN's ability to make slight, local changes without damaging the overall appearance.
The process begins with inversion~\cite{karras2020analyzing}, where the input image \( \mathbf{x} \) is converted to a latent code \( \mathbf{w}_p \) in StyleGAN's latent space \( \mathcal{W} \). This involves optimizing both the latent code \( \mathbf{w} \) and the noise vector \( \mathbf{n} \) using the LPIPS perceptual loss~\cite{zhang2018unreasonable}:
\[
\mathbf{w}_p, \mathbf{n} = \arg \min_{\mathbf{w}, \mathbf{n}} \mathcal{L}_{\text{LPIPS}}(\mathbf{x}, G(\mathbf{w}, \mathbf{n}; \theta)) + \lambda_n \mathcal{L}_n(\mathbf{n}),
\]
where \( G(\mathbf{w}, \mathbf{n}; \theta) \) is the generated image.
Next, Pivotal Tuning fine-tunes the generator to better reconstruct the input image using the latent code \( \mathbf{w}_p \). The tuning uses the loss function:
\[
\mathcal{L}_{\text{PT}} = \mathcal{L}_{\text{LPIPS}}(\mathbf{x}, \mathbf{x}^p) + \lambda_{L2} \mathcal{L}_{L2}(\mathbf{x}, \mathbf{x}^p),
\]
where \( \mathbf{x}^p = G(\mathbf{w}_p; \theta^*) \).
To maintain local edits, the method introduces locality regularisation~\cite{shen2020interpreting,harkonen2020ganspace}. This involves interpolating between a random vector \( \mathbf{w}_z \) and the pivot code \( \mathbf{w}_p \):
\[
\mathbf{w}_r = \mathbf{w}_p + \alpha \frac{\mathbf{w}_z - \mathbf{w}_p}{\|\mathbf{w}_z - \mathbf{w}_p\|_2},
\]
and optimising the distance between images generated with the original and tuned weights:
\[
\mathcal{L}_{\text{R}} = \mathcal{L}_{\text{LPIPS}}(\mathbf{x}_r, \mathbf{x}^*_r) + \lambda_{L2} \mathcal{L}_{L2}(\mathbf{x}_r, \mathbf{x}^*_r).
\]
The final optimization objective is:
\[
\theta^* = \arg \min_{\theta^*} \mathcal{L}_{\text{PT}} + \lambda_R \mathcal{L}_{\text{R}}.
\]

\sstitle{Light-weight GAN}
Li et al.~\cite{li2020lightweight} design a light-weight version of GAN for instruction-based image editing. First, the discriminator's objective includes both unconditional and conditional adversarial losses to distinguish real images from fake ones and to align images with their text descriptions~\cite{nam2018text}. To ensure the model remains lightweight and efficient, a word-level discriminator is introduced. Each word in the description is labeled with specific targets, such as objects or attributes, and the discriminator learns to associate each word with the corresponding region in the image~\cite{li2019controllable}. This method leverages word-level feedback to help the discriminator retain the semantic meaning of individual words, ensuring the generated images accurately reflect the textual description~\cite{xu2018attngan}. Finally, the generator comprises a text encoder, image encoders, and an image generator. It modifies the input image based on the text description, using encoded features from pre-trained networks. This ensures efficient and accurate image manipulation, guided by both adversarial and perceptual losses.

\sstitle{Multi-task}
DE-Net~\cite{tao2023net} introduces a novel GAN model with three key components. First, the Dynamic Editing Block (DEBlock) adaptively manipulates image features by dynamically combining spatial and channel-wise editing techniques~\cite{wang2022manitrans}. It employs C-Affine and S-Affine transformations, 
respectively. These transformations are integrated through a weighted sum,
allowing dynamic feature manipulation.
Second, the Composition Predictor (Comp-Pred) predicts combination weights for feature manipulation based on text and visual inputs. It maps image features to visual vectors and concatenates them with text vectors, enhancing the alignment between text and visual information to ensure accurate image editing.
Third, the Dynamic Text-adaptive Convolution Block (DCBlock) improves editing precision by dynamically adjusting convolution kernel parameters based on text guidance. 
This dynamic adjustment allows the DCBlock to distinguish between text-relevant and text-irrelevant parts of the image.
Inspired by the multi-head attention mechanism in Transformers, DE-Net uses multiple dynamic convolution kernels as attention heads, enabling joint attention to various image features from different text representation subspaces. These kernels predict attended feature maps, which are combined to form a comprehensive attended feature map, used to predict spatial affine parameters for the DEBlock.

\subsection{Diffusion-based Controls}

\begin{figure}
    \centering
    \includegraphics[width=\linewidth]{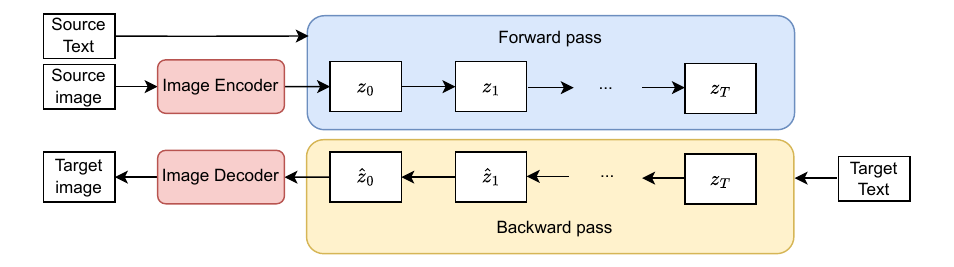}
    \caption{Latent-diffusion-based control.}
    \label{fig:diffusion-control}
\end{figure}

\sstitle{Latent diffusion}
Chandramouli et al.~\cite{chandramouli2022ldedit} propose the adaptation of Latent Diffusion Models (LDMs) for image manipulation using a shared latent representation between source and target images. As shown in \autoref{fig:diffusion-control}, the source image \( x_{\text{src}} \) is encoded into a latent code \( z_0 \) by the encoder \( \mathcal{E} \), and forward diffusion is performed using DDIM sampling, conditioned on the source text \( y_{\text{src}} \):
\begin{multline*}
z_{t+1} = \sqrt{\alpha_{t+1}} \left( \frac{z_t - \sqrt{1 - \alpha_t} \epsilon_{\theta}(z_t, t, \tau_{\theta}(y_{\text{src}}))}{\sqrt{\alpha_t}} \right) \\ + \sqrt{1 - \alpha_{t+1}} \epsilon_{\theta}(z_t, t, \tau_{\theta}(y_{\text{src}}))
\end{multline*}
The reverse diffusion, conditioned on the target text \( y_{\text{tar}} \), starts from the same latent code \( z_{t_{\text{stop}}} \):
\begin{multline*}
z_{t-1} = \sqrt{\alpha_{t-1}} \left( \frac{z_t - \sqrt{1 - \alpha_t} \epsilon_{\theta}(z_t, t, \tau_{\theta}(y_{\text{tar}}))}{\sqrt{\alpha_t}} \right) \\ + \sqrt{1 - \alpha_{t-1}} \epsilon_{\theta}(y_t, t, \tau_{\theta}(y_{\text{tar}}))
\end{multline*}
The deterministic sampling ensures cycle-consistency between images. This method supports various image manipulations using pretrained LDMs without additional training.

\sstitle{Denoising diffusion}
Brooks et al.~\cite{brooks2023instructpix2pix} proposes InstructPix2Pix, a model trained on generated data for instruction-based image editing. Many other works share a similar process such as UniTune~\cite{valevski2023unitune}, Prompt-to-Prompt~\cite{hertz2023prompt}, DiffusionCLIP~\cite{kim2022diffusionclip}, and GLIDE~\cite{nichol2022glide}. The model applies classifier-free guidance~\cite{ho2022classifier} on Stable Diffusion~\cite{rombach2022high} with two conditionings: input image \( c_I \) and text instruction \( c_T \). It operates in a latent space created by a variational autoencoder (VAE) and optimizes a latent diffusion objective:
\begin{equation}
	L = \mathbb{E}_{\varepsilon \sim \mathcal{N}(0,1), t} \left[ \| \varepsilon - \theta(\mathbf{z}_t, t, \mathcal{E}(c_I), c_T) \|_2^2 \right]
\end{equation} 
It employs classifier-free guidance~\cite{ho2022classifier} to shift the generated sample's probability distribution, using the following modified score estimate for conditional and unconditional denoising using \autoref{eq:diffusion_guidance}.
During training, the model randomly applies unconditional denoising to some examples to become adept at handling both conditional or unconditional inputs~\cite{liu2022compositional}. The final score estimate is adjusted with guidance scales \( s_I \) and \( s_T \) to balance between fidelity to the input image and intensity of the edit as per the instruction:
\begin{multline}
\tilde{\theta}(\mathbf{z}_t, c_I, c_T) = \theta(\mathbf{z}_t, \emptyset, \emptyset) + s_I \cdot (\theta(\mathbf{z}_t, c_I, \emptyset) - \theta(\mathbf{z}_t, \emptyset, \emptyset)) \\ + s_T \cdot (\theta(\mathbf{z}_t, c_I, c_T) - \theta(\mathbf{z}_t, c_I, \emptyset))	
\end{multline}

The model learns the conditional probability distribution \( P(\mathbf{z}|c_I, c_T) \) which is decomposed as follows:
\begin{equation}
	P(\mathbf{z}|c_T, c_I) = \frac{P(c_T|c_I, \mathbf{z})P(c_I|\mathbf{z})P(\mathbf{z})}{P(c_T, c_I)}
\end{equation} 
By taking the log and the derivative, they relate the terms to the gradient of the log probability of the model's output:
\begin{equation}
	\nabla_{\mathbf{z}} \log(P(\mathbf{z}|c_T, c_I)) = \nabla_{\mathbf{z}} \log(P(\mathbf{z})) + \nabla_{\mathbf{z}} \log(P(c_I|\mathbf{z})) + \nabla_{\mathbf{z}} \log(P(c_T|c_I, \mathbf{z}))
\end{equation} 
This corresponds to the terms in the classifier-free guidance formulation~\cite{ho2022classifier} and allows the model to adjust the influence of each conditioning on the generated sample. The model is capable of learning these implicit classifiers by differentiating between estimates with and without the given conditional input. This method improves how well the generated samples match the conditional inputs, providing a more directed approach to text-conditioned image editing tasks.

Fu et al.~\cite{fu2024guiding} uses a sequence-to-sequence model \( T \)~\cite{li2023blip,li2024blip} that incorporates an edit head to translate visual tokens $[IMG]$ into a semantically meaningful latent space \( U \):
\[ u_t = T(\{u_1, ..., u_{t-1}\} | \{e_{\text{[IMG]}} + h_{\text{[IMG]}}\}), \]
where \( e \) is the word embedding and \( h \) is the hidden state associated with the image tokens. The method leverages a latent diffusion model \( \mathcal{F} \) for visual guidance~\cite{rombach2022high}, which includes a variational autoencoder (VAE) that generates a latent goal \( o = Enc_{VAE}(\mathcal{O}) \). The model uses denoising diffusion by adding noise \( z_t \) over timesteps \( t \) and employs a UNet architecture to predict the added noise~\cite{ho2020denoising}.
The editing loss \( \mathcal{L}_{edit} \) is computed using the score estimation \( s_{\theta}(z_t, \mathcal{O}, \{u\}) \) from classifier-free guidance~\cite{ho2022classifier}, formulated as:
\begin{equation}
 \mathcal{L}_{edit} = \mathbb{E}_{o,\{u\},\varepsilon \sim \mathcal{N}(0,1)_t}[\| \varepsilon - \theta(z_t, v, \{u\}) \|_2^2 ],	
\end{equation}
where \( \alpha_y \) and \( \alpha_x \) are the weights for the guidance scale for the image and the instruction. During training, these scales are adjusted, and at inference \( \alpha_y = 1.5 \) and \( \alpha_x = 7.5 \).
Overall optimization is performed on the combined loss:
\begin{equation}
	 \mathcal{L}_{all} = \mathcal{L}_{ins} + 0.5 \cdot \mathcal{L}_{edit},
\end{equation}
where \( \mathcal{L}_{ins} \) is the instruction loss in \autoref{eq:instruction_loss}.

\sstitle{Conditional Diffusion}
Nichol et al.~\cite{nichol2022glide} train a 3.5 billion parameter text-conditioned diffusion model at \(64 \times 64\) resolution and a smaller 1.5 billion parameter model for upsampling to \(256 \times 256\)~\cite{dosovitskiy2021an}. They also train a noised CLIP model for guidance.
First, they utilize the ablated diffusion model (ADM) architecture~\cite{dhariwal2021diffusion}, augmented with text conditioning, modeling \(p(x_{t-1} \mid x_t, c)\). Second, they encode the text into tokens, which are passed into a Transformer, and use the output embeddings as class embeddings or attention context~\cite{ramesh2021zero}.
Third, they fine-tune the model for unconditional image generation by replacing 20\% of text sequences with empty sequences during training, enabling text-conditioned and unconditional outputs. Next, they also fine-tune for inpainting by masking image regions and conditioning on masked versions, adding RGB and mask channels for the ground-truth image and inpainting mask~\cite{meng2022sdedit,saharia2022palette}. For classifier guidance, they train noised CLIP models using an image encoder \(f(x_t, t)\) for noised images \(x_t\) at \(64 \times 64\) resolution, maintaining the same noise schedule as their base model~\cite{dhariwal2021diffusion}.

\sstitle{Contextual Diffusion}
Yang et al.~\cite{yang2023imagebrush} propose a method for context learning by progressive denoising. Given the denoised result \( x_{t-1} = \text{Grid}(\{E, E', I, I'\}_{t-1}) \) from the previous time step \( t \), the objective is to refine this grid-like image using contextual descriptions. The model uses a pre-trained variational autoencoder, reducing computational resources and enhancing image quality. Specifically, for an image \( x_t \), the diffusion process removes Gaussian noise from its encoded latent input \( z_t = \mathcal{E}(x_t) \), with the final image \( x_0 = \mathcal{D}(z_0) \) decoded from the latent variable \( z_0 \)~\cite{rombach2022high}.

\begin{figure}
    \centering
    \includegraphics[width=\linewidth]{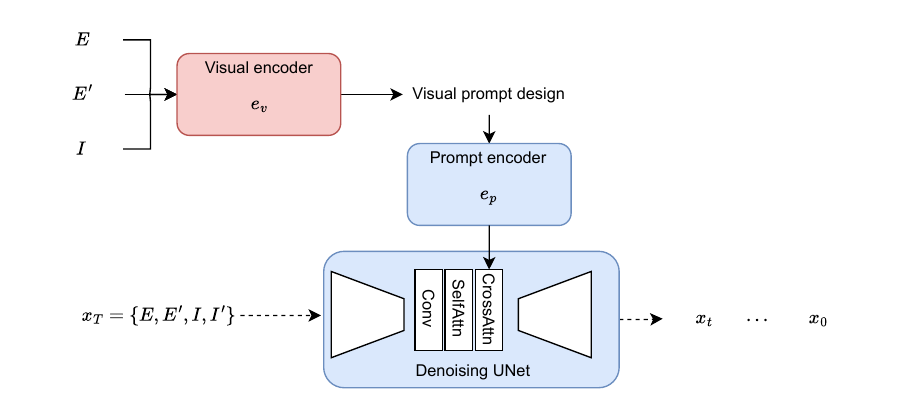}
    \caption{Visual prompt design using contextual diffusion.}
    \label{fig:contextual-diffusion}
\end{figure}

The model employs a UNet-like network with self-attention blocks to capture interdependencies within image channels, focusing on high-level semantic relationships using visual prompts \( vp = \{E, E', I\} \), as shown in \autoref{fig:contextual-diffusion}. This involves a shared visual encoder \( e_v \) and a prompt encoder \( e_p \), extracting tokenized features \( f_{\text{img}} \) from an image \( I \), which are processed by a bi-directional transformer \( e_p \).

High-level semantic information~\cite{li2023gligen,brooks2023instructpix2pix,voynov2023p} is integrated into the UNet architecture using classifier-free guidance, injecting contextual features \( f_c \) through cross-attention at specific layers:
\[ \phi^{l-1} = \phi^{l-1} + \text{Conv}(\phi^{l-1}) \]
\[ \phi^{l-1} = \phi^{l-1} + \text{SelfAttn}(\phi^{l-1}) \]
\[ \phi^l = \phi^{l-1} + \text{CrossAttn}(f_c) \]
For enhancing user intent comprehension, the interface module allows users to designate areas of interest via bounding boxes or automated tools, processed by a box encoder \( e_b \)~\cite{kirillov2023segment}. GroundingDINO~\cite{liu2023grounding} labels the focused region based on textual instructions, reducing manual labeling and enhancing flexibility in user interactions~\cite{li2023gligen}.

Chen et al.~\cite{chen2023photoverse} integrate dual-branch conditioning in both textual and visual domains. This includes projecting the reference image into pseudo-words and image features for accurate concept representation, which are then injected into the text-to-image model.
Stable Diffusion, based on Latent Diffusion Model (LDM) architecture, is utilised~\cite{rombach2022high}. It includes an autoencoder to compress images and a denoising network to perform the diffusion process. The model uses a CLIP text encoder to project conditions into intermediate latent representations, employed in UNet via cross-attention.
The training objective is to predict the noise added to the latent image.
For dual-branch concept injection, only cross-attention module weights are fine-tuned~\cite{hu2022lora,kumari2023multi,gal2023encoder}. 
Visual conditions are integrated similarly. The multimodal fusion is a weighted sum:
\[
O = \gamma \text{Attn}(Q, K^T, V^T) + \sigma \text{Attn}(Q, K^S, V^S),
\]
with \(\gamma\) and \(\sigma\) as scale factors, and a random seed determining the combination~\cite{deng2019arcface}. Regularization terms for token embedding and image values are:
\[
\mathcal{L}^T_{\text{reg}} = \text{Mean}[\|p_f\|_1] \quad \text{and} \quad \mathcal{L}^S_{\text{reg}} = \text{Mean}[\|V^S\|_1].
\]
The total loss combines three losses, i.e. diffusion, facial identity, and regularization:
\[
\mathcal{L}_{\text{total}} = \mathcal{L}_{\text{diffusion}} + \lambda_{\text{face}} \mathcal{L}_{\text{face}} + \lambda_{\text{rt}} \mathcal{L}^T_{\text{reg}} + \lambda_{\text{rv}} \mathcal{L}^S_{\text{reg}}.
\]
Lightweight adapters and UNet are trained to enable fast personalization based on user-provided prompts~\cite{karras2019style}.

\sstitle{Mask Conditioning (Cascaded Diffusion)}
Conventional approaches may inadvertently alter irrelevant parts of an image when editing it based on a query. To resolve this, DiffEdit~\cite{couairon2023diffedit} utilizes a text-conditioned diffusion model, which focuses on modifying only relevant areas while preserving other parts.
The authors start by denoising an image with a text-conditioned diffusion model, which produces varying noise estimates based on different text conditionings. For example, comparing a `zebra' query to a `horse' reference helps identify which parts need editing. Using Gaussian noise at 50\% strength, they stabilize differences by averaging spatial variations across ten input noises. The mask is then rescaled to [0, 1] and binarized with a 0.5 threshold, highlighting the regions needing attention. Then, the authors encode the input image \(x_0\) into a latent space at timestep \(r\) using the DDIM~\cite{song2021denoising} encoding function \(E_r\). This step uses an unconditional model, without any specific text conditioning~\cite{meng2022sdedit}. After obtaining the latent representation \(x_r\), they decode it using a diffusion model conditioned on the editing text query \(Q\), such as `zebra.' They guide the diffusion process with the mask \(M\), combining masked areas and the original image as follows:
\[
   \tilde{y}_t = M y_t + (1 - M)x_t,
\]
   where \(y_t\) is computed from the previous timestep. The encoding ratio \(r\) controls the edit strength, balancing accuracy and deviation from the original image. 

\begin{figure*}
    \centering
    \includegraphics[width=0.9\linewidth]{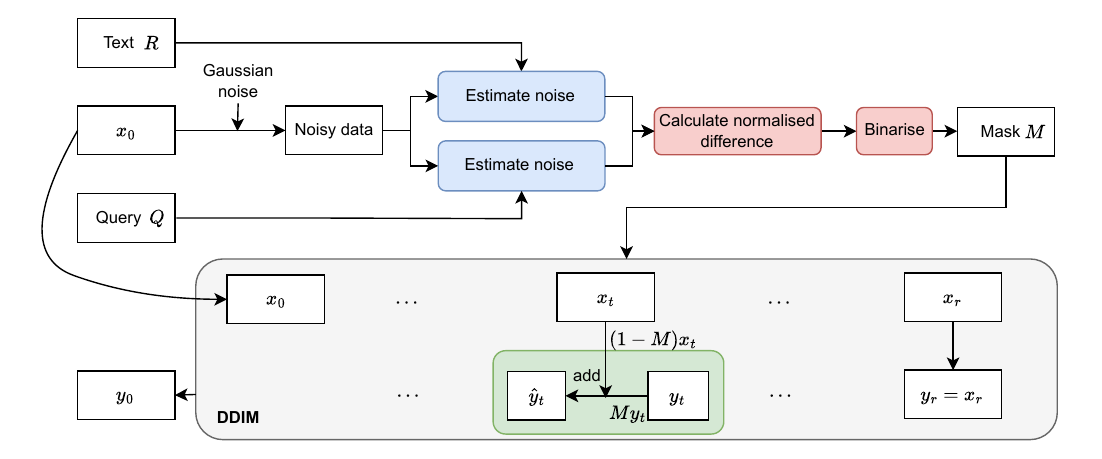}
    \caption{The pipeline of conditioning a denoising diffusion implicit model (DDIM) with a mask.}
    \label{fig:mask-condition}
\end{figure*}

Imagen Editor~\cite{wang2023imagen} is another mask-conditioned image inpainting model. It operates by taking three primary inputs from the user: the image to be edited, a binary mask specifying the edit region, and a text prompt, which collectively guide the output samples. To address the challenge of mask relevance, the model employs an object detector (e.g. SSD Mobilenet v2~\cite{howard2018inverted}) to create masks that are well-aligned with the text prompt~\cite{yu2018generative,yu2019free,ding2022cogview2}. For high-resolution editing, Imagen Editor conditions on both the image and the mask by concatenating them with diffusion latents along the channel dimension, akin to methods like SR3~\cite{saharia2022image}, Palette \cite{saharia2022palette}, and GLIDE~\cite{nichol2022glide}, ensuring the conditioning image and mask are always at \( 1024 \times 1024 \) resolution. The base diffusion model operates at smaller resolutions (e.g., \( 64 \times 64 \) or \( 256 \times 256 \)), necessitating downsampling. A parameterised downsampling convolution is preferred over simple bicubic downsampling to avoid significant artefacts along mask boundaries and enhance fidelity. The new input channel weights are initialised to zero~\cite{nichol2022glide}, ensuring that the model starts identical to Imagen~\cite{saharia2022photorealistic}, focusing on high-quality inpainting results by effectively utilising the conditioning image and mask inputs.

Wang et al.~\cite{wang2023instructedit} also minimise noise prediction errors using DDIM~\cite{song2021denoising} for sample generation. They use ChatGPT~\cite{chang2024survey} to extract segmentation prompts and captions through in-context learning, assisted by BLIP2 for image content understanding.
Grounded Segment Anything combines Grounding DINO~\cite{liu2023grounding} and Segment Anything~\cite{kirillov2023segment} to identify and mask objects. Grounding DINO finds object bounding boxes, and Segment Anything refines these into binary masks.
Mask-guided image editing uses these masks to direct the denoising process, ensuring edits are applied only within the masked regions while preserving the rest of the image~\cite{couairon2022diffedit}.

\begin{figure}
    \centering
    \includegraphics[width=\linewidth]{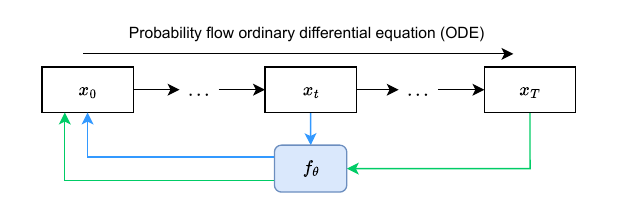}
    \caption{Consistent diffusion.}
    \label{fig:consistent-diffusion}
\end{figure}

\sstitle{Consistent Diffusion}
Consistency models (CMs)~\cite{song2023consistency} have been introduced to accelerate the generation process compared to previous diffusion models (DMs). A key feature of CMs is self-consistency, ensuring samples along a trajectory map to the initial sample. This is achieved using a consistency function \( f(\mathbf{z}_t, t) \), optimized as:
\begin{equation}
\min_{\theta, \theta^-; \phi} \mathbb{E}_{\mathbf{z}_0, t} \left[ d \left( f_\theta(\mathbf{z}_{t_{n+1}}, t_{n+1}), f_{\theta^-} (\hat{\mathbf{z}}^\phi_{t_n}, t_n) \right) \right]
\end{equation}
Here, \( f_\theta \) is a trainable neural network (see \autoref{fig:consistent-diffusion}), while \( f_{\theta^-} \) is a slowly updated target model. The variable \(\hat{\mathbf{z}}^\phi_{t_n}\) denotes a one-step estimation of \(\mathbf{z}_{t_n}\).
Sampling in CMs involves a sequence of timesteps \(\tau_{1:n} \in [t_0, T]\), starting from initial noise \(\hat{\mathbf{z}}_T\) and iteratively updating the process using:
\[
\hat{\mathbf{z}}_{\tau_i} = \mathbf{z}_0^{(\tau_i+1)} + \sqrt{\tau_i^2 - t_0^2} \mathbf{\epsilon}
\]
\[
\mathbf{z}_0^{(\tau_i)} = f_\theta (\hat{\mathbf{z}}_{\tau_i}, \tau_i)
\]
Latent Consistency Models (LCMs)~\cite{luo2023latent} extend this to include text conditions \( c \) for text-guided image manipulation. Sampling in LCMs adjusts the process to incorporate \( c \):
\[
\hat{\mathbf{z}}_{\tau_i} = \sqrt{\alpha_{\tau_i}} \mathbf{z}_0^{(\tau_i+1)} + \sigma_{\tau_i} \mathbf{\epsilon}
\]
\[
\mathbf{z}_0^{(\tau_i)} = f_\theta (\hat{\mathbf{z}}_{\tau_i}, \tau_i, c)
\]

Denoising Diffusion Consistent Models (DDCM)~\cite{xu2024inversion} offer a non-Markovian forward process where \(\mathbf{z}_t\) directly maps to the ground truth \(\mathbf{z}_0\). This ensures exact consistency between original and reconstructed images.
DDCM enables Virtual Inversion, an efficient method starting from random noise and supporting multistep consistency sampling~\cite{xu2024inversion}. This method ensures exact consistency without requiring explicit inversion, maintaining \(\mathbf{z} = \mathbf{z}_0\) throughout the process.

\sstitle{Controllable Diffusion}
Nguyen et al.~\cite{nguyen2024flexedit} propose FlexEdit, which starts by encoding an input image \( I \) into a latent representation \( z_0 \) using DDIM Inversion~\cite{song2021denoising}. It then denoises through intermediate source latents \( \{z_t\}_{t=0}^T \), crucial for real and synthesised image editing, eventually decoding \( z_0^* \) to generate the edited image \( I^* \).
The framework involves two main stages: a forwarding stage to collect noisy latents and an editing stage to manipulate them. Dynamic object binary masks are extracted and refined using cross-attention and self-attention maps~\cite{hertz2023prompt}, converted into binary masks \( M_{j,t} \).
FlexEdit integrates latent blending and optimisation at each denoising step~\cite{nguyen2024dataset}. Latent optimisation aligns the noisy latent code with user constraints, adjusting for object size and position using centroid \( \text{centroid}_{j,t} \) and size \( \text{size}_{j,t} \) calculations~\cite{epstein2023diffusion}.
Attention separation for object addition is addressed using a separation loss \( L_{\text{sep}} \) based on cosine similarity, ensuring the new object doesn't interfere with existing ones~\cite{agarwal2023star,li2023divide}. Latent blending employs an adaptive binary mask, combining source and target tokens for flexibility and background preservation~\cite{ren2024grounded}.

\sstitle{Blending}
Avrahami et al.~\cite{avrahami2022blended} propose a method for editing specific regions of an image defined by a text description, using a binary mask to isolate the area of interest. This approach uses a local CLIP-guided diffusion process, adapted from DDPM~\cite{dhariwal2021diffusion}. The core technique estimates a ``clean'' image from a noisy one using:
\[
\hat{x}_0 = \frac{x_t - \sqrt{1 - \hat{\alpha}_t \epsilon_\theta(x_t, t)}}{\sqrt{\hat{\alpha}_t}}
\]
and optimizes a CLIP-based loss that minimizes the cosine distance between the CLIP embeddings of the text prompt and the image, focusing the transformation on the masked region.
To improve background preservation and ensure seamless integration between edited and unedited areas, the method introduces a new blended diffusion technique. This involves blending noisy image versions at various diffusion stages, using a combination of the mean squared error (MSE) and learned perceptual image patch similarity (LPIPS) metrics~\cite{zhang2018unreasonable} to evaluate the background preservation:
\begin{equation}
D_{bg}(x_1, x_2, m) = d(x_1 \odot (1-m), x_2 \odot (1-m))	
\end{equation}
\begin{equation}
d(x_1, x_2) = \frac{1}{2} (\text{MSE}(x_1, x_2) + \text{LPIPS}(x_1, x_2))
\end{equation}
Additionally, to prevent adversarial results and enhance robustness, extended projective augmentations are applied to the diffusion process. This strategy forces the model to maintain high fidelity to the text description across different transformations, improving the coherence of the final edited image~\cite{ramesh2021zero,razavi2019generating}.

\begin{figure*}
    \centering
    \includegraphics[width=0.9\linewidth]{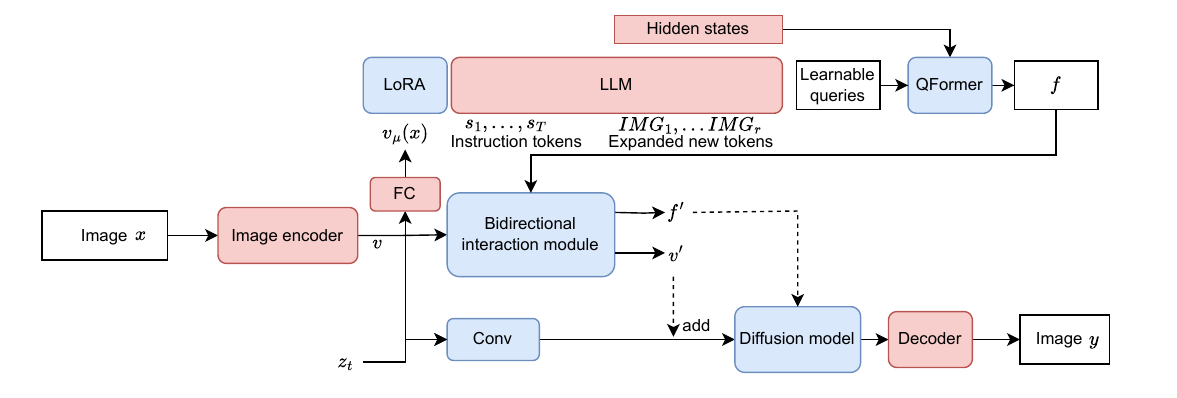}
    \caption{The framework of SmartEdit~\cite{huang2023smartedit}.}
    \label{fig:bidirectioninteract}
\end{figure*}

\sstitle{Bidirection Interaction}
SmartEdit~\cite{huang2023smartedit} is designed to handle complex understanding and reasoning scenarios, a focus not extensively covered by MGIE~\cite{fu2024guiding}. It introduces a Bidirectional Interaction Module (BIM) that fosters extensive two-way interactions between images and language model outputs, improving upon MGIE's approach (see~\autoref{fig:bidirectioninteract}). In contrast to InstructDiffusion~\cite{geng2023instructdiffusion}, which offers a generic framework for aligning vision tasks with human instructions, SmartEdit specializes in instruction-based image editing. It utilises the pretrained weights from InstructDiffusion for its diffusion model, noting that while InstructDiffusion performs adequately in general perception tasks, it falls short in complex scenarios. SmartEdit addresses these gaps by integrating its BIM with the LLaVA language model~\cite{liu2024visual,zhu2024minigpt} and augmenting its training data, leading to better outcomes in sophisticated understanding and reasoning tasks.

\sstitle{Embedding Interpolation}
Kawar et al.~\cite{kawar2023imagic} propose a process involving three main stages.
First, the target text is encoded into a text embedding \( e_{\text{tgt}} \)~\cite{raffel2020exploring}, which is then optimised using a denoising diffusion~\cite{ho2020denoising}.
This optimization produces an embedding \( e_{\text{opt}} \) that closely matches the input image.
Second, model fine-tuning is performed. The optimised embedding \( e_{\text{opt}} \) is used to adjust the model parameters \( \theta \) with the same loss function, ensuring the generative model accurately reconstructs the input image while adhering to the target text.
Third, text embedding interpolation is used to apply the desired edit. A linear interpolation between \( e_{\text{tgt}} \) and \( e_{\text{opt}} \) is performed:
\[
\bar{e} = \eta \cdot e_{\text{tgt}} + (1 - \eta) \cdot e_{\text{opt}}
\]
where \( \eta \in [0, 1] \). This interpolated embedding \( \bar{e} \) guides the generation of a low-resolution edited image, which is then super-resolved using fine-tuned auxiliary models, resulting in the final high-resolution edited image \( \bar{\mathbf{x}} \).

\sstitle{Location Awareness}
Bodur et al.~\cite{bodur2023iedit} incorporate masks for localized image editing to better align source and pseudo-target images. Masks guide the learning process during training and optionally during inference to generate localized edits. CLIPSeg~\cite{luddecke2022image} is used to generate these masks.
During training, the method incorporates masks to focus on editing specific image regions. The optimization process is modified to predict target noise \( \epsilon_2 \) on the masked region and source noise \( \epsilon_1 \) on the inverse mask region, resulting in the loss function:
\begin{equation}
\mathcal{L}_{\text{mask}} = \mathbb{E}_{\mathcal{E}(x), y_2, \epsilon_2, t} \mathcal{L}_{\text{mask}}^{\text{fg}} + \mathbb{E}_{\mathcal{E}(x), y_2, \epsilon_1, t} \mathcal{L}_{\text{mask}}^{\text{bg}},
\end{equation}
where the foreground and background losses are:
\begin{equation*}
\begin{split}
\mathcal{L}_{\text{mask}}^{\text{fg}} = \left\| [\epsilon_2 \odot M_2 - \epsilon_\theta(z_t, t, \tau_\theta(y_2))] \odot M_2 \right\|^2, \\
\mathcal{L}_{\text{mask}}^{\text{bg}} = \left\| [\epsilon_1 \odot \overline{M_1} - \epsilon_\theta(z_t, t, \tau_\theta(y_2))] \odot \overline{M_1} \right\|^2,
\end{split}
\end{equation*}
where \(\overline{M_1}\) represents the inverse of the mask, and \(\odot\) denoting element-wise multiplication.
Additionally, a perceptual loss ensures visual similarity between the edited and target images in masked regions:
\begin{equation}
\mathcal{L}_{\text{perc}} = \mathbb{E}_{\hat{x}_1, x_2} \left\| V(\hat{x}_1 \odot M_1) - V(x_2 \odot M_2) \right\|^2.
\end{equation}
and a localized CLIP loss aligns the generated image's masked area with the edit prompt:
\[
\mathcal{L}_{\text{loc}}(\hat{x}_1, y_2^{\text{diff}}) = D_{\text{CLIP}}(\hat{x}_1 \odot M_1, y_2^{\text{diff}}).
\]

The final loss function for fine-tuning is:
\[
\left( 1 - \frac{t}{T} \right) \mathcal{L}_{\text{loc}} + \mathcal{L}_{\text{global}} + \frac{t}{T} \mathcal{L}_{\text{mask}} + \lambda_{\text{perc}} \mathcal{L}_{\text{perc}}.
\]

During inference, random noise is added to the input image, and masks guide the denoising process. Gaussian noise corruption is applied iteratively with a sampling ratio. At each sampling step, pixel values of the reconstructed image in the inverse mask region are replaced with the corrupted version of the input image:
\[
\tilde{z}_t = \overline{M} \odot z_t + M \odot \tilde{z}_t.
\]

This process preserves details of the original latent input \( z_1 \) and follows the decoding step to obtain the final image \( \hat{x}_1 \).

\sstitle{Mask extraction}
\begin{figure*}
    \centering
    \includegraphics[width=\linewidth]{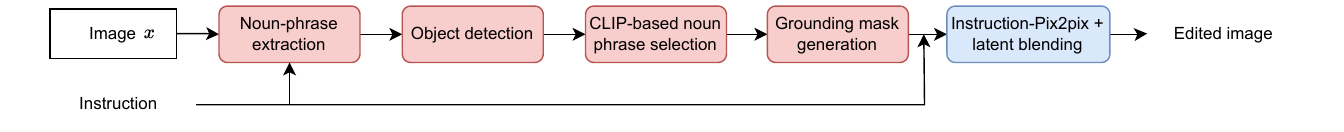}
    \caption{The pipeline of mask extraction.}
    \label{fig:maskextract}
\end{figure*}
Shagidanov et al.~\cite{shagidanov2024grounded} propose Grounded-Instruct-Pix2Pix framework, is divided into two main stages: grounding mask generation and localized image editing within the grounding mask. 
First, as shown in \autoref{fig:maskextract}, grounding mask generation starts with extracting noun phrases from the instruction using NLP tools like spaCy. For instance, in the instruction ``change the right panda to a brown bear'', the phrases ``the right panda" and "a brown bear'' are extracted. These phrases are matched with image sub-regions using the CLIP-Score~\cite{hessel2021clipscore} to identify the relevant object. This process is guided by the patterns in the user's edit instruction (e.g., ``change/make/turn/etc. A to/into/become/etc. B'') and involves using an object detector like GroundingDINO~\cite{liu2023grounding} to predict image sub-regions for each noun phrase. The best match is selected, and its bounding box is processed by the Segment-Anything model to produce the final grounding mask.
Second, the localized image editing is performed by modifying Instruct-Pix2Pix~\cite{brooks2023instructpix2pix} with mask-guided edit localization inspired by Blended Latent Diffusion~\cite{avrahami2023blended}. The process uses predefined steps of DDIM Inversion~\cite{song2021denoising} on Instruct-Pix2Pix for inversion latents \(\{x_1, x_2, ..., x_N\}\). Starting from a random noise latent \(x_T\), the denoising process begins, modifying the intermediate latent \(x_t\) using:
\[ 
x_t = 
\begin{cases} 
x_t & \text{if } t \geq N, \text{ or } t = 0 \\
x_t \ast M + \hat{x_t} \ast (1 - M) & \text{otherwise} 
\end{cases}
\]
where \(M\) is the grounding mask and \(\hat{x_t}\) is \(x_t\)'s corresponding inversion latent stored previously. This approach blends forward and backward latents with the grounding mask \(M\), preserving necessary areas during the denoising process. A low value for \(N\) (e.g., \(N = 0.1T\)) is chosen to allow most editing during initial steps, followed by blending non-edit areas.

\sstitle{Feature injection}
Tumanyan et al.~\cite{tumanyan2023plug} achieve fine-grained control over the generated structure by manipulating spatial features within the model. Spatial features from intermediate decoder layers encode localised semantic information and are less affected by appearance details~\cite{baranchuk2022labelefficient}. Self-attention helps retain fine layout and shape details.
The framework extracts features from \( I^G \) and injects them into the generation process of \( I^* \) along with \( P \), without requiring training or fine-tuning. This approach is applicable to both text-generated and real-world guidance images, using DDIM inversion for initial noise \( x_T^G \).
PCA analysis of features from each decoder layer shows that coarse layers separate foreground and background, intermediate layers encode shared semantic information, and deeper layers capture high-frequency details.
Feature injection is formalised by:
\[ z_{t-1}^* = \epsilon_{\theta} (x_t^*, P, t; \{ f_t^l \}) \]
Self-attention matrices \( A_t^l \) are used for fine-grained control~\cite{tumanyan2022splicing}:
\[ z_{t-1}^* = \epsilon_{\theta} (x_t, P, t; \{ A_t^l \}) \]
Negative prompting~\cite{liu2022compositional} balances neutral and negative guidances:
\[ \tilde{ \epsilon } = \alpha \epsilon_{\theta} (x_t, \emptyset, t) + (1 - \alpha) \epsilon_{\theta} (x_t, P_n, t) \]

\sstitle{Context Matching}
Meng et al.~\cite{meng2024instructgie} propose InstructGIE based on inpainting~\cite{bar2022visual,wang2023images}, capturing low-level visual editing contexts. The visual prompted condition is reformulated as a single image \( \text{Img}_{\text{VPcon}} = \{\text{Img}_0^{\text{in}}, \text{Img}_0^{\text{out}}, \text{Img}_1^{\text{in}}, \text{Grey}\} \), enhancing the global effective receptive field (ERF) to capture visual contexts. A vision encoder based on Zero-VMamba~\cite{liu2024vmamba} processes this visual prompted condition in latent space \( x_{\text{vpc}} \) as \( e_{\text{vpc}} = G_{\text{ven}}(x_{\text{vpc}}; \Theta_g) \). The conditioned latent diffusion model computes the output as \( y_{\text{vpc}} = \mathcal{F}(x; \Theta) + \mathcal{G}(\mathcal{F}(x + \mathcal{G}(x_{\text{vpc}}; \Theta_{g1}); \Theta_c); \Theta_{g2}) \).

An editing-shift-matching technique is introduced for enhanced in-context learning~\cite{cheng2022masked}. For each training ground truth \( \text{Img}_{\text{train}} = \{\text{Img}_0^{\text{in}}$, $\text{Img}_0^{\text{out}}, \text{Img}_1^{\text{in}}, \text{Img}_1^{\text{out}}\} \), an implicit editing shift value \( \mathcal{T}(\text{Img}) = \frac{1}{2} \sum_{i=0}^{1} \text{CLIP}(\text{Img}_i^{\text{in}}) - \text{CLIP}(\text{Img}_i^{\text{out}}) \) is calculated. The framework optimizes the editing shift loss \( \mathcal{L}_{\text{es}} = 1 - \frac{\mathcal{T}(\text{Img}^{\text{PO}}) \cdot \mathcal{T}(\text{Img}_{\text{train}})}{\|\mathcal{T}(\text{Img}^{\text{PO}})\| \times \|\mathcal{T}(\text{Img}_{\text{train}})\|} \).

To address low-quality details in diffusion-based image editing models, selective area matching targets differences in detailed editing areas between the original training ground truth \( \text{Img}_{\text{train}} \) and the reversed pseudo output \( \text{Img}^{\text{PO}} \). A frozen Mask2Former model \( \mathcal{M} \) obtains segmented masks and class information \( C \), filtering out predefined classes requiring special attention. The selective-area matching loss is optimised during training as 
\[ \mathcal{L}_{\text{sam}} = \frac{1}{N} \sum_{i=1}^{N} ((\text{Img}^{\text{PO}}_i \cdot [\text{mask}]_i) - (\text{Img}^{\text{train}}_i \cdot [\text{mask}]_i))^2 \].

\sstitle{Concept Forgetting}
Li et al.~\cite{li2024text} propose a method called learning and forgetting (LaF). LaF identifies and selectively forgets specific elements in the image that need modification. The process involves aligning the text prompt with the image's semantic content, using tools like OpenFlamingo~\cite{awadalla2023openflamingo} to generate detailed descriptions. The framework uses dependency parsing to locate and compare key elements in the image and text, guiding the editing process to target relevant objects~\cite{song2023bridge}. The goal is to forget specific concepts in the model's knowledge during inference~\cite{gandikota2024unified}, balancing positive and negative guidance signals with weights \( w \) and \( \eta \):
\[ \bar{e}_{\theta}^{(t)}(z_t) = e_{\theta}^{(t)}(z_t) + w \left( e_{\theta}^{(t)}(z_t, c_p) - e_{\theta}^{(t)}(z_t) \right) - \eta \left( e_{\theta}^{(t)}(z_t, c_n) - e_{\theta}^{(t)}(z_t) \right), \]
where \( c_p \) and \( c_n \) are input prompt and forgetting concepts. This ensures modifications align with the user's intent by targeting the correct elements in the image.

\sstitle{Human Alignment}
Geng et al.~\cite{geng2023instructdiffusion} introduce InstructDiffusion, a generalist modeling interface for various vision tasks, leveraging the Denoising Diffusion Probabilistic Model (DDPM)~\cite{brooks2023instructpix2pix}. It handles tasks like segmentation, keypoint detection, and image synthesis. 
InstructDiffusion focuses on three outputs: 3-channel RGB images, binary masks, and key points. This setup supports tasks such as keypoint detection, semantic segmentation, and image enhancement (e.g., deblurring and denoising), guided by detailed instructions. For example, keypoint detection uses instructions to improve accuracy, while segmentation employs semi-transparent masks to highlight specific regions.
The unified framework relies on diffusion models. Training includes pretraining adaptation, task-specific training, and instruction tuning. Pretraining adapts Stable Diffusion~\cite{rombach2022high} to handle specific tasks, introducing noise to the encoded latent \(\mathbf{z} = E(t_i)\), generating a noisy latent \(\mathbf{z}_t\). 
Human alignment further refines the model by using a curated dataset of the best-edited images, fine-tuning the model to ensure high-quality results in multi-turn editing scenarios~\cite{wei2022finetuned}.

\sstitle{Dynamic prompt}
Yang et al.~\cite{yang2024dynamic} address cross-attention leakage in prompt-based image editing with a dynamic prompt learning. The method refines initial prompts \( \mathcal{P} \) to improve cross-attention maps, crucial for generating accurate images based on user modifications. The main issues it tackles are distractor-object leakage and background leakage, which cause artifacts in generated images.
The method optimises word embeddings \( \mathbf{v}_t \) for each timestamp \( t \), rather than using a single embedding for all timestamps. The initial prompt \( \mathcal{P} \) consists of \( K \) noun words with learnable tokens updated over time.
Several losses are used to optimize the embeddings. The \emph{Disjoint Object Attention Loss} minimises overlap between attention maps of different objects:
\[
\mathcal{L}_{dj} = \sum_{i=1}^K \sum_{\substack{j=1 \\ j \ne i}}^K \cos(A_t^{v_i}, A_t^{v_j}).
\]
The \emph{Background Leakage Loss} minimises overlap with the background region \( \mathcal{B} \):
\[
\mathcal{L}_{bg} = \sum_{i=1}^K \cos(A_t^{v_i}, \mathcal{B}).
\]
The \emph{Attention Balancing Loss} ensures attention activations are concentrated on relevant objects:
\[
\mathcal{L}_{at} = \max_{v_t^k \in \mathcal{V}_t} (1 - \max(\mathcal{F}(A_t^{v_k}))).
\]
These losses are combined into the final optimization objective:
\[
\arg \min_{\mathbf{v}_t} \mathcal{L} = \lambda_{at} \cdot \mathcal{L}_{at} + \lambda_{dj} \cdot \mathcal{L}_{dj} + \lambda_{bg} \cdot \mathcal{L}_{bg}.
\]
Finally, Null-Text embeddings reconstruct the original image using null embeddings \( \emptyset_t \) for each timestamp.

\sstitle{Multi-task}
Sheynin et al.~\cite{sheynin2023emu} propose Emu-Edit, a diffusion model designed for a wide range of image editing tasks, including region-based and free-form edits as well as traditional computer vision tasks like detection and segmentation. Emu Edit uses task embeddings to identify the required semantic edit based on user instructions, addressing challenges with ambiguous or unique instructions by learning a unique task embedding for each task during training. This approach allows the model to adapt to new tasks through few-shot learning.
Emu Edit builds on the Emu model~\cite{dai2023emu}, which follows a two-stage training approach with pre-training on a large dataset and fine-tuning on a smaller, high-quality dataset~\cite{rombach2022high,brooks2023instructpix2pix}. The model employs a large U-Net (\(\epsilon_\theta\)) with 2.8 billion parameters and text embeddings from CLIP ViT-L~\cite{radford2021learning} and T5-XXL~\cite{raffel2020exploring}. 
Task embeddings (\(v_i\)) are integrated into the U-Net through cross-attention.
For few-shot learning, a new task embedding (\(v_{\text{new}}\)) is learned while keeping the model weights frozen:
\[
\min_{v_{\text{new}}} \mathbb{E}_{y, \epsilon, t} \left[ \| \epsilon - \epsilon_\theta (z_t, t, E(c_I), c_T, v_{\text{new}}) \|_2^2 \right].
\]
where \(\epsilon\) is the noise added by the diffusion process~\cite{lin2024common}, and \(y\) is a triplet of instruction, input image, and target image. 

\sstitle{Multi-turn}
To preserve quality in multi-turn editing, sequential edit thresholding is used. This involves computing the absolute difference image \(d\) and applying the following thresholding:
\[
c_I^{s+1} = \begin{cases} 
c_I^s & \text{if } \bar{d} < \alpha, \\
c_I^{s+1} & \text{otherwise},
\end{cases}
\]
where \(\bar{d}\) is a low-pass filtered version of \(d\), and \(\alpha = 0.03\) is the threshold value. This method helps smooth transitions and reduce artifacts in multi-turn edits.

\subsection{CLIP-based Controls}

\sstitle{Latent space manipulation}
Patashnik et al.~\cite{patashnik2021styleclip} explores three methods for text-driven image manipulation combining StyleGAN~\cite{karras2020analyzing} and CLIP~\cite{radford2021learning}. The first method uses simple latent optimization in $W+$ space, which is versatile but slow and hard to control. In particular, it adjusts a latent code \( w_s \) in StyleGAN's \( W+ \) space based on a text prompt \( t \). The optimization minimizes the function:
\begin{equation}
\text{arg min}_{w \in W+} \ D_{CLIP}(G(w), t) + \lambda_{L2} \|w - w_s\|_2 + \lambda_{ID} L_{ID}(w)
\end{equation}
where \( G \) is a StyleGAN generator, \( D_{CLIP} \) measures cosine distance in CLIP space, and \( L_{ID} \) uses ArcFace~\cite{deng2019arcface} to maintain identity similarity~\cite{richardson2021encoding}. The method uses gradient descent to achieve image edits that can explicitly alter visual characteristics or shift identity, adjusting \( \lambda_{L2} \) and \( \lambda_{ID} \) to control the extent of modification~\cite{tov2021designing}.

The second method employs a local mapper trained to infer manipulation steps in $W+$, achieving similar manipulation directions across different starting points but struggling with fine-grained, disentangled effects~\cite{patashnik2021styleclip}. More precisely, the method is called latent mapper, a more efficient approach for image manipulation in StyleGAN's W+ space~\cite{karras2019style} using a mapping network trained for specific text prompts. The network consists of three parts corresponding to coarse, medium, and fine layers of StyleGAN, each adjusted through:
\begin{equation}
M_t(w) = (M^c_t(w_c), M^m_t(w_m), M^f_t(w_f))
\end{equation}
The method minimizes the CLIP loss:
\begin{equation}
\label{eq:clip_loss}
L_{CLIP}(w) = D_{CLIP}(G(w) + M_t(w), t)
\end{equation}
combined with L2 regularization and identity preservation losses:
\begin{equation}
L(w) = L_{CLIP}(w) + \lambda_{L2} \| M_t(w) \|_2 + \lambda_{ID} L_{ID}(w)
\end{equation}
This method achieves precise control over specific image attributes while maintaining other visual qualities. It's noted that despite custom manipulations for each image, the diversity in manipulation directions across different prompts is limited.

The third method generates global manipulation directions in StyleGAN's style space \( S \)~\cite{wu2021stylespace} for fine-grained, disentangled image manipulations from text prompts~\cite{patashnik2021styleclip}. The process involves:
(1) Translating a text prompt into a stable vector \( \Delta t \) in CLIP's embedding space~\cite{radford2021learning}, optimized through prompt engineering to accurately reflect the desired attribute change.
(2) Mapping \( \Delta t \) into a manipulation direction \( \Delta s \) in \( S \). This is achieved by assessing the relevance of each style channel to \( \Delta t \), using image pairs to calculate cosine similarity changes, and filtering through:
   \begin{equation}
   \Delta s = \begin{cases} 
   \Delta i_c \cdot \Delta i & \text{if } |\Delta i_c \cdot \Delta i| \geq \beta \\
   0 & \text{otherwise}
   \end{cases}
      \end{equation}
This approach allows precise control over specific attributes in image manipulation, ensuring that changes are targeted and do not affect unrelated visual aspects.

\begin{figure}
    \centering
    \includegraphics[width=\linewidth]{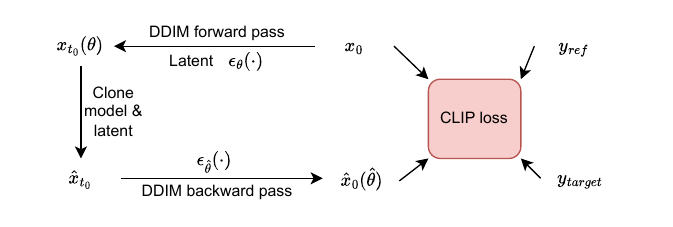}
    \caption{Domain translation using CLIP loss.}
    \label{fig:domain-translate}
\end{figure}

\sstitle{Domain translation}
Kim et al.~\cite{kim2022diffusionclip} propose a model, namely DiffusionCLIP, that can translate images between unseen domains and synthesizing images based on stroke inputs in novel domains. The model employs pretrained diffusion models to facilitate zero-shot translation, adapting images to new styles and domains without needing training examples from those domains. Additionally, it can generate images that maintain the identity and semantics of the original while adopting the characteristics of the target domain, utilizing a combination of forward and reverse DDIM processes~\cite{choi2021ilvr,meng2022sdedit} (see \autoref{fig:domain-translate}). Formally, the model involves converting an initial image, \(x_0\), into a latent space representation, \(x_t(\theta)\), using a pretrained diffusion model~\cite{ho2020denoising}. This model is then fine-tuned to generate new samples driven by a target text, \(y_{tar}\), using the CLIP loss and a reverse diffusion process known as deterministic DDIM~\cite{song2021denoising}.
The fine-tuning of the reverse diffusion model, \(\theta\), employs a loss function that includes directional CLIP loss, \(L_{direction}\), and identity loss, \(L_{ID}\). The directional CLIP loss ensures the generated image aligns with the text prompt, while the identity loss preserves the original image's characteristics like expression or hair color. Specifically, the identity loss used is \(L_1\) loss, which measures the pixel-wise absolute difference between the original and the generated images. Additionally, a face identity loss, \(L_{face}\), is included for face manipulations, which helps in retaining facial features~\cite{deng2019arcface}.
The combined objective function for fine-tuning is given by:
\begin{multline}
L_{direction} (\hat{x}_0(\theta), y_{tar}, x_0, y_{ref}) + \lambda_1 L_1(\hat{x}_0(\theta), x_0) \\ + \lambda_{face} L_{face}(\hat{x}_0(\theta), x_0)
\end{multline}
where \(\lambda_1\) and \(\lambda_{face}\) are weights for the \(L_1\) and face identity losses, respectively.
The architecture also involves a shared U-Net~\cite{ronneberger2015u} with positional embeddings similar to those in Transformer models~\cite{vaswani2017attention}, ensuring the generated images adhere to the text prompts while preserving identity elements like facial features~\cite{dhariwal2021diffusion}.
Kim et al.~\cite{kim2022diffusionclip} also propose a fast sampling strategy. Instead of iterating the diffusion process up to the final time step \(T\), the model accelerates both the forward and reverse diffusion by reducing the number of steps (referred to as the 'return step'). This is achieved by tuning parameters \(t_0\), \(S_{for}\), and \(S_{gen}\) which control the starting, forward, and generative steps respectively.

\begin{figure}
    \centering
    \includegraphics[width=\linewidth]{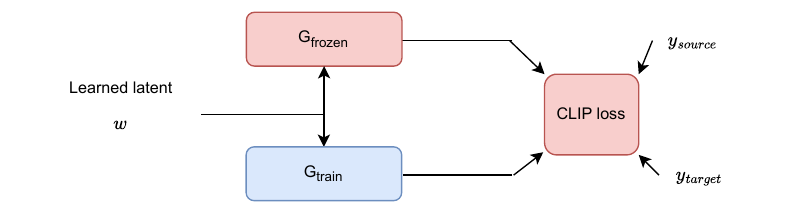}
    \caption{Domain adaptation using CLIP loss.}
    \label{fig:domain-adapt}
\end{figure}

\sstitle{Domain adaption}
Gal et al.~\cite{gal2022stylegan} propose a ``Layer-Freezing'' technique for adapting neural networks to domain shifts like transforming a photo to a sketch (see \autoref{fig:domain-adapt}). This method involves selectively training specific network weights to stabilize learning and improve output quality, particularly for complex transformations~\cite{mo2020freeze}. It identifies critical weights and involves a two-phase process:
(1) \emph{Layer Selection Phase:} Latent codes are optimized using a Global CLIP loss~\autoref{eq:clip_global} with all network weights fixed~\cite{abdal2019image2stylegan}.
(2) \emph{Optimization Phase:} Selected layers are unfrozen and their weights are optimized using a Directional CLIP loss to achieve targeted changes (e.g., ``Dog" to ``Cat").
This approach focuses on maintaining realistic image generation by training only the most relevant layers, thus avoiding overfitting and unrealistic results. However, there are challenges with shape transformations using generative models, specifically when attempting to transform images of dogs into cats. The generator does not fully convert dogs to cats but produces a mix of cat, dog, and intermediate images. To address this, Gal et al.~\cite{gal2022stylegan} propose a latent mapper to edit latent codes within the generator's domain to enhance the transformation towards more cat-like features, effectively refining the output to better match the target domain.

\sstitle{Latent space discretisation}
Crowson et al.~\cite{crowson2022vqgan} propose VQGAN-CLIP, a generative editing approach that combines the VQGAN~\cite{esser2021taming} and CLIP~\cite{radford2021learning} models. The process involves using a GAN to generate candidate images, which are then optimized to align with the semantics of the text prompt by minimizing the squared spherical distance between their embeddings. The approach employs discrete latent spaces to address the continuous nature of image data, using vector quantization to create a discrete codebook of embeddings that facilitate the generation and reconstruction of images through an autoencoder.
Additionally, a regularization term is incorporated into the loss function to enhance the structural coherence and reduce unwanted textures in the generated images:
\begin{equation}
	\text{Loss} = L_{\text{CLIP}} + \alpha \cdot \frac{1}{N} \sum_{i=0}^N Z_i^2
\end{equation}
with \(\alpha\) being the regularization weight.
This method enables flexibility, allowing the integration of various techniques like handling multiple text prompts simultaneously and employing masking to refine image editing~\cite{couairon2022flexit,frans2022clipdraw}, which enhances the semantic coherence of the output~\cite{michel2022text2mesh}.

\sstitle{Multi-channel Directions}
Kocasari et al.~\cite{kocasari2022stylemc} propose a methodology using StyleGAN2, focusing on the style space \( S \)~\cite{tov2021designing,wu2021stylespace}, which offers a disentangled representation ideal for instruction-based image modifications. The method, named StyleMC (Style Space based Multi-Channel Directions), employs a pre-trained StyleGAN2 generator \( G \) and manipulates the style code \( s \in S \) by introducing a direction \( \Delta s \) to align the generated image with changes specified in the text prompt~\cite{karras2020analyzing}.
To optimize \( \Delta s \), the approach uses a combination of CLIP loss \( L_{CLIP} \) and identity loss \( L_{ID} \). The CLIP loss minimizes the cosine distance between the CLIP embeddings of the generated image and the text (\autoref{eq:clip_loss}).
The identity loss ensures that modifications maintain the original image's identity features, using a network \( R \) to measure similarity:
\begin{equation}
L_{ID} = 1 - (R(G(s)), R(G(s + \Delta s)))
\end{equation}
These losses are combined to find the optimal manipulation direction:
\begin{equation}
\text{arg min}_{\Delta s} \lambda_C L_{CLIP} + \lambda_D L_{ID}
\end{equation}
where \( \lambda_C \) and \( \lambda_D \) weight the CLIP and identity losses, respectively.
This efficient method focuses on low-resolution layers to find the manipulation direction and operates on small image batches within the \( S \) space, allowing for precise and complex edits that are closely aligned with textual modifications while preserving the original image's identity.

\sstitle{Null-text inversion}
Instead of using random noise vectors~\cite{roich2022pivotal}, Mokady et al.~\cite{mokady2023null} employ a single noise vector inspired by GAN literature for efficient optimization. It begins with DDIM inversion at \( w = 1 \) for a rough approximation, then optimizes around this trajectory using a larger guidance scale \( w > 1 \) to balance accuracy and editability.
The method optimises only the unconditional embedding \( \varnothing \), initialized with the null-text embedding, enabling high-quality reconstruction and intuitive editing~\cite{gal2023an,ruiz2023dreambooth,valevski2023unitune}. This global null-text optimization uses different ``null embeddings'' \( \varnothing_t \) for each timestep \( t \), enhancing reconstruction quality and efficiency~\cite{kawar2023imagic}.

\sstitle{Localization-aware inversion}
Tang et al.~\cite{tang2024locinv} propose an editing framework focusing on Dynamic Prompt Learning (DPL)~\cite{yang2024dynamic} and Localization-aware Inversion (LocInv). These methods aim to improve alignment between attention maps and nouns, addressing cross-attention leakage.
DPL introduces three losses, leveraging localization priors from segmentation or detection boxes~\cite{zou2024segment}, denoted as \( S \). Cross-attention maps are derived from the Diffusion Model UNet, using features of the noisy image \( \psi(z_t) \) projected to \( Q_t = l_Q(\psi(z_t)) \) and textual embeddings to \( K = l_K(C) \). The attention map \( A_t \) is computed as \( \text{softmax}(Q_t \cdot K^T / \sqrt{d}) \). Tokens \( V_t \) are updated to ensure alignment~\cite{yang2024dynamic}.
LocInv uses similarity loss \( L_{\text{sim}} \) and overlapping loss \( L_{\text{ovl}} \):
\[
L_{\text{sim}} = \sum_{i=1}^{K} \left[ 1 - \cos(A_t^{v_i}, S_t^{v_i}) \right]
\]
\[
L_{\text{ovl}} = 1 - \frac{\sum_{i=1}^{K} A_t^{v_i} \cdot S_t^{v_i}}{\sum_{i=1}^{K} A_t^{v_i}}
\]
The combined loss function is:
\[
\arg \min_{V_t} L = \lambda_{\text{sim}} \cdot L_{\text{sim}} + \lambda_{\text{ovl}} \cdot L_{\text{ovl}}
\]
Adjective binding aligns adjectives with corresponding nouns using a Spacy parser, with loss \( L_{\text{adj}} \):
\[
L_{\text{adj}} = \sum_{i=1}^{K} \left[ 1 - \cos(A_t^{v_i}, A_t^{a_i}) \right]
\]
This loss is added when adjective-editing is required.

\sstitle{Multi-entities}
Wang et al.~\cite{wang2022manitrans} propose the ManiTrans framework using transformers, focusing on autoencoder models for downsampling and quantizing images into discrete tokens, and transformer models for modeling their joint distribution.
The autoencoder-based Trans model includes a convolutional encoder \( E \), a convolutional decoder \( G \), and a codebook \( \mathbf{Z} \in \mathbb{R}^{K \times n_z} \). Given an image \( \mathbf{X} \in \mathbb{R}^{H \times W \times 3} \), the encoder \( E \) generates a latent feature map \( \mathbf{Q} \in \mathbb{R}^{h \times w \times n_z} \). Quantisation replaces each pixel embedding with its closest latent variable from the codebook:
$ \hat{Q}_{ij} = \arg \min_{z_k} \| Q_{ij} - z_k \|^2$.
The decoder \( G \) reconstructs the image \( \hat{\mathbf{X}} \approx \mathbf{X} \).
Image generation treats the quantized feature map \( \hat{\mathbf{Q}} \) as a sequence of discrete tokens, predicted autoregressively:
$ P(I_{\leq i} | \mathbf{T}) = \prod_{j} P(I_j | I_{< j}, \mathbf{T})$.
The losses for text and image tokens are:
\[ \mathcal{L}_{\text{txt}} = -\mathbb{E}_{\mathbf{T}_1} \log P(T_i | T_{<i}) \]
\[ \mathcal{L}_{\text{img}} = -\mathbb{E}_{\mathbf{I}_1} \log P(I_i | I_{<i}, \mathbf{T}) \]

Language guidance uses the CLIP model for visual-semantic alignment, with the semantic loss:
$ \mathcal{L}_{\text{semantic}} = 1 - D(G(\hat{\mathbf{I}}), \mathbf{T})$,
where \( D \) is the cosine similarity between image and text embeddings.
Vision guidance involves appending grayscale image tokens \( \mathbf{V} \) to the text sequence, sharing positional embeddings with \( \mathbf{I} \), and applying the vision guidance loss:
$ \mathcal{L}_{\text{gray}} = -\mathbb{E}_{\mathbf{V}_i} \log P(V_i | V_{<i})$.
The total training loss combines these losses:
\[ \mathcal{L}_{\text{ar}} = \lambda_1 \mathcal{L}_{\text{img}} + \lambda_2 \mathcal{L}_{\text{gray}} + \lambda_3 \mathcal{L}_{\text{txt}} \]
\[ \mathcal{L}_{\text{total}} = \mathcal{L}_{\text{ar}} + \lambda_4 \mathcal{L}_{\text{semantic}} \]

During inference, entity guidance uses a semantic alignment module to find and manipulate relevant entities in the image. Entity segmentation maps each entity to the latent feature map, and a text prompt selects the relevant entities by calculating token-level similarities, ensuring precise manipulation based on their alignment with the prompt word.

\sstitle{Region generation}
Lin et al.~\cite{lin2024text} propose mask-free local image editing using a region generation network.
Given an input image \( X \) and text \( T \), features are extracted using a pre-trained visual transformer model, ViT-B/16~\cite{caron2021emerging}. Anchor points \( \{C_i\} \) are initialised at high-scoring patches from the self-attention map. For each anchor point, bounding box proposals \( B_i \) are generated~\cite{simeoni2021localizing}. A region generation network then predicts scores for these proposals using a softmax function, with the Gumbel-Softmax trick aiding in training.
The CLIP model~\cite{radford2021learning} guides the image editing through a composite loss function, combining CLIP guidance loss \( \mathcal{L}_{\text{Clip}} \), directional loss \( \mathcal{L}_{\text{Dir}} \), and structural loss \( \mathcal{L}_{\text{Str}} \)~\cite{patashnik2021styleclip,kolkin2019style}. These losses ensure the edited image aligns with text descriptions while preserving texture and spatial layout.
During inference, a quality score \( S \) ranks edited images from different anchor points, using \( S_{t2i} \) (similarity between text and edited images) and \( S_{i2i} \) (similarity between source and edited images). The final image is selected based on the highest score \( S \).
The method is compatible with various image editing models~\cite{avrahami2023blended,nichol2022glide,chang2023muse}, such as MaskGIT~\cite{chang2022maskgit} and Stable Diffusion~\cite{rombach2022high}, demonstrating its versatility and applicability across different architectures.

\subsection{Attention-based Controls}

Zhao et al.~\cite{zhao2024instructbrush} propose methods for optimizing instructions in image editing. Inspired by Textual Inversion~\cite{gal2023an}, it focuses on optimizing features in cross-attention layers rather than directly in the CLIP space.
The proposed method projects features from textual embeddings to align with image features, enhancing detailed representation~\cite{nguyen2023visual}. The cross-attention is given by:
\[ \text{Attention} (Q, K, V) = \text{Softmax} \left( \frac{Q K^T}{\sqrt{d}} \right) V \]
where \( K \) and \( V \) represent the token length and feature dimension of the instruction~\cite{hertz2023prompt}. The optimization objective for feature embeddings is:
\[ \gamma = \arg \min_{\gamma} \mathbb{E}_{e_{i},c_{i}} \mathbb{E}_{\tau, \epsilon} \left[ \|e - e_{\theta} (z_{t}, t, E(e_{i}, c_{i}, \gamma))\|_{2}^{2} \right] \]

\begin{figure}
    \centering
    \includegraphics[width=\linewidth]{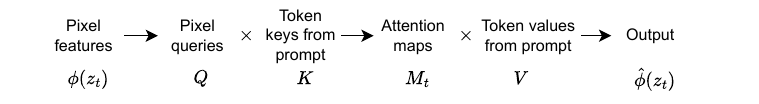}
    \caption{Cross attention for text-driven image edits.}
    \label{fig:cross-attn}
\end{figure}

\sstitle{Cross Attention}
Hertz et al.~\cite{hertz2023prompt} use cross-attention mechanisms~\cite{vaswani2017attention} within the diffusion model~\cite{saharia2022photorealistic,rombach2022high} to direct edits based on changes to a text prompt while preserving the original image's structure (see \autoref{fig:cross-attn}). The technique focuses on the interaction between image pixels and text embeddings, forming attention maps through a calculated softmax function on the scaled dot product of query \(Q\) and key \(K\):
\[
M = \text{Softmax} \left( \frac{QK^T}{\sqrt{d}} \right)
\]
where \(d\) is the dimensionality of the keys and queries. These attention maps guide pixel updates to specific regions relevant to the modified text, allowing for precise and localized image modifications without using masks.
Additionally, the method includes strategies for enhancing the coherence and visual continuity of the original image during editing, leveraging both cross-attention and self-attention layers' roles. This method aims to improve the precision and contextual sensitivity of text-driven image edits, ensuring that modifications are integrated seamlessly and maintain the integrity of the original image.

\sstitle{Mutual Self-Attention}
Cao et al.~\cite{cao2023masactrl} maintain the object contents (e.g., textures and identity) by proposing mutual self-attention to query similar contents from the sourced image \( I_s \) during the denoising process, combining the semantic layout from the instruction with the details of \( I_s \).
It involves assembling self-attention inputs by keeping query features \( Q \) unchanged and obtaining key and value features \( K_s \) and \( V_s \) from \( I_s \). The method includes a mask-aware strategy~\cite{hertz2023prompt,tang2023daam} to distinguish foreground and background using masks \( M_s \) and \( M \), defined as:
\[ f_o^l = \text{Attention}(Q^l, K_s^l, V_s^l; M_s) \]
\[ f_b^l = \text{Attention}(Q^l, K_s^l, V_s^l; 1 - M_s) \]
\[ \bar{f}^l = f_o^l \cdot M + f_b^l \cdot (1 - M) \]
The approach integrates with controllable diffusion models (e.g., T2I-Adapter~\cite{mou2024t2i}, ControlNet~\cite{zhang2023adding}) for more precise image synthesis, allowing for desired poses and shapes by querying image contents from \( I_s \).

\sstitle{Unified Attention}
Hertz et al.~\cite{hertz2023prompt} found that text-image interactions occur in the noise prediction network, leading to attention control methods to align noise \(\mathbf{\epsilon}_{\theta}^{\text{tgt}}\) with language prompts.
The U-Net noise predictor uses cross-attention and self-attention modules~\cite{avrahami2022blended}. Queries (Q) are projected from spatial features, and keys (K) and values (V) from text features in cross-attention. In self-attention, keys and values come from spatial features. 
For rigid semantic changes, cross-attention replaces generated images' attention maps with those of original images. Global and local attention refinement integrates source and target prompts to preserve and incorporate changes.
Non-rigid changes are managed by mutual self-attention control, which captures object layouts. MasaCtrl~\cite{cao2023masactrl} noted that self-attention layers effectively handle these changes by combining structural layouts from the source with target prompts.

Unified Attention Control (UAC)~\cite{xu2024inversion} combines cross-attention and mutual self-attention control, adding a layout branch to manage composition and structure. During diffusion, mutual self-attention initiates refinement, followed by cross-attention and layout adjustments to reflect non-rigid changes.

\sstitle{Grouped Cross-Attention}
Feng et al.~\cite{feng2024item} propose the D-Edit framework for versatile image editing using diffusion models and disentanglement control. 
Text prompts are integrated through cross-attention mechanisms~\cite{radford2021learning,tang2023daam}, encoding the prompt \( P \) into an embedding \( c = g_\phi(P) \) and using it with image latent \( z_t \). The attention matrix is calculated as \( A = \text{softmax}(q k^T) \).
The image \( I \) is segmented into items, each linked to a distinct prompt \( P_i \), creating disentangled cross-attentions~\cite{tang2023emergent}. Linking prompts to items involves adding new tokens to the text encoder and finetuning the model to optimise UNet parameters~\cite{ruiz2023dreambooth}.
The optimised model supports various editing operations: replacing items via text prompts, swapping items between images, editing item appearance through masks, and removing items by deleting their masks and prompts, filling the region with nearby masks and prompts.

\sstitle{Self-Attention Injection}
Kwon et al.~\cite{kwon2024unified} propose a method combining single image editing with self-attention injection (PNP)~\cite{tumanyan2023plug} and sequential video editing using a shared attention framework (Pix2Vid)~\cite{ceylan2023pix2video}, enhancing performance and consistency across 3D, video, and panorama editing.

Assuming a series of images \( I_1, I_2, ..., I_N \) for editing, one image \( I_{ref} \) is used as a reference to ensure consistent context. The U-Net architecture extracts self-attention and resnet features from DDIM inversion paths, injecting them during sampling to edit attributes without altering image structure.
Inversion and Feature Extraction involve sending \( I_{ref} \) and each \( I_i \) to the noisy latent space through DDIM Inversion, extracting features. During sampling step \( t \), self-attention layer query and key features \( Q_{ref}^{l,t} \) and \( K_{ref}^{l,t} \) are obtained and used.
Disentangled Self-Attention Injection begins with the noisy latent \( z_{ref}^T \) and \( z_i^T \). Reverse diffusion injects features obtained earlier to edit while maintaining \( I_{ref} \)'s structure, propagating to other frames. This process ensures structure and context consistency.
Injection Scheduling varies self-attention injection across timesteps to ensure consistent image editing.

\subsection{Hybrid Controls}

\sstitle{Mixture-of-Expert}
Li et al.~\cite{li2023moecontroller} propose the use of mixture-of-expert controllers via global and local categories. 
The system includes a cross-attention network to fuse text and image features, expert models with various capabilities, and a gated system for automatic adaptation.

\begin{figure}[!h]
    \centering
    \includegraphics[width=\linewidth]{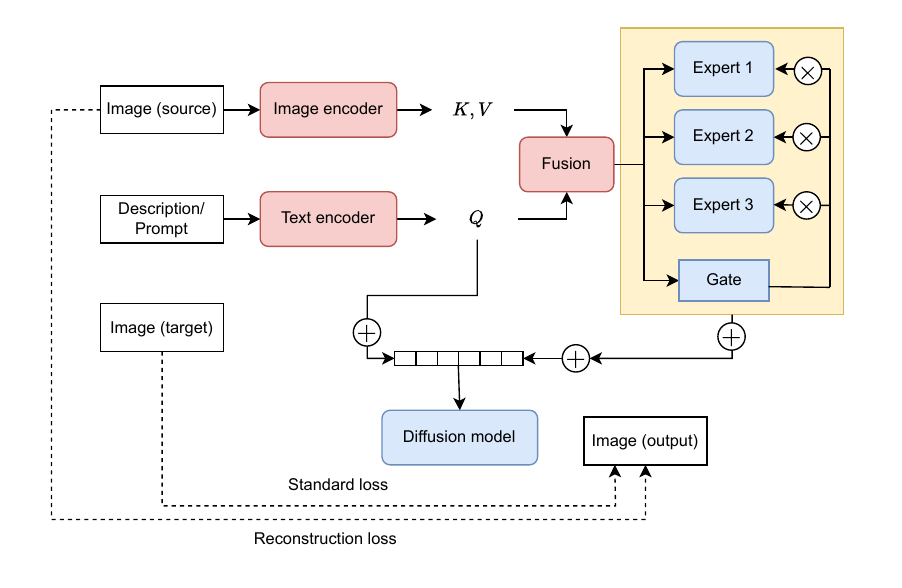}
    \caption{The training procedure of mixture-of-expert.}
    \label{fig:mix-expert}
\end{figure}

The approach divides tasks into global editing, local editing, and fine-grained editing, using three expert models: Expert 1 for fine-grained local tasks, Expert 2 for global style transfer, and Expert 3 for local editing (see \autoref{fig:mix-expert}). The fusion module integrates text features \(Q\) and image features \(K\) and \(V\) from the CLIP Image Encoder, combining expert outputs as:
\[
c = \sum_{i=1}^{n} g_i(x) f_i(x) + E_t(y_t),
\]
where \(g_i(x)\) is the gate's softmax output, and \(CA(Q, K, V)\) is the cross-attention network.
To ensure consistency between generated and original images, a reconstruction loss is added during training~\cite{ruiz2023dreambooth}, as follows:
\[
\mathcal{L} = \mathbb{E}_{z, \epsilon \sim \mathcal{N}(0, I), t} \left[\|\epsilon - f_\theta (z_t, tgt, t, c)\|_2^2 + w \|\epsilon - f_\theta (z_t, src, t, c)\|_2^2\right].
\]

\sstitle{Disentangled Transfer}
Dalva et al.~\cite{dalva2024gantastic} propose GANTASTIC framework, in which semantic directions are transferred from StyleGAN to the diffusion process. Firstly, the StyleGAN generation process involves multiple latent spaces (\( \mathcal{Z}, \mathcal{W}, \mathcal{W^+}, \) and \( \mathcal{S} \))~\cite{simsar2023fantastic,wu2021stylespace}. The generator \( G \) maps \( \mathcal{Z} \) to the target image domain \( \mathcal{X} \). The latent code \( \mathbf{z} \) is drawn from a Gaussian distribution and transformed into \( \mathcal{W} \) using a mapper function. Latent vectors in \( \mathcal{W} \) are then transformed into channel-wise style parameters in \( \mathcal{S} \), facilitating extensive image editing.
Secondly, diffusion models create data samples via iterative denoising, reversing a noise sequence over steps \( t \)~\cite{ho2020denoising}. The denoising network \( \epsilon_\theta \) predicts the noise component \( \epsilon \) during the reverse process. The objective function for training this network is:
$ \mathcal{L}_{DM} = \mathbb{E}_{x_0, x_t \sim \mathcal{N}(0,1), t} \left[ \left| \left| \epsilon^t - \epsilon_\theta(x_t, t) \right| \right|^2_2 \right] $.
For classifier-free guidance~\cite{ho2022classifier}, the conditional noise prediction \( \tilde{\epsilon}_\theta(x_t, c) \) is:
\[ \tilde{\epsilon}_\theta(x_t, c) = \epsilon_\theta(x_t, \phi) + \lambda_g (\epsilon_\theta(x_t, c) - \epsilon_\theta(x_t, \phi)) \]
The method aims to learn a latent direction \( \mathbf{d} \) for translating \( X_{input} \) to \( X_{edited} \). The learning objective \( \mathcal{L} \) combines semantic alignment loss \( \mathcal{L}_{sem} \) and latent alignment loss \( \mathcal{L}_{latent} \)~\cite{radford2021learning}:
\[ \mathcal{L} = \mathcal{L}_{sem} + \mathcal{L}_{latent} \]
Image editing with a latent direction \( \mathbf{d_i} \) is performed to reflect the desired semantic. Using classifier-free guidance, the editing term is:
$ \tilde{\epsilon}_\theta(x_t, c, d) = \tilde{\epsilon}_\theta(x_t, c) + \lambda_e (\epsilon_\theta(x_t, d) - \epsilon_\theta(x_t, \phi))$.
For multiple edits, this term is expanded over direction set \( D \):
$ \tilde{\epsilon}_\theta(x_t, c) + \sum_{i=1}^{|D|} \lambda_e (\epsilon_\theta(x_t, d_i) - \epsilon_\theta(x_t, \phi))$.
Real image editing uses DDPM Inversion~\cite{huberman2024edit} to obtain \( x_T \) and reformulates the noise prediction \( \tilde{\epsilon}_\theta(x_t, d) \):
\[ \tilde{\epsilon}_\theta(x_t, d) = \epsilon_\theta(x_t, \phi) + \lambda_e (\epsilon_\theta(x_t, d) - \epsilon_\theta(x_t, \phi)) \]

\section{Learning Paradigms}
\label{sec:learning}

\subsection{Self-Supervision}

\sstitle{Adversarial training}
Li et al.~\cite{li2020manigan} propose a self-supervised way to train a generative model for text-driven image manipulation. It follows a similar adversarial training approach to ControlGAN~\cite{li2019controllable}, with the addition of a regularisation term in the generator's objective function:
\begin{equation}
 L_{reg} = 1 - \frac{1}{CHW} ||I' - I||	
\end{equation}
This term discourages the model from learning an identity function where the generated image \( I' \) is identical to the input image \( I \). Training differs from~\cite{li2019controllable} in that it does not use paired text-to-image data, instead it constructs a loss function using a natural question approach.
The training stops when the model reaches the best balance between generating text-described attributes and preserving text-irrelevant content, measured by an evaluation metric called manipulative precision.

\sstitle{Counterfactual reasoning}
Fu et al.~\cite{fu2020sscr} develops a training approach using Self-Supervised Counterfactual Reasoning (SSCR) to enhance data diversity in the face of scarcity. New instructions \( I' \) are generated through interventions on original instructions \( I \), forming an augmented dataset \( U' = \{I', O\} \).
For each turn \( t \), a counterfactual image \( V'_t \) is predicted from the previous image features and the counterfactual instruction history \( h'_t \) as:
\begin{equation}
V'_t = G(f_{t-1}, h'_t)	
\end{equation}
An iterative explainer \( E \) computes the loss \( L'_E \) for updating the generator \( G \), based on the difference between the original and reconstructed instructions:
\begin{equation}
L'_E = \sum_{i=1}^{L} CE_{loss}(\hat{w}'_i, w'_i)	
\end{equation}
The training process optimizes \( G \) by balancing the discriminator's loss \( L_G \), the token-level loss \( L_E \), and the discriminator's binary loss \( L_D \),
aiming to train the generator to better interpret and visualize diverse counterfactual instructions, improving its generalizability.

\sstitle{Instance-level preservation}
Xia et al.~\cite{xia2021tedigan} propose an instance-level optimization for face manipulation, focusing on preserving identity. Overcoming the difficulty of maintaining facial identity while altering features~\cite{deng2019arcface}, the authors propose a new method for text-guided image manipulation using an optimization module that balances desired changes with identity preservation~\cite{ververas2020slidergan,richardson2021encoding}.
The optimization objective is:
\begin{equation}
\mathbf{z}^* = \text{argmin}_{\mathbf{z}} \| \mathbf{x} - G(\mathbf{z}) \|^2_2 + \lambda_1' \| F(\mathbf{x}) - F(G(\mathbf{z})) \|^2_2 + \lambda_2' \| \mathbf{z} - E_v(G(\mathbf{z})) \|^2_2.	
\end{equation}
Here, \(\mathbf{x}\) is the original image, \(G\) is the generator, and \(\lambda_1'\) and \(\lambda_2'\) are loss weights.

\subsection{Supervised Fine-Tuning}

\sstitle{Instruction tuning}
Gan et al.~\cite{gan2024instructcv} propose the use of an instruction-tuning dataset \( D_T = \{(x_i, v(y_i), I_i)\} \) to train a conditional diffusion model~\cite{brooks2023instructpix2pix}. This model performs vision tasks specified by instructions \( I \) on input images \( x \), producing task outputs \( v(y) \). By fine-tuning a text-to-image diffusion model with \( D_T \), the model evolves into a language-guided multi-task learner.
Diffusion models denoise a normally distributed random variable, reversing a Markov chain's dynamics~\cite{ho2020denoising}. The process transforms \( v(y) \) into a latent representation \( z_0 = E(v(y)) \), injecting Gaussian noise over \( T \) steps:
\[
q(z_t|z_{t-1}) = \mathcal{N}(z_t; \sqrt{1 - \beta_t}z_{t-1}, \beta_t \mathbf{I}),
\]
where \(\beta_t\) controls the noise. The reverse process recovers \( z_0 \):
\[
p_\theta(z_{t-1}|z_t) = \mathcal{N}(z_{t-1}; \mu_\theta(z_t, t), \sigma^2_t \mathbf{I}),
\]
where \(\mu_\theta\) is parameterized by neural networks. The model is trained to optimize~\cite{song2019generative}:
\[
\mathcal{L}_{\text{unconditional}} = \mathbb{E}_{v(y), \epsilon, t} \left[ \|\epsilon - \epsilon_\theta(z_t, t)\|^2_2 \right].
\]
For conditional generation~\cite{brooks2023instructpix2pix,rombach2022high}, the objective is:
\[
\mathcal{L}_{\text{conditional}} = \mathbb{E}_{v(y), E(x), I, \epsilon, t} \left[ \|\epsilon - \epsilon_\theta(z_t, t, E(x), I)\|^2_2 \right].
\]
Similarly, Fu et al. \cite{fu2024guiding} uses multimodal embeddings to align instructions and images in a shared space, optimizing with cross-entropy loss for better instruction comprehension and visual feature alignment for consistency. SmartEdit \cite{huang2023smartedit} extends multimodal editing to handle complex, sequential, or multi-object edits by using a reasoning segmentation loss to enhance perception and object localization within scenes.

Finally, classifier-free guidance~\cite{ho2022classifier} is applied to enhance alignment between outputs and conditioning, combining unconditional and conditional noise predictors.

\sstitle{Classifier-free guidance (CFG)}
This technique biases samples towards a specific conditioning, like a text prompt, improving text-image alignment and fidelity at the expense of mode coverage~\cite{ho2022classifier}. In text-guided image inpainting, CFG ensures strong alignment between the generated image and the text prompt~\cite{patashnik2021styleclip,yang2023paint,saharia2022photorealistic}. Using high guidance weights with oscillation, particularly a schedule oscillating between 1 and 30, provides the best balance between sample fidelity and text-image alignment~\cite{wang2023imagen,ho2022imagen}.
In the case of diffusion models, CFG enhances the generation quality by directing the probability mass toward more likely data points as per an implicit classifier \( p_{\theta}(c|\mathbf{z}_t) \)~\cite{gafni2022make,nichol2022glide}. This method is often applied in class-conditional and text-conditional image generation to ensure that generated images align better with their specified conditions~\cite{brooks2023instructpix2pix}. It involves training the diffusion model on both conditional and unconditional denoising tasks. At inference time, the model uses a guidance scale \( s \geq 1 \), which modulates the influence of the conditioning. The modified score estimate \( \tilde{\theta}(\mathbf{z}_t, c) \) used for this guidance is calculated as follows:
\begin{equation}
\label{eq:diffusion_guidance}
 \tilde{\theta}(\mathbf{z}_t, c) = \theta(\mathbf{z}_t, \emptyset) + s \cdot (\theta(\mathbf{z}_t, c) - \theta(\mathbf{z}_t, \emptyset)) 
 \end{equation}
Here, \( \theta(\mathbf{z}_t, c) \) is the score estimate for the conditional generation, \( \theta(\mathbf{z}_t, \emptyset) \) for the unconditional generation, and \( s \) is the guidance scale, enhancing the conditioning effect on the generated output.

Classifier-free guidance offers two advantages~\cite{nichol2022glide}. Firstly, it enables a model to use its own knowledge for guidance, eliminating the need for a separate classification model. Secondly, it simplifies guidance when conditioning on challenging-to-classify information, like text.

\sstitle{CLIP guidance}
Contrastive Language-Image Pre-training (CLIP) involves pretraining a text encoder and an image encoder to match texts with images in a dataset, enabling efficient visual concept learning via natural language supervision~\cite{radford2021learning}. Two specific loss functions have been developed for better knowledge extraction from CLIP~\cite{kim2022diffusionclip,patashnik2021styleclip,gal2022stylegan}:
\begin{compactitem}
  \item \emph{Global Target Loss:} This loss minimizes the cosine distance in the CLIP space between the generated image and a given target text.
\begin{equation}
\label{eq:clip_global}
 L_{global}(x_{gen}, y_{tar}) = D_{CLIP}(x_{gen}, y_{tar}) 	
\end{equation}
where \( y_{tar} \) is the target text description, \( x_{gen} \) is the generated image, and \( D_{CLIP} \) calculates the cosine distance between their CLIP space vectors.
  \item \emph{Directional CLIP Loss:} This loss aligns the direction between the embeddings of a pair of reference and target texts with the direction between the embeddings of the reference and generated images. 
\begin{equation}
\label{eq:clip_direction}
 L_{direction} (x_{gen}, y_{tar}; x_{ref}, y_{ref}) = 1 - \frac{\Delta I \cdot \Delta T}{||\Delta I|| ||\Delta T||}	
\end{equation}
where \( \Delta T = E_T(y_{tar}) - E_T(y_{ref}) \) and \( \Delta I = E_I(x_{gen}) - E_I(x_{ref}) \). \( E_I \) and \( E_T \) represent the image and text encoders of CLIP, respectively.
\end{compactitem}
The directional CLIP loss is highlighted for its robustness against mode-collapse and adversarial attacks because it ensures that the perturbations in the image are dependent on the specific target and reference images, thus generating distinct outcomes. This approach helps in producing more diverse and secure image manipulations guided by textual descriptions.

\sstitle{Instructional guidance}
LLM and projectors are fine-tuned on multi-modal edit instructions and pairs of source images and edit results~\cite{li2023instructany2pix}. The input projector maps embeddings of reference images and audio pieces to the LLM's space, while the output projector maps the LLM output back to the multi-modal encoder's space. The target sequence is ``[base] [sub] .. [sub] [gen] .. [gen]'' with [sub] tokens for referenced objects included only if needed.

For instructions like ``add [a]'', ``remove [b]'', ``replace [c] with [d]'', only [a] and [d] have corresponding [sub] tokens. Cross entropy classification loss is applied to the LLM's output. Embeddings of the target sequence are fetched from the LLM's last layer and projected into the multi-modal encoder's space. L2 regression loss is applied between the [base] embedding and source image, and between the [gen] embedding and target images. Especially, image generation is not required as loss is applied directly in the embedding space.

\sstitle{Chain-of-Thought finetuning}
Zhang et al.~\cite{zhang2024tie} formulates image editing as an inpainting task requiring precise masks and text prompts. Given LISA's language generation and segmentation capabilities, GPT-4V-generated reasoning samples are used to fine-tune LISA~\cite{yang2023improved}, known as Fine-tune-CoT~\cite{ho2023large}, to improve its performance in in-context learning.
For instance, an editing prompt like "Place a single vase of flowers and a glass of soda on the table, and also add a bottle of beer" is broken down into simpler steps: (1) Put a glass of soda on the table, (2) Keep just one vase of flowers, and (3) Add a bottle of beer. The assistant locates regions for each task, creating segmentation prompts such as \texttt{[SEG] a glass of soda on the table}.
The model's vocabulary is expanded with the new token \texttt{[SEG]}, used to generate the final segmentation mask and inpainting prompt. These are then applied in the Kandinsky-2.2-decoder-inpaint model~\cite{kirillov2023segment} to produce images accurately reflecting the user's instructions, enhancing image quality and relevance.

\sstitle{Noise combination}
Kim et al.~\cite{kim2022diffusionclip} propose a noise combination method, which allows for the manipulation of multiple attributes within an image through a single sampling process, without the need for fine-tuning new models for each combination of attributes. The approach involves the use of multiple fine-tuned models, each representing different attributes, and combining their effects through a weighted sum of their respective noises~\cite{choi2018stargan,choi2020stargan}.
The process starts with inverting an input image into a latent representation using the forward DDIM process with a pretrained model. This latent representation is then manipulated using the reverse DDIM process, where the noise contributions from each fine-tuned model are combined according to time-dependent weights \(\gamma_i(t)\) that sum to one. This allows the manipulation of the image to reflect multiple attributes simultaneously.
Furthermore, Kim et al.~\cite{kim2022diffusionclip} propose a continuous transition method, where the noise from the original model and a fine-tuned model can be mixed in varying proportions to smoothly transition between the original and manipulated images. This enables precise control over the degree of change during the manipulation, enhancing the versatility of the model for complex image editing tasks.

\sstitle{Additional Supervisions}
Using InstructPix2Pix protocol \cite{brooks2023instructpix2pix}, Chakrabarty et al. \cite{chakrabarty2023learning} add noise to the encoded latent \( z = E(x) \) of an image \( x \), with noise level \( z_t \) increasing over time steps \( t \). A network \( \theta \) is trained to predict this noise given image conditioning \( c_I \) and text instruction conditioning \( c_T \).
For image conditioning, input channels are added to the first convolutional layer, combining \( z_t \) and \( E(c_I) \). The model uses pre-trained weights for the diffusion process, with new channels initialised to zero, and adapts the text conditioning mechanism for edit instructions.
Fine-tuning uses bounding box supervision, setting \( c_I \) to an image with a bounding box, and segmentation mask supervision, setting \( c_I \) to an image with a segmentation mask.

\subsection{Zero-shot}

\sstitle{Tuning-free} 
Masactrl \cite{cao2023masactrl} introduces a tuning-free mechanism for self-attention control in diffusion models to ensure consistency in image synthesis and editing. It employs mutual self-attention to selectively align features across image regions according to the textual prompt, enhancing efficiency without model fine-tuning. The key formula updates attention weights via:
\[
A'_{ij} = A_{ij} \times (1 + \alpha \cdot M_{ij})
\]
where \(A_{ij}\) are original attention weights, \(M_{ij}\) is the mutual attention mask, and \(\alpha\) is a control parameter. This method allows coherent and consistent edits by aligning attention with text prompts.
Similarly, Hertz et al.~\cite{hertz2023prompt} uses cross-attention mechanisms to modify specific areas of an image based on text prompts.

\sstitle{Contrastive Language-Image Pre-training (CLIP)}
Contrastive Language-Image Pre-training (CLIP)~\cite{radford2021learning} is an approach for learning joint representations between text and images. It consists of two models: an image encoder \(f(x)\) and a caption encoder \(g(c)\). CLIP optimizes a contrastive cross-entropy loss to match image-caption pairs. If the image \(x\) aligns with the caption \(c\), the dot product \(f(x) \cdot g(c)\) is high, and vice versa for mismatched pairs.
Given that CLIP provides a similarity score between images and captions, it has been used to steer generative models like GANs~\cite{patashnik2021styleclip,gal2022stylegan}. Nichol et al.~\cite{nichol2022glide} apply this to diffusion models, using a CLIP model for classifier guidance. They adjust the reverse-process mean with the gradient of the dot product of the image and caption encodings, leading to:
\begin{equation}
\hat{\mu}_\theta(x_t | c) = \mu_\theta(x_t | c) + s \cdot \Sigma_\theta(x_t | c) \nabla_{x_t} \left(f(x_t) \cdot g(c)\right),
\end{equation}
where \(s\) is a scaling factor.
Similar to classifier guidance, CLIP must be trained on noised images \(x_t\) for effective gradient derivation. Nichol et al.~\cite{nichol2022glide} use noise-aware CLIP models, noting that while public CLIP models can guide diffusion, their noised CLIP achieves better results without extra techniques like data augmentation. They suggest that using public CLIP impacts quality as noised images are out-of-distribution. Additionally, Patashinik et al. \cite{patashnik2021styleclip} explore manipulating StyleGAN images with text prompts using CLIP methods like Latent Optimization and Direction Mapping, enabling consistent and detailed edits across images.

\sstitle{Inversion-free}
Inversion-based editing methods are slow and struggle with consistency due to their reliance on time-consuming inversion processes and calibration efforts. These methods iteratively calibrate the target branch \(\mathbf{z}_t^{\text{tgt}}\) to match the source branch, but this does not ensure consistency. 
To overcome these issues, Xu et al.~\cite{xu2024inversion} propose the Inversion-Free Image Editing (InfEdit) framework, which uses a dual-branch paradigm but directly calibrates the initial \(\mathbf{z}_0^{\text{tgt}}\) without iterative adjustments. Starting from a random terminal noise \(\mathbf{z}_{\tau_1}^{\text{src}} = \mathbf{z}_{\tau_1}^{\text{tgt}} \sim \mathcal{N}(\mathbf{0}, \mathbf{I})\), the source branch follows the DDCM sampling process without explicit inversion. The framework computes the distance \(\Delta \mathbf{\epsilon}^{\text{cons}}\) and calibrates the target branch prediction accordingly. InfEdit eliminates the need for inversion branch anchors, saving computation time and reducing cumulative errors. It is also compatible with efficient Consistency Sampling using LCMs, enabling effective target image sampling within fewer steps.

\sstitle{Equivariant finetuning}
To address temporal inconsistencies with anchor-based attention, Wu et al.~\cite{wu2023fairy} propose an equivariant finetuning approach using data augmentation. If an input frame \( I^t \) differs from \( I^{t-1} \) only in camera position, the output frames \( \hat{I}^t \) and \( \hat{I}^{t-1} \) should differ similarly.  The strategy involves applying a random affine transformation \( g : \mathcal{I} \rightarrow \mathcal{I} \) to both the original and edited image pairs (\( I, I' \)), using random rotations, translations, scaling, and shearing. The images are then cropped and resized. The model is fine-tuned with the transformed images \( g(I) \) and \( g(I') \), making it equivariant to these transformations. This finetuning method significantly improves temporal consistency.

\sstitle{Training-free}
Santos et al.~\cite{santos2024pix2pix} propose training-free and tuning-free image editing by using pre-trained models: Stable Diffusion~\cite{rombach2022high} for diffusion, BLIP~\cite{li2022blip} for captioning, and Phi-2~\cite{gunasekar2023textbooks} for generating edit directions.
First, the initial image is captioned using BLIP, and the corresponding noise vector is obtained through DDIM Inversion. 
Next, the edit direction embedding is created using two captions: one before editing and one after. The difference between their embeddings, calculated with the CLIP model, guides the editing.
Finally, Stable Diffusion generates the edited image, using the noise vector, caption, and edit direction embedding, enabling text-based image modifications without further training~\cite{parmar2023zero}.

\subsection{Weak Supervision}

Bodur et al.~\cite{bodur2023iedit} propose a method for fine-tuning the Latent Diffusion Model (LDM)~\cite{rombach2022high} for image editing with a weakly-supervised dataset. The process involves encoding the source image \( x_1 \) into the latent space \( z_1 := \mathcal{E}(x_1) \) and adding noise \( \epsilon_1 \) to create a noisy image $z_t = \sqrt{\alpha_t} z_1 + \sqrt{1 - \alpha_t} \epsilon_1,
$,
where \( \epsilon_1 \sim \mathcal{N}(0,1) \) and \( \alpha_t \) is the Gaussian transition sequence~\cite{song2021denoising}.
The target image \( x_2 \) is encoded as \( z_2 := \mathcal{E}(x_2) \), and the ground truth noise $\epsilon_2 = \frac{z_t - \sqrt{\alpha_t} z_2}{\sqrt{1 - \alpha_t}}.
$
The objective is to minimize the L2 loss between \( \epsilon_2 \) and the noise predicted by the network \( \epsilon_\theta \):
\begin{equation}
\mathcal{L}_{\text{paired}} = \mathbb{E}_{\mathcal{E}(x), y_2, \epsilon_2, t} \left[ || \epsilon_2 - \epsilon_\theta(z_t, t, \tau_\theta(y_2)) ||^2 \right].
\end{equation}

To align the generated image with the edit prompt, a global CLIP loss is introduced~\cite{patashnik2021styleclip}:
\[
\mathcal{L}_{\text{global}}( \hat{x}_1, y_2 ) = D_{\text{CLIP}}( \hat{x}_1, y_2 ),
\]
where \( \hat{x}_1 \) is decoded from $
\hat{z}_1 = \frac{z_t - \sqrt{1 - \alpha_t} \epsilon_\theta(z_t, t, \tau_\theta(y_2))}{\sqrt{\alpha_t}}.
$
The losses are weighted inversely to the noise level \( t \), as in Fan et al. \cite{fan2023target}, for balanced performance:
\[
\left( 1 - \frac{t}{T} \right) \mathcal{L}_{\text{global}} + \frac{t}{T} \mathcal{L}_{\text{paired}},
\]
where \( T \) is the maximum number of noise steps.

\subsection{Reinforcement Learning with Human Feedback (RLHF)}

\sstitle{Interactive dialogue}
Jiang et al.~\cite{jiang2021talk} propose a dialog-based system, which provides natural language feedback. The feedback function \( \text{feedback}_t = \text{Talk}(\text{feedback}_{t-1}, r, s, e_r, h) \) handles user requests (\(r\)), system state (\(s\)), editing encoding (\(e_r\)), and editing history (\(h\)). Feedback includes checking if the edit meets user expectations, offering alternative options, and requesting more instructions.
For user requests, the system uses a language encoder \(E\) to generate \(e_r = E(r)\). The encoding directs further dialogue based on request type, attribute of interest, editing direction, and change degree. User requests are classified into three types: specifying a target degree, indicating a relative degree of change, and describing the attribute without specifying the degree of change. The system employs templates to generate and train these requests.

\sstitle{Reward learning}
Zhang et al.~\cite{zhang2024hive} learns a reward function \( R_{\Phi}(\hat{x}, c) \) that mimics human preferences on evaluating the quality of image edits made by a diffusion model. This reward function takes into account the original input image, the text instruction condition \( c = [c_I, c_E] \), and the edited image \( \hat{x} \) to output a scalar value. The reward model architecture utilizes vision-language models like BLIP~\cite{li2022blip} for multimodal encoding to create an image-grounded text encoder that produces a multimodal embedding from the image and text instruction.
The training of the reward function involves using a dataset \( D_{human} \) of images ranked by human annotators based on the quality of the edit. The pairwise loss function for the reward model is defined using the Bradley-Terry model~\cite{ouyang2022training}:
\begin{equation}
 L_{RM}(\Phi) = - \sum_{\hat{x}_i \succ \hat{x}_j} \log \left( \frac{exp(R_{\Phi}(\hat{x}_i, c))}{\sum_{k=i,j} exp(R_{\Phi}(\hat{x}_k, c))} \right),
 \end{equation}
where \( \hat{x}_i \succ \hat{x}_j \) indicates a human preference for the edited image \( \hat{x}_i \) over \( \hat{x}_j \).

\sstitle{RLHF-based fine-tuning} 
Zhang et al. \cite{zhang2024hive} introduced a method to enhance image editing by incorporating human feedback into generative model training, specifically models like InstructPix2Pix. HIVE's three main stages involve supervised training with a large instructional dataset, developing a reward model from human rankings, and fine-tuning using this feedback to align edits with user preferences. 
The training uses over one million examples to enhance dataset diversity, followed by a reward model built from human rankings of edited outputs. This model is then used to fine-tune the editing model for better quality edits. Evaluations on both synthetic and real-world datasets showed that HIVE outperforms InstructPix2Pix, improving user satisfaction and demonstrating the value of integrating human feedback.

\section{Applications}
\label{sec:application}

\subsection{Extended Image Editing}

\sstitle{Style editing}
StyleBooth~\cite{han2024stylebooth} unifies text and exemplar-based style editing by mapping inputs into a shared hidden space through a trainable matrix \( W \), integrating both vision and text instructions. Using identifiers like "\(<\text{style}>\)" and "\(<\text{image}>\)," the system follows diverse instructions and balances them with a Scale Weights Mechanism.
A high-quality dataset is created by generating style images and de-styling them into plain images. Iterative refinement through style-tuning and filtering preserves content structure while enhancing image quality and usability. Efficient style tuners trained for each style improve model accuracy.
Multimodal instructions involve encoding style images with a CLIP encoder \( C_I \) and integrating these features into text space with \( W \). This forms the final instruction \( h = F_{\text{insert}}(h_T, h_V) \), allowing the model to interpret and apply style instructions from both text and visuals. The training objective ensures the model accurately follows these instructions by minimizing differences between generated and latent space embeddings. Fine-tuning with style and de-style tuners further enhances fidelity and consistency in editing.

\sstitle{Fashion editing}
Pernuvs \cite{pernuvs2023fice} propose FICE, which aims to edit a given fashion image \( I \in \mathbb{R}^{3 \times n \times n} \) according to an appearance-related text description \( t \) and produce a corresponding output image \( I_f \in \mathbb{R}^{3 \times n \times n} \) that closely aligns with the semantics of \( t \). The synthesized image \( I_f \) should preserve the pose, identity, and other appearance characteristics of the subject in \( I \), while ensuring edits are localized to desired fashion items and maintaining realistic clothing appearances.
FICE employs a three-stage procedure: Initialization, Constrained GAN Inversion, and Image Stitching. In the Initialization stage, a latent code \( w \) is generated from the input image \( I \) using a GAN inversion encoder \( E \)~\cite{tov2021designing}. This code approximates the original appearance of \( I \) through the pre-trained GAN generator \( G \).
In the Constrained GAN Inversion stage, \( w \) is optimized to match \( t \) while preserving visual characteristics, ensuring the synthesized image \( I_g = G(w) \) contains the semantics of \( t \).
The Image Stitching stage combines the optimized GAN-synthesized image \( I_g^* \) with \( I \) to ensure identity preservation, producing the final output image \( I_f \).
FICE relies on several models: the pre-trained StyleGAN generator \( G \)~\cite{karras2020analyzing}, the CLIP model \( C \) for semantic editing~\cite{radford2021learning}, DensePose \( D \) for pose preservation~\cite{guler2018densepose}, and a segmentation model \( S \) for artifact-free results~\cite{chen2017rethinking}.
The final optimization objective combines multiple loss functions: \( L_{\text{clip}} \) for semantic content, \( L_{\text{pose}} \) for pose preservation, \( L_{\text{reg}} \) for latent code regularisation~\cite{abdal2019image2stylegan,abdal2020image2stylegan++,tov2021designing}, \( L_{\text{im}} \) for image consistency, and \( L_{\text{head}} \) for head preservation:
\[
L = \lambda_{\text{clip}} L_{\text{clip}} + \lambda_{\text{pose}} L_{\text{pose}} + \lambda_{\text{reg}} L_{\text{reg}} + \lambda_{\text{im}} L_{\text{im}} + \lambda_{\text{head}} L_{\text{head}}
\]
where \( \lambda_{\text{clip}}, \lambda_{\text{pose}}, \lambda_{\text{reg}}, \lambda_{\text{im}}, \lambda_{\text{head}} \) are balancing weights.
The final image \( I_f \) is obtained by:
\[
I_f = S_{\text{head}}(I) \odot I + (1 - S_{\text{head}}(I)) \odot I_g^*
\]
with \( I_g^* = G(w^*) \), where \( w^* \) is the optimized latent code.

Wang et al.~\cite{wang2024texfit} propose a two-stage approach called TexFit. 
First, the Editing Region Location Module (ERLM) uses text prompts to locate hidden editing regions in an image~\cite{jiang2022text2human}. A text prompt \( P \) and fashion image \( \mathbf{x_0} \) produce a region mask \( M \). The ERLM processes \( \mathbf{x_0} \) and the text embedding \( \mathbf{f_p} \) from a CLIP model, using an encoder \( E \) and a decoder \( D \).
In the second stage, the masked image \( \mathbf{x_m} = (1 - M) \odot \mathbf{x_0} \) guides Latent Diffusion Models (LDMs). The mask condition \( z_t' \) includes \( z_t \), \( m \), and \( z_m \). Using a classifier-free guidance technique~\cite{ho2022classifier}, predictions combine unconditional and conditional elements: \( \tilde{\epsilon_t} = \epsilon_\theta(z_t', t, \mathbf{c_\emptyset}) + w \cdot (\epsilon_\theta(z_t', t, c) - \epsilon_\theta(z_t', t, \mathbf{c_\emptyset})) \). The final edited image \( \tilde{\mathbf{x}} \) combines the edited and original images: \( \tilde{\mathbf{x}} = M \odot \mathbf{x_e} + (1 - M) \odot \mathbf{x_0} \).

The GaussianVTON framework~\cite{chen2024gaussianvton} edits 3D scenes for virtual try-on using 3D Gaussian Splatting and 2D VTON models~\cite{morelli2023ladi}. It processes a reconstructed 3D scene with captured images, camera poses, and calibration data.
The Editing Recall Reconstruction (ERR) strategy~\cite{chen2024gaussianvton} renders the entire dataset during editing, using the original images \( I^0_v \) for conditioning inputs and the target garment \( X \). ERR ensures consistency by editing all images before updating the dataset.
A three-stage refinement strategy enhances quality: Stage-1 aligns facial features using FaceMesh, Stage-2 segments garment regions and detects errors~\cite{kirillov2023segment}, and Stage-3 uses denoising models like DDNM~\cite{wang2023zeroshot} to refine the final rendering.

\sstitle{Scene Editing}
Yildirim et al.~\cite{yildirim2023inst} propose a data construction pipeline for instruction-based scene editing. It starts with the selection of objects from scene graphs generating text prompts. Objects are categorised into two groups:
(1) \emph{Removable Objects}: These can be uniquely referred to and removed, such as a man, boat, or kite.
(2) \emph{Non-removable Objects}: These appear in referring expressions but are not sensible to remove, such as walls or the sky.
Objects with bidirectional relationships can be removed, while those with unidirectional relationships cannot~\cite{hudson2019gqa}. Items corresponding to multiple objects (e.g., apples, bikes) and objects that are too large or too small are filtered out. Implicit parts of objects and items worn by objects are also excluded.
Relations in scene graphs are simplified, and segmentation masks are obtained using Detectron2 and Detic frameworks~\cite{zhou2022detecting}. These masks help identifying objects for removal using a method called CRFill~\cite{zeng2021cr} for high-quality inpainting results. Textual prompts are generated based on object relations and attributes, with random attributes used for data augmentation.

\sstitle{Face Editing}
TediGAN by Xia et al. \cite{xia2021tedigan} is a framework for generating and manipulating facial images using text descriptions. It employs a two-step GAN-based approach to align textual prompts with facial features, providing control over attributes like hair color, age, and expression. TediGAN maps text descriptions into a joint latent space using the StyleGAN architecture. Specifically, latent inversion transforms real images into latent codes in StyleGAN's \( \mathcal{W} \)-space for manipulation. The optimization goal is given by:
\[
L_{\text{recon}} = \|x - G(E(x))\|_2^2 + \lambda_1 \|F(x) - F(G(E(x)))\|_2^2 - \lambda_2 \mathbb{E}[D(G(E(x)))]
\]
where \( E(x) \) is the encoding function, \( G \) is the generator, \( F \) denotes feature extraction, and \( D \) is the discriminator. This setup allows for fine-grained control over facial attributes while balancing reconstruction fidelity and semantic manipulation.

Talk-to-Edit \cite{jiang2021talk} introduces a dialog-based system for iterative control over facial edits, using the CelebA-Dialog dataset to modify facial attributes incrementally based on user feedback. Dialog instructions are parsed into semantic vectors, with a conditional GAN applying changes. The mapping function \( M \) translates dialog into attribute adjustments:
\[
z_{\text{edit}} = z + s \times M(z)
\]
where \( z \) is the initial latent code and \( s \) controls editing strength. This iterative process ensures personalized and coherent facial editing.
On the other hand, Yue et al. \cite{yue2023chatface} encode the input image into a semantic latent space \( z \in \mathbb{R}^{512} \), deriving a noise latent code \( x_T \). A residual mapping network provides manipulation directions:
\[
z_{\text{edit}} = z + s \times \text{Mapping}(z).
\]
Linear interpolation between \( z_{\text{edit}} \) and \( z_0 \) ensures temporal alignment:
\[
z_t = \text{Lerp}(z_{\text{edit}}, z_0; \nu)
\]
where \( \nu = t/T \). Editing is achieved through three types of losses: reconstruction loss, face identity loss using the ArcFace network, and CLIP direction loss, enhancing control over facial attributes while preserving desired features.

\sstitle{Chart Editing}
Yan et al.~\cite{yan2024chartreformer} categorise chart edits into style, layout, format, and data-centric. Style edits change visual attributes like colours and fonts for readability~\cite{liu2023matcha,han2023chartllama}. Layout re-composition adjusts elements like axes and grids to improve systematic representation. Format conversion changes chart types (e.g., line to bar charts) to highlight different data aspects. Data-centric modifications involve precise data manipulations, enabling custom views.

ChartReformer~\cite{yan2024chartreformer} addresses chart editing by de-rendering charts into data and visual attributes for accurate reconstruction. The dataset is created from paired chart images using real-world data sources~\cite{davila2021icpr}. Custom software generates diverse chart images stored in JSON for easy adjustments. Edit pair synthesis ensures visual consistency while updating data. The pipeline pre-trains ChartReformer on a large dataset to learn mappings between attributes and data, then fine-tunes on paired images and prompts to adjust charts accurately~\cite{lee2023pix2struct,liu2023deplot}. The goal is to enhance real-world plotting by suggesting JSON repair and default parameters for incomplete predictions.

\sstitle{Remote sensing images}
After training, SinDDM~\cite{kulikov2023sinddm} can be used with CLIP~\cite{radford2021learning} for remote sensing image editing~\cite{han2024exploring}. The CLIP loss \( L_{CLIP} \) measures the cosine distance between image and text embeddings:
\[ L_{CLIP} = - \frac{f_I(\hat{x}_0^s) \cdot f_T(\text)}{\| f_I(\hat{x}_0^s) \| \cdot \| f_T(\text) \|} \]

The prompts can include both ROI limiting the editing area and corresponding text, constraining affected regions spatially. The original image \( x_0^s \) is added with the same noise and blended with the masked image to continue the diffusion process.

For prompt-based fine-tuning, selecting the right prompt is crucial. Different text prompts can significantly affect the image editing result, as shown by comparing ``Large Fire'' with ``Heavily Burning''. Prompt Ensembling (PE)~\cite{han2024exploring} is used to generate robust text guidance by combining multiple prompts with the same semantics. Utilizing GPT for generating high-quality prompts ensures diversity and better image editing outcomes. The embeddings of the prompts are averaged as the guidance for the image editing model.

\sstitle{Iterative Editing}
Joseph et al.~\cite{joseph2024iterative} propose a method called Multi-granular Image Editing, which allows diffusion models to apply spatially constrained edits~\cite{wang2023imagen}. It treats the diffusion model as an energy-based model (EBM)~\cite{liu2022compositional}, enabling selective gradient updates to specific regions.
An EBM models data likelihood using an energy function \( E_\psi(\cdot) \), mapping latent embedding \( z = \mathcal{E}(x) \) to a scalar value. The probability density for \( z \) is:
\[ p_\psi(z) = \frac{\exp(-E_\psi(z))}{\int_z \exp(-E_\psi(z)) \, dz} \]
Langevin Sampling~\cite{welling2011bayesian} is used to update \( z \) iteratively:
\[ z_t = z_{t-1} - \frac{\lambda}{2} \nabla_z E_\psi(z_{t-1}) + \mathcal{N}(0, \omega_t^2 I) \]
This aligns with the latent diffusion model's noise predictions, allowing control by zeroing out gradients for non-interest regions:
\[ z_{t-1} = z_t - \mathbf{m} \ast \epsilon_\theta(z_t, t) + \mathcal{N}(0, \sigma_t^2 I) \]
The method involves passing an image \( I^e \) through a pretrained VQ-VAE encoder \( \mathcal{E} \) to obtain \( z_{\text{img}}^e \), stacking it with \( z_t \), and denoising over \( T \) iterations. Initial approaches revealed noise accumulation, so latent iteration was proposed.

In the iterative multi-granular image editor, \( z_{\text{init}} \sim \mathcal{N}(0, I) \) is initialized. For each edit instruction \( y_i \), \( z_{\text{img}}^e \) is set to \( \mathcal{E}(I_0) \) for the first instruction or the previous latent for subsequent ones. \( z_{\text{init}} \) and \( z_{\text{img}}^e \) are concatenated, then denoised for each \( t \) from \( T \) to 0. After denoising, \( z_0 \) is decoded and appended to the set of edited images, then reused for subsequent instructions.

\sstitle{Multi-instruction Editing}
Guo et al.~\cite{guo2023focus} propose a method for multi-instruction image editing.
First, masks are extracted for each instruction using segmentation capabilities of large-scale diffusion models~\cite{wu2023diffumask}. Cross-attention maps identify precise locations, which are processed iteratively to enhance contrast and derive masks for areas of interest~\cite{agarwal2023star,xie2023boxdiff}.
Cross-condition attention modulation confines each instruction within its mask to reduce interference among instructions~\cite{chefer2023attend}. The attention map is adjusted to maintain focused attention for each sub-instruction using the function:
\[ A'_{t, \text{ins}} = \text{softmax}\left( \frac{\left( \mathcal{X} + \Delta \mathcal{X} \right) \odot \mathcal{M} + \mathcal{Y} \odot \left(1 - \mathcal{M}\right)}{\sqrt{d}} \right) \]
where \( \mathcal{X} \) and \( \mathcal{Y} \) are the query and key matrices respectively, and \( \Delta \mathcal{X} \) enhances attention values.
For fine-grained editing, mask-guided disentangle sampling isolates the editing area from irrelevant regions. Combined masks are upsampled to match the latent space resolution.
This sampling method is applied in 75\% of the diffusion steps, with standard sampling for the remaining 25\%, enhancing precision and coherence in multi-instruction image editing~\cite{guo2023focus}.

\sstitle{Multi-object Editing}
Yang et al.~\cite{yang2024objectaware} propose a method using Stable Diffusion~\cite{rombach2022high} and DDIM Inversion for multi-object edits. It represents an editing task as \(\langle I_o, P_o, P_t \rangle\), where \(I_o\) is the original image, \(P_o\) is its description, and \(P_t\) is the target prompt. The goal is to modify \(I_o\) to produce \(I_t\), aligning with \(P_t\).
To achieve this, \(I_o\) is inverted to \(I_{\text{noise}}\) using DDIM Inversion guided by \(P_o\), then denoised to generate \(I_t\) guided by \(P_t\). Editing pairs \(\langle O_o, O_t \rangle\) are defined, where \(O_o\) and \(O_t\) represent corresponding objects in \(P_o\) and \(P_t\). Each pair may require a distinct optimal inversion step~\cite{tumanyan2023plug}.
Each inversion step \( i \) produces an edited image \( I_t^i \), creating a set of candidate images \(\{ I_t^i \}\). The best candidate \( I_t^{i^*} \) that aligns with the target prompt \( P_t \) and preserves non-editing regions is selected as the optimal inversion step \( i^* \).
To automate this process, a mask generator extracts the editing region \( M_e \) and the non-editing region \( M_{ne} \) from the original image \( I_o \)~\cite{liu2023grounding,kirillov2023segment,couairon2023diffedit,cao2023masactrl}. The quality of candidate images is evaluated using two metrics: \( S_e \) for alignment with the target prompt, calculated with CLIP scores~\cite{hessel2021clipscore}, and \( S_{ne} \) for similarity with the non-editing region, using negative mean squared error. 
Parallelisation and splicing strategies speed up candidate image generation.
Yang et al.~\cite{yang2024objectaware} handle concept mismatch and poor editing by processing each editing pair \(\{(O_o, O_t)_k\}\) independently. Guided prompts \(\{ P_t^k \}\) are used to find the optimal inversion step \( i_k^* \) for each pair~\cite{radford2021learning}. During reassembly, each region starts from its optimal inversion step, and denoising continues up to the reassembly step \( i_r \), set to 20\% of total inversion steps~\cite{xu2023restart,meng2023distillation,song2023consistency}. Re-inversion smooths edges and enhances fidelity, followed by a final denoising step guided by \( P_t \).

\sstitle{Local editing}
Given an RGB image \( I_G \) and a textual instruction \( T_I \), the goal is to locate and edit specific regions while preserving non-edited areas. Inspired by Text2LIVE~\cite{bar2022text2live}, Li et al.~\cite{li2024zone} extracts an edited layer \( I_L \) and composites it over \( I_G \) without using user-defined masks or complex prompts.
Traditional methods require users to specify objects to edit with prompts or masks, which is often unintuitive~\cite{avrahami2022blended,hertz2023prompt,parmar2023zero,couairon2023diffedit}. ZONE~\cite{li2024zone} edits objects based on instructions like ``make her old'', using token-aware models that link text to spatial structures. The initial mask \( M_b \) and image \( I_{sty} \) are refined using SAM~\cite{kirillov2023segment} and a Region-IoU (rIoU) scheme for accurate segmentation. The refined mask \( M_f \) and edited layer \( I_L' = I_{sty} \odot M_f \) are created. An FFT-based edge smoother addresses over-edit issues, creating edge-dilated layers \( I_{L,d} \) and \( I_{G,d} \). The final mask \( M_f^* \) is derived, and the final edited result \( I_C \) is obtained by combining \( I_G \) and \( I_L \).

\sstitle{Visual instruction}
Prior textual inversion methods use an image reconstruction loss to capture the essence of the concept for synthesis in new contexts but lack pixel-level detail for editing~\cite{kawar2023imagic}. Instead, Nguyen et al.~\cite{nguyen2023visual} use a pre-trained text-conditioned image editing model~\cite{brooks2023instructpix2pix}, which avoids additional fine-tuning. Given two images \( \{x, y\} \) representing ``before'' and ``after'' states of an edit \( c_T \), the goal is to recover image \( y \)~\cite{radford2021learning}. The instruction \( c_T \) is optimized based on the pair \( \{x, y\} \).

Relying only on image reconstruction may lead to learning the edited image \( y \) rather than the editing instruction. The CLIP embedding helps in determining the editing direction~\cite{patashnik2021styleclip,gal2022stylegan,parmar2023zero}. For a pair \( \{x, y\} \), the image editing direction \( \Delta_{x \to y} \) is:
\[
\Delta_{x \to y} = \mathcal{E}_{\text{clip}}(y) - \mathcal{E}_{\text{clip}}(x)
\]
The instruction \( c_T \) is aligned with this direction by minimizing the cosine distance:
\[
\mathcal{L}_{\text{clip}} = \cos(\Delta_{x \to y}, c_T)
\]
For a pair \( \{x, y\} \), visual prompting is formulated as an instruction optimization using image reconstruction loss and CLIP loss. The method computes the mean difference \( \Delta_{x \to y} \) for all examples and optimizes \( c_T \) during each step. Once learned, \( c_T \) can edit a new image \( x_{\text{test}} \) into \( y_{\text{test}} \). This approach allows combining \( c_T \) with additional information, providing fine-grained control over edits.

\subsection{3D Editing}
Michel et al.~\cite{michel2022text2mesh} propose a methodology for applying style to 3D meshes based on text prompts using a Neural Style Field (NSF) network. This method facilitates high-resolution styling with enhanced detail through spectral bias techniques~\cite{rahaman2019spectral,tancik2020fourier}, improving the mesh's structural and visual quality. The process starts with a fixed input mesh \(M\) of vertices \(V\) and faces \(F\), which is stylised into \(M^S\) through the influence of a text prompt \(t\).
The NSF generates a style field for each vertex by employing Multi-Layer Perceptrons (MLP). Positional encoding \(\gamma(p)\) is applied to each vertex, defined as:
\begin{equation}
\gamma(p) = [\cos(2\pi Bp), \sin(2\pi Bp)]^T
\end{equation}
where \(B\) is a Gaussian matrix determining the frequency of style changes controlled by \(\sigma\).
These encoded positions are input to the MLP \(N_s\), which branches into \(N_d\) for displacement along the vertex normal and \(N_c\) for colouring, both within specific constraints to maintain the content's integrity~\cite{chen2019learning}. The style is visualised using a differentiable renderer, and the network weights are optimised by matching the stylised mesh renderings to the target text via CLIP~\cite{radford2021learning}. 
Then, the method leverages the CLIP model to guide the neural optimisation of stylising 3D meshes via text-based correspondence. The method involves rendering the stylised mesh \(M^S\) and its geometric variant \(M^S_{disp}\) from multiple views, and for each view, generating two 2D projections. These projections are augmented and embedded into CLIP space, producing averaged embeddings across all views.
The optimisation targets a loss function based on the cosine similarity between these averaged embeddings and the target text embedding. This approach ensures the application of global and local augmentations to maintain view consistency and encourage meaningful changes in geometry and color, aligning the 3D mesh's appearance closely with the specified text description.

Fan et al.~\cite{fan2023target} propose a Cyclic Manipulation GAN (cManiGAN) framework aiming to generate semantically accurate images from a reference image \( I_r \) and a textual instruction \( T \). It consists of an image editor \( G \), divided into a localiser \( L \) and an image in-painter \( P \).
The reasoning module \( R \) is trained using the objective function:
\begin{equation}
\mathcal{L}_R = \mathcal{L}_{\text{2s2s}} \left( R(T^o_r \oplus T^o_t \oplus T_{\text{loc}}), T \right) + \mathcal{L}_{\text{2s2s}} \left( R(T^o_r \oplus T \oplus T^o_t), T^o_t \right) 
\end{equation}
where \( T^o_r \) and \( T^o_t \) are original and target instructions, \( T_{\text{loc}} \) is location-based, and \( \oplus \) denotes concatenation~\cite{shetty2018adversarial}.
The localiser \( L \) identifies the target object/location in \( I_r \) based on \( T \), generating a binary mask $
M = L(I_r, T_{\text{loc}}) 
$.
The in-painter \( P \) then generates the image \( I_g \) using the text feature \( f^T_{\text{how}} \) and masked input \( (1 - M) \odot I_r \)~\cite{raffel2020exploring}:
\begin{equation}
I_g = P(f^T_{\text{how}}, (1 - M) \odot I_r) 
\end{equation}
The reconstruction loss \( \mathcal{L}_{\text{rec}} \) ensures consistency of unmasked regions:
\begin{equation}
\mathcal{L}_{\text{rec}} = \mathcal{L}_{\text{MSE}} \left( (1 - M) \odot I_r, (1 - M) \odot I_g \right) 
\end{equation}
A classification loss \( \mathcal{L}_{\text{CE}} \) ensures semantic accuracy~\cite{iizuka2017globally,lu2019vilbert,li2020lightweight}.
The localiser \( L \) also includes classification losses for masked and unmasked regions:
\begin{equation}
\mathcal{L}^L_{\text{in}} = \mathcal{L}_{\text{CE}} \left( \text{MLP}(E((1 - M) \odot I_r)), y_{\text{in}}^r \right) 
\end{equation}
\begin{equation}
\mathcal{L}^L_{\text{out}} = \mathcal{L}_{\text{BCE}} \left( \text{MLP}(E((1 - M) \odot I_r)), y_{\text{out}} \right) 
\end{equation}
The cyclic-consistent training scheme allows \( T' \) to be inferred from \( \hat{T} = T_{\text{loc}} \oplus T^o_r \oplus T^o_t \) with swapped labels, providing pixel-level supervision and ensuring \( I_g \) is consistent with \( T \)~\cite{odena2017conditional,johnson2016perceptual}.

Sella et al.~\cite{sella2023vox} propose another method for editing 3D objects using multiview images and text prompts. Initially, the object is represented with a grid-based volumetric representation. 
The volumetric representation uses a 3D grid \( G \), where each voxel holds a 4D feature vector~\cite{sun2022direct}. The object's geometry is modeled with a feature channel for spatial density values, passed through a ReLU nonlinearity~\cite{karnewar2022relu}, and three additional channels representing appearance, mapped to RGB colors via a sigmoid function. Unlike recent neural 3D scene representations, this approach avoids view-dependent appearance effects to reduce artifacts. Volumetric rendering is performed as in NeRF~\cite{mildenhall2021nerf}, using images and camera poses to learn a grid \( G_i \).
Text-guided object editing is performed by optimizing \( G_e \), a grid initialized from \( G_i \). The optimization combines a generative component guided by the text prompt and a pullback term to maintain initial values. A Score Distillation Sampling (SDS) loss~\cite{lin2023magic3d}, applied over Latent Diffusion Models (LDMs), encourages the feature grid to match the desired edit. At each iteration, noise \( \epsilon_t \) is added to a generated image \( x \), with gradients computed as \( \nabla_x \mathcal{L}_{SDS} = w(t) (\epsilon_t - \epsilon_{\phi}(x_t, t, s)) \). An annealed SDS loss decreases the maximal time-step, focusing on high-frequency information.

To avoid over-fitting, a volumetric regularisation term is introduced: \( \mathcal{L}_{reg3D} = 1 - \frac{Cov(f_i^{\sigma}, f_e^{\sigma})}{\sqrt{Var(f_i^{\sigma}) Var(f_e^{\sigma})}} \). This term ensures correlation between density features of the input grid \( f_i^{\sigma} \) and the edited grid \( f_e^{\sigma} \).
The refinement step uses cross-attention layers to create a volumetric binary mask \( M \) that marks voxels for editing~\cite{hertz2023prompt,brooks2023instructpix2pix}. The refined grid \( G_r \) is obtained by merging \( G_i \) and \( G_e \) as \( G_r = M \cdot G_e + (1 - M) \cdot G_i \). Cross-attention maps, elevated to a 3D grid, guide segmentation through energy minimization. This involves a unary term penalizing label probability disagreements and a smoothness term penalizing large pairwise color differences. The smoothness term, computed from local color differences in \( G_e \), is given by \( w_{pq} = \exp \left( -\frac{(c_p - c_q)^2}{2\sigma^2} \right) \). The energy minimization problem is solved via graph cuts~\cite{boykov2001fast}, resulting in high-quality segmentation masks.

\sstitle{Dynamic scaling}
Kamata et al.~\cite{kamata2023instruct} propose Instruct 3D-to-3D, a method for converting a source 3D model into a new target model based on text instructions using InstructPix2Pix~\cite{brooks2023instructpix2pix}. The process begins with initializing the target model with the source model. A target image \( I_{tgt} \) is rendered from a random viewpoint, encoded by StableDiffusion to obtain the latent feature \( L_{tgt} \). Noise \( \epsilon \) is added to create a noisy latent \( x_t \), which, along with the source image \( I_{src} \) and text \( y \), is input into InstructPix2Pix. The noise is estimated as:
\begin{multline*}
\tilde{\epsilon}_{\phi}(x_t; y, I_{src}, t) = \epsilon_{\phi}(x_t; \emptyset, \emptyset, t) + s_I (\epsilon_{\phi}(x_t; \emptyset, I_{src}, t) - \epsilon_{\phi}(x_t; \emptyset, \emptyset, t)) \\ + s_T (\epsilon_{\phi}(x_t; y, I_{src}, t) - \epsilon_{\phi}(x_t; \emptyset, I_{src}, t))
\end{multline*}
Here, \( s_I \) and \( s_T \) are hyper-parameters for fidelity to the source image and text. The target model is updated by:
\[
\nabla_{\theta_{tgt}} \mathcal{L}_{SDS} = \mathbb{E}_{t, \epsilon} \left[ w(t) (\tilde{\epsilon}_{\phi}(x_t; y, I_{src}, t) - \epsilon) \frac{\partial L_{tgt}}{\partial \theta_{tgt}} \right]
\]
DVGO (Dynamic Voxel Grid Optimization)~\cite{sun2022direct} is used for fast 3D-to-3D conversions, maintaining a 3D voxel grid. The number of voxels starts at \( N \) and is dynamically scaled to \( N / 2^l \) during conversion, then gradually returned to \( N \). This method allows for gradual global and detailed structural changes~\cite{lombardi2019neural}.

\sstitle{Iterative dataset update}
Instruct-NeRF2NeRF~\cite{haque2023instruct} starts with a reconstructed NeRF scene and a dataset of calibrated images. Using text instructions, the model is iteratively fine-tuned to produce an edited NeRF by alternating updates between the dataset images and NeRF.

InstructPix2Pix~\cite{brooks2023instructpix2pix} is used to edit each image with inputs: an input conditioning image \( c_I \), a text instruction \( c_T \), and a noisy input \( z_t \). The original image \( I_v^0 \) is updated iteratively, ensuring the diffusion model prevents drift~\cite{meng2022sdedit}.
The Iterative Dataset Update (Iterative DU) involves rendering images from NeRF, updating them via the diffusion model, and refining NeRF. Starting with captured images \( I_v^0 \) from various viewpoints \( v \), a series of updates maintains a mixture of old and new information~\cite{nguyen2022snerf}.
Over time, images converge on a consistent depiction of the edited scene. Iterative DU retains edited images across NeRF updates, aligning with the score distillation sampling (SDS) loss from DreamFusion~\cite{poole2023dreamfusion}, improving stability and efficiency.

\sstitle{Multimodal signals}
Sabat et al.~\cite{sabat2024nerf} propose NeRF-Insert, a method to improve Neural Radiance Fields (NeRF) by iteratively updating datasets~\cite{haque2023instruct}. Users define a 3D inpainting region with masks or meshes, rendered from various views to create inpainting masks. An additional loss term ensures edits stay within this region, enhancing scene quality~\cite{shi2024mvdream,zhu2024hifa}.
The method involves replacing training images with inpainted versions and uses a visual hull field to define the 3D region efficiently. NeRFs are optimized using 2D supervision and a spatially constraining loss term to maintain the original NeRF's density and color outside the mask.

\sstitle{Iterative refinement}
Huang et al.~\cite{huang2024blenderalchemy} propose a method refining Blender 3D programs by decomposing the initial state into a base state \( S_{base} \) and programs \(\{ p_0^{(1)}, p_0^{(2)}, \ldots, p_0^{(k)} \}\):
\[ S_{init} = F \left( \{ p_0^{(i)} \}_{i=1}^k, S_{base} \right) \]
For specific tasks, it decomposes into \( S_{base} \) and a script \( p_0 \):
\[ S_{init} = F(\{p_0\}, S_{base}) \]
The goal is to find an edited version \( p_1 \) that aligns with user intention \( I \). An evaluator \( V(S_1, S_2, I) \) helps refine the program iteratively.
Edits can be minor tweaks or major changes, balancing restrictions between the neighborhood of \( p_{best} \) and the program space \( \mathcal{P} \).
Textual user intentions are enhanced with text-to-image understanding, using generated images to guide edits for accurate results in Blender.

\subsection{Audio Editing}

Wang et al.~\cite{wang2023audit} presents AUDIT, a method for implementing audio editing based on human instructions. The method focuses on generating triplet data (instruction, input audio, output audio) for each editing task. The method has three main advantages: using triplet data ensures high edit quality; input audio as a condition helps preserve unedited parts; and using human instructions as text input makes it practical for real-world scenarios.

Firstly, the system includes an autoencoder that projects input mel-spectrograms to a low-dimensional latent space and reconstructs them back. It features a text encoder for input instructions, a diffusion network for editing in the latent space, and a vocoder for waveform reconstruction. The autoencoder, comprising an encoder \( E \) and a decoder \( G \), transforms mel-spectrograms \( x \) into latent representations \( z \) and reconstructs \( \hat{x} \). A variational autoencoder (VAE) is used, trained with \( L_1 \) and \( L_2 \) reconstruction loss~\cite{isola2017image}, Kullback-Leibler loss \( L_{KL} \)~\cite{rombach2022high}, and GAN loss \( L_{GAN} \). The total training loss is \( L_{VAE} = \lambda_1 L_1 + \lambda_2 L_2 + \lambda_{KL} L_{KL} + \lambda_{GAN} L_{GAN} \).

A pre-trained T5 model serves as the text encoder, converting text input into embeddings with frozen parameters during training. The latent diffusion model learns \( p(z_{out} | z_{in}, c_{text}) \), with a U-Net and cross-attention mechanism~\cite{ho2020denoising}. The training loss is \( L_{LDM} = \mathbb{E}_{(z_{in}, z_{out}, text)} \mathbb{E}_{\epsilon \sim \mathcal{N}(0, I)} \| \epsilon_\theta (z_t, t, z_{in}, c_{text}) - \epsilon \|^2 \). HiFi-GAN~\cite{kong2020hifi} is used as the vocoder for converting mel-spectrograms to audio, balancing quality and efficiency. Triplet data for training the model focuses on tasks like adding, dropping, replacement, inpainting, and super-resolution~\cite{ouyang2022training}.

\sstitle{Speech Editing}
Borsos et al.~\cite{borsos2022speechpainter} propose SpeechPainter, a model designed to fill gaps in speech segments using corresponding transcripts~\cite{zen2019libritts}. It operates on log-mel spectrograms and synthesises audio using a neural vocoder. Built on Perceiver IO~\cite{jaegle2022perceiver}, the model efficiently handles varying input and output sizes.
The model consists of an encoder and decoder~\cite{larsen2016autoencoding}. The encoder processes masked log-mel spectrograms \( X' \) and transcripts \( T \) using cross-attention and self-attention mechanisms~\cite{jaegle2021perceiver}, creating a latent representation \( L \) that guides the decoder to produce the complete spectrogram~\cite{kumar2019melgan,zeghidour2021soundstream}.
Training involves two phases. The first phase uses \( L_1 \) reconstruction loss to train the model:
\[ L_{\text{rec}}^G = \frac{1}{n d_{\text{mel}}} \mathbb{E}_{(X, T)} [\| X - G(X', T) \|_1], \]
where \( G \) is the model.
The second phase introduces adversarial training with a discriminator \( D \) to improve perceptual quality. The discriminator uses hinge loss:
\[ L_D = \mathbb{E}_{(X, T)} \left[ \sum_t \max(0, 1 - D_t(X)) + \sum_t \max(0, 1 + D_t(G(X', T))) \right]. \]

Feature matching loss ensures the generated spectrogram closely matches real data:
\[ L_{\text{feat}}^G = \mathbb{E}_{(X, T)} \left[ \frac{1}{\ell} \sum_{i=1}^\ell \frac{1}{d_i} \| D^i(X) - D^i(G(X', T)) \|_1 \right], \]
where \( \ell \) is the number of discriminator layers.
The final generator loss combines reconstruction and feature matching losses:
\[ L_G = L_{\text{rec}}^G + \lambda_{\text{feat}} L_{\text{feat}}^G, \]
with \( \lambda_{\text{feat}} = 10 \).

\sstitle{Music Editing}
Han et al.~\cite{han2023instructme} proposes InstructME for instruction-based music editing. Text instructions \( y \) are converted into embeddings \( \mathcal{T}(y) \) using a pretrained T5 model~\cite{raffel2020exploring}, and audio segments \( x_s \) are transformed into latent embeddings \( z_s \) by a variational auto-encoder (VAE). A diffusion process generates new audio embeddings, converted back to waveforms by the VAE decoder~\cite{wang2023audit}. The VAE includes an encoder, decoder, and discriminator for enhancing sound quality~\cite{song2021scorebased}. The diffusion model uses Gaussian noise injection and a time-conditional U-Net for denoising, aiming to restore the original latent space \( z_0 \) of the target music~\cite{liu2023audioldm}. For efficient processing of lengthy music sequences, a chunk transformer models long-term dependencies, reducing computational cost. Strategies like multi-scale aggregation and chord-conditional diffusion improve consistency and harmony by capturing high-level music characteristics and emphasising chord progressions~\cite{huang2023noise2music,huang2022mulan}. For remix operations, InstructME employs classifier guidance and classifier-free guidance for controllable generation, mixing classifier gradients or training a conditional model directly.

Instruct-MusicGen~\cite{zhang2024instruct} is a model designed for text-to-music generation and editing, capable of modifying existing music audio based on text instructions~\cite{zhang2024llamaadapter,lin2023content}. It takes two inputs: music audio ($X_{cond}$) and a text instruction ($X_{instruct}$), such as ``Add guitar''. The architecture consists of two main modules: the audio fusion module and the text fusion module.
The audio fusion module converts music audio into embeddings and integrates them into MusicGen using self-attention and cross-attention mechanisms. It processes the audio through EnCodec tokens and re-encodes them into embeddings ($Z_{cond}$), then fuses these with $Z_{music}$.
The text fusion module encodes text instructions with the T5 encoder. It fine-tunes the cross-attention mechanism to handle text instructions, applying Low-Rank Adaptation (LoRA) to reduce the number of parameters needing fine-tuning.

\subsection{Video Editing}
Video editing is more challenging than image editing due to difficulties in maintaining geometric and temporal consistency~\cite{wu2023tune,molad2023dreamix}. 
Given an input video with \( N \) frames (\( I = \{I^1, \ldots, I^N\} \)), the goal is to edit it into a new video (\( I' = \{I'^1, \ldots, I'^N\} \)) based on a natural language instruction \( c \). This preserves the video's semantics. The baseline approach uses an image-based editing model \( f : (\mathcal{I}, \mathcal{T}) \to \mathcal{I} \) to edit each frame individually, \( I' = \{f(I^1, c), \ldots, f(I^N, c)\} \).

Temporal propagation ensures consistency across video frames \cite{esser2023structure,singer2023makeavideo,yu2023video,xu2022temporally}. Methods like Omnimattes and Atlas tackle temporal consistency by decomposing videos into unified 2D atlas layers~\cite{lu2022associating,kasten2021layered}. 

\sstitle{Cascaded diffusion} 
Molad et al.~\cite{molad2023dreamix} propose a text-guided video diffusion model (VDM) for editing, extended to image animation. This approach allows global content mapping back to the video with minimal effort. It corrupts videos by downsampling and adding noise, then uses cascaded VDMs~\cite{ho2022imagen} to upscale and refine them based on a text prompt \( t \).
Input video degradation involves downsampling and adding Gaussian noise, with the noise level \( s \). The corrupted video is then mapped to a high-resolution output guided by the text prompt \( t \).
To improve fidelity, the model undergoes mixed video-image finetuning~\cite{ruiz2023dreambooth}. The input video \( v \) is treated as both a single clip and unordered frames \( u \). The finetuning objective for video \( \mathcal{L}_{\theta}^{\text{vid}}(v) \) ensures exact reconstruction:
\[
\mathcal{L}_{\theta}^{\text{vid}}(v) = \mathbb{E}_{z \sim \mathcal{N}(0, I), s \in \mathcal{U}(0, 1)} || D_{\theta}(z_s, s, t^*, c) - v ||^2.
\]
Frames are finetuned individually using the denoising model \( D_{\theta}^{a} \):
\[
\mathcal{L}_{\theta}^{\text{frame}}(u) = \mathbb{E}_{z \sim \mathcal{N}(0, I), s \in \mathcal{U}(0, 1)} || D_{\theta}^{a}(z_s, s, t^*, c) - u ||^2.
\]
The mixed finetuning objective combines these:

\[
\theta = \arg \min_{\theta'} \alpha \mathcal{L}_{\theta'}^{\text{vid}}(v) + (1 - \alpha) \mathcal{L}_{\theta'}^{\text{frame}}(u),
\]
where \( \alpha \) balances the objectives.

\sstitle{Atlas Edit Layer}
Bar et al.~\cite{bar2022text2live} propose a method for making semantic, localized edits using simple text prompts, such as changing textures or adding effects like smoke. The approach uses CLIP~\cite{radford2021learning} to learn a generator from a single input image or video, addressing the challenge of controlling edits' localization and preserving original content~\cite{frans2022clipdraw}.
The framework has three key components. First, layered editing outputs an RGBA layer composited over the input image with dedicated losses. Second, content preservation and localization losses use CLIP features to guide edits~\cite{avrahami2022blended}. Third, an internal generative prior creates a dataset by augmenting the input image or video and text to train the generator.

For image editing, the generator \( G_{\theta} \) takes a source image \( I_s \) and creates an edit layer \( \mathcal{E} = \{C, \alpha\} \) (color image \( C \) and opacity map \( \alpha \)). The final image \( I_o \) is:
\[ I_o = \alpha \cdot C + (1 - \alpha) \cdot I_s \]
The objective is to align \( I_o \) with a text prompt \( T \). Auxiliary prompts \( T_{\text{screen}} \) and \( T_{\text{ROI}} \) help initialize and localize the edits. The loss function combines composition loss \( L_{\text{comp}} \), screen loss \( L_{\text{screen}} \), structure loss \( L_{\text{structure}} \), and regularization \( L_{\text{reg}} \):
\[ L_{\text{Text2LIVE}} = L_{\text{comp}} + \lambda_g L_{\text{screen}} + \lambda_s L_{\text{structure}} + \lambda_r L_{\text{reg}} \]
Composition loss matches \( I_o \) to \( T \)~\cite{patashnik2021styleclip,gal2022stylegan}:
$ L_{\text{comp}} = L_{\text{cos}}(I_o, T) + L_{\text{dir}}(I_s, I_o, T_{\text{ROI}}, T)$.
Screen loss supervises the edit layer over a green background \( I_{\text{green}} \):
$ L_{\text{screen}} = L_{\text{cos}}(I_{\text{screen}}, T_{\text{screen}})$.
Structure loss preserves spatial layout using self-similarity matrices~\cite{kolkin2019style,tumanyan2022splicing}:
$ L_{\text{structure}} = \| S(I_s) - S(I_o) \|_F$.
Sparsity regularization encourages sparse opacity maps~\cite{lu2020layered}:
$ L_{\text{reg}} = \gamma (\| \alpha \|_1 + \Psi_0(\alpha))$.
A text-driven relevancy loss initializes \( \alpha \):
$ L_{\text{init}} = \text{MSE}(R(I_s), \alpha)$.
The generator is trained with an internal dataset of augmented image-text pairs~\cite{chefer2021generic}.

For video editing, Bar et al.~\cite{bar2022text2live} use Neural Layered Atlases (NLA) for temporal consistency~\cite{kasten2021layered}. NLA decomposes the video into 2D atlases. Each video location \( p = (x, y, t) \) is mapped to a UV location and opacity value. The atlas edit layer \( \mathcal{E}_A = \{C_A, \alpha_A\} \) is mapped to each frame:
\[ \mathcal{E}_t = \text{Sampler}(\mathcal{E}_A, \mathcal{S}) \]
Training involves augmenting atlas crops and applying losses to the edited frames, ensuring plausible and consistent edits.

\sstitle{Inter-frame propagation}
Chai et al.~\cite{chai2023stablevideo} ensure geometric consistency in video editing by fixing the mappings \( UV^b \) and \( UV^f \) and generating edited atlases \( A^b \) and \( A^f \). It uses a pre-trained latent diffusion model~\cite{rombach2022high,zhang2023adding,mou2024t2i} with guided conditions as generators \( G^b(\cdot) \) and \( G^f(\cdot) \). Inter-frame propagation is applied to the foreground atlas, reconstructing the video frame by frame using:
\[
I_i = \alpha_i \circ UV^f_i(G^f(A^f)) + (1 - \alpha_i) \circ UV^b_i(G^b(A^b)),
\]
where \( \circ \) denotes pixel-wise product.
The aggregation network edits video frames, capturing more details from different viewpoints, ensuring higher fidelity. A two-layer 2D convolution network aggregates edited key frames, aligning them with the original. The reconstruction loss \( L_{\text{rec}} \) ensures alignment:
\[
L_{\text{rec}} = \sum_{i=1}^{N} ||E_i - UV^f_i(A^f)||_1.
\]

\sstitle{Zero-shot}
Wang et al.~\cite{wang2023zero} propose a video editing framework, namely vid2vid-zero, that uses  a text-to-image diffusion model~\cite{mokady2023null} in a zero-shot manner. It includes three main components: video inversion for text-to-video alignment, spatial regularization for fidelity, and cross-frame modeling for temporal consistency.

Vid2vid-zero's algorithm begins with real video inversion, using DDIM to map \( X_0 \) to the noise space, resulting in \( \{ X_t^{\text{inv}} \}_{t=1}^T \). Null-text optimization~\cite{mokady2023null} aligns this latent noise with the source prompt \( c \)~\cite{liu2023pre}. Reference cross-attention maps are computed, and for each timestep \( t \) from \( T \) to 1, null-text embeddings are injected, and sampling is performed using DDIM.
In the editing phase, attention masks \( M_t \) and null-text embeddings \( \emptyset_t \) are updated based on the thresholds \( \tau_{\text{null}} \) and \( \tau_M \). DDIM sampling with guidance produces the edited video frames.
Spatial-temporal attention~\cite{wu2023tune} is proposed for temporal modeling, attending to both previous and future frames to ensure bi-directional temporal consistency~\cite{singer2023makeavideo}. Spatial regularization uses cross-attention maps from the inversion phase to maintain the fidelity of the original video, injecting these maps during denoising to focus on prompt-related areas and preserve temporal consistency across frames~\cite{hong2023cogvideo}.

\sstitle{Cross-frame attention}
The integration of cross-frame attention \cite{liu2024video,wang2023zero} with correspondence estimation facilitates feature propagation in video-to-video generation. The primary goal of self-attention is to select values \( V \) using attention scores determined by \( QK^T \). For cross-frame attention, given a token location \( p \) in a frame, the attention score is computed by the cosine similarity between \( Q_p \) and each token in \( K^* \), where key values \( V^* \) represent features across spatial and temporal dimensions. This formulation resembles feature propagation, with the attention score serving as estimated correspondence, resulting in a fused representation of warped features from successive frames.

Attention scores in cross-frame attention serve as correspondence estimation across frames. By designating a query point \( p \) at time \( t \) and determining its corresponding coordinate \( q \) at time \( t' \):
\[
q = \arg \max_{p'} A_{p, p'}, \quad \text{where} \quad A = \text{softmax} \left( \frac{Q^t K^{t'T}}{\sqrt{d}} \right)
\]
Here, \( A_{p, p'} \) represents the matrix element at row \( p \) and column \( p' \). Correspondence is estimated by selecting the location \( p' \) with the highest attention score for \( p \), averaged across multi-head attention, verifying effectiveness in feature propagation.

Fairy~\cite{wu2023fairy}, a video-to-video framework leveraging cross-frame attention, propagates value features from anchor frames to candidate frames. Anchor frames are edited using an image-based model \( f : (\mathcal{I}, \mathcal{T}) \to \mathcal{I} \), and cross-frame attention is used for successive frames. For frame \( I^t \):
\[
\text{softmax} \left( \frac{Q[K, K_{\text{anc}}]^T}{\sqrt{d}} \right) [V, V_{\text{anc}}]
\]
This facilitates cross-frame tracking by estimating temporal correspondence. Editing frame \( I^t \) uses cached features \( K_{\text{anc}} \) and \( V_{\text{anc}} \), allowing long video edits with multi-GPU parallelisation, achieving significant speedup and superior quality without memory constraints, enhancing scalability and practicality.

\sstitle{Decoupled classifier-free guidance}
The EffiVED model's training~\cite{zhang2024effived} involves transforming input video \( V_I \) and editing video \( V_E \) into latent representations \( x_I \) and \( x_E \) using a pre-trained VAE encoder. These representations, along with a noisy latent \( \epsilon \), are processed by a 3D U-Net with cross-attention, guided by a text instruction \( c \). The model aims to transform \( x_I \) to \( x_E \) by minimizing the objective:
\[
\mathbb{E}_{x_I, x_E, \epsilon \sim \mathcal{N}(0,1), t} \left[ \left\| \epsilon - \epsilon_\theta \left( [x_I, x_E], c, t \right) \right\|^2_2 \right],
\]
where \( t \) is the denoising step.
EffiVED uses decoupled classifier-free guidance~\cite{wang2024videocomposer}, balancing input videos and instructions. During training, it randomly selects \( x_I \) or \( c \). During inference, the predicted noise at step \( t \) is computed as:
\begin{multline*}
\tilde{\epsilon}(x_I, c, t) = \tilde{\epsilon}(\phi, \phi, t) + \lambda_1 \left( \tilde{\epsilon}(x_I, c, t) - \tilde{\epsilon}(\phi, c, t) \right) \\ + \lambda_2 \left( \tilde{\epsilon}(x_I, c, t) - \tilde{\epsilon}(x_I, \phi, t) \right),
\end{multline*}
where \( \lambda_1 \) and \( \lambda_2 \) control the influence of text and vision guidance.

\sstitle{Decoupled-guidance attention control}
Liu et al.~\cite{liu2024video} propose Video-P2P, a video editing framework using a Prompt-to-Prompt setting~\cite{hertz2023prompt}. A real video \( \mathcal{V} \) with \( n \) frames, a source prompt \( \mathcal{P} \), and an edited prompt \( \mathcal{P}^* \) generate an edited video \( \mathcal{V}^* \). The object of interest is assumed present in the first frame~\cite{wu2023tune}.
A Text-To-Set (T2S) model for approximate inversion uses $1 \times 3 \times 3$ pattern convolution kernels and temporal attention~\cite{ho2022imagen,ho2022video}, replacing self-attentions with frame-attentions:
\[ Q = W^Q v_i, \quad K = W^K v_0, \quad V = W^V v_0 \]
This processes video pair-by-pair \( n \) times. The model is fine-tuned for noise prediction with a shared unconditional embedding \( \mathcal{E}_t \):
\[ \min_{\mathcal{E}_t} \sum_{i=1}^{n} \| z^*_{t-1, i} - z_{t-1, i} \|_2^2, \]
where \( z_{t-1, i} \) is updated as:
$ z_{t-1, i} = \text{align}(z_t, z_{t, 0}, \mathcal{E}_t, C)$.
Attention control combines two pipelines: an optimized unconditional embedding for the source prompt and an initialized embedding for the target prompt. Initial steps swap attention maps:
\[ \text{Edit}(M_t, M^*_t, t) = \begin{cases} 
M^*_t & \text{if } t < \tau \\
M_t & \text{otherwise}
\end{cases}, \]
where \( M_t, M^*_t \) are cross-attention maps. The binary mask \( B(M_{t, w}) \) determines changes above a threshold for controlled attention.

\sstitle{Audio-driven instruction}
Recently, several approaches for for video scene editing integrate audio properties to enhance visual edits~\cite{shen2024audioscenic,biner2024sonicdiffusion}. Utilizing a modified Stable Diffusion framework, AudioScenic~\cite{shen2024audioscenic} combines Stable Diffusion, VAE autoencoder, and U-Net architectures to guide the editing process. The method integrates audio semantics into a visual editing framework to ensure that visual changes are guided by audio conditions. It employs strategies to preserve key visual details, adjust effects based on audio intensity, and maintain consistency over time by aligning audio and visual features.

\section{Published Resources}
\label{sec:resources}

\subsection{Published Algorithms}
Several key algorithms and models have been instrumental in foundational experiments for instructional image editing controls. \autoref{tab:algorithms} summarises these algorithms, covering a wide array of tasks such as image editing, mask-conditioned and mask-free editing, local and multi-object editing, as well as style editing. The models employ techniques including diffusion, diffusion+CLIP, and various hybrid methods. Each model is linked to a GitHub repository, enabling convenient access to implementation details for further exploration and practical use.

\begin{table*}[!h]
    \centering
    \caption{Published Algorithms and Models}
    \label{tab:algorithms}
    \begin{adjustbox}{max width=\textwidth}
    \begin{threeparttable}
    \begin{tabular}{l|c|c|c|c|l}
        \toprule
	\textbf{Algorithm} & \textbf{Year} & \textbf{Editing Task} & \textbf{Model} & \textbf{Instruction Type} &  \textbf{Repository} \\
	\midrule
	UnifyEditing~\cite{kwon2024unified} & 2024 &  &  & & \url{https://unifyediting.github.io/} \\
	FlexEdit~\cite{nguyen2024flexedit} & 2024 & Image editing & Diffusion & & \url{https://flex-edit.github.io/} \\
	Pix2Pix-OnTheFly~\cite{santos2024pix2pix} & 2024 & Image editing & Diffusion & LLM-powered & \url{https://github.com/pix2pixzero/pix2pix-zero} \\
	LocInv~\cite{tang2024locinv} & 2024 & Image editing & Diffusion+CLIP & & \url{https://github.com/wangkai930418/DPL} \\
	D-Edit~\cite{feng2024item} & 2024 & Mask-conditioned editing & Diffusion & Freestyle & \url{https://github.com/asFeng/d-edit} \\
	LearnableRegions~\cite{lin2024text} & 2024 & Mask-free editing & Diffusion+CLIP & Caption-based & \url{https://yuanze-lin.me/LearnableRegions_page/} \\
	ZONE~\cite{li2024zone} & 2024 & Local editing & Diffusion & Localisation & \url{https://github.com/lsl001006/ZONE} \\
	OIR~\cite{yang2024objectaware} & 2024 & Multi-object & Diffusion+CLIP & Description-based & \url{https://aim-uofa.github.io/OIR-Diffusion/} \\
	InstructCV~\cite{gan2024instructcv} & 2024 & Multi-task & Diffusion & LLM-powered & \url{https://github.com/AlaaLab/InstructCV} \\
	MGIE~\cite{fu2024guiding} & 2024 & & Diffusion &  MLLM-powered & \url{https://mllm-ie.github.io/} \\
	Grounded-Instruct-Pix2Pix~\cite{shagidanov2024grounded} & 2024 & Image editing & Diffusion & Description-based & \url{https://github.com/arthur-71/Grounded-Instruct-Pix2Pix} \\
	InfEdit~\cite{xu2024inversion} & 2024 & &  & & \url{https://sled-group.github.io/InfEdit/} \\
	FoI~\cite{guo2023focus} & 2024 & & Diffusion & Multi-instruction & \url{https://github.com/guoqincode/Focus-on-Your-Instruction} \\
	DPL~\cite{yang2024dynamic} & 2024 & & Diffusion & Command-based & \url{https://github.com/wangkai930418/DPL} \\
	GANTASTIC~\cite{dalva2024gantastic} & 2024 & Image editing & Hybrid & Disentangled directions & \url{https://gantastic.github.io/} \\ 
	StyleBooth~\cite{han2024stylebooth} & 2024 & Style editing & & Multimodal & \url{https://ali-vilab.github.io/stylebooth-page/} \\
	GaussianVTON~\cite{chen2024gaussianvton} & 2024 & Fashion editing & & & \url{https://haroldchen19.github.io/gsvton/} \\
	TexFit~\cite{wang2024texfit} & 2024 & Fashion editing & Diffusion & Caption-based &  \url{https://texfit.github.io/} \\
	ChartReformer~\cite{yan2024chartreformer} & 2024 & Chart editing & & & \url{https://github.com/pengyu965/ChartReformer} \\
	BlenderAlchemy~\cite{huang2024blenderalchemy} & 2024 & 3D editing & & & \url{https://ianhuang0630.github.io/BlenderAlchemyWeb/} \\
	Instruct-MusicGen~\cite{zhang2024instruct} & 2024 & Music editing & & & \url{https://github.com/ldzhangyx/instruct-musicgen} \\
	Video-P2P~\cite{liu2024video} & 2024 & Video editing & Diffusion & & \url{https://video-p2p.github.io/} \\
	EffiVED~\cite{zhang2024effived} & 2024 & Video editing & & & \url{https://github.com/alibaba/EffiVED} \\
	vid2vid-zero~\cite{wang2023zero} & 2023 & Video editing & Diffusion & & \url{https://github.com/baaivision/vid2vid-zero} \\
	Instruct-NeRF2NeRF~\cite{haque2023instruct} & 2023 & 3D editing & &  & \url{https://instruct-nerf2nerf.github.io/} \\
	Instruct 3D-to-3D~\cite{kamata2023instruct} & 2023 & 3D editing & Diffusion & & \url{https://sony.github.io/Instruct3Dto3D-doc/} \\
	InstructME~\cite{han2023instructme} & 2023 & Music editing & Diffusion & & \url{https://musicedit.github.io/} \\
	Fairy~\cite{wu2023fairy} & 2023 & Video editing & & & \url{https://fairy-video2video.github.io/} \\
	AUDIT~\cite{wang2023audit} & 2023 & Audio editing & Diffusion & & \url{https://audit-demo.github.io/} \\
	InstructAny2Pix~\cite{li2023instructany2pix} & 2023 & Image/Audio editing & Diffusion &  LLM-powered & \url{https://github.com/jacklishufan/InstructAny2Pix} \\
	StableVideo~\cite{chai2023stablevideo} & 2023 & Video editing & Diffusion & & \url{https://github.com/rese1f/StableVideo} \\
	Dreamix~\cite{molad2023dreamix} & 2023 & Video editing & Diffusion & & \url{https://dreamix-video-editing.github.io/} \\
	Vox-E~\cite{sella2023vox} & 2023 & 3D Objects & Diffusion+CLIP & & \url{https://tau-vailab.github.io/Vox-E/} \\
	VISII~\cite{nguyen2023visual} & 2023 & Image editing & Diffusion & Visual & \url{https://thaoshibe.github.io/visii/} \\
	InstructPix2Pix~\cite{brooks2023instructpix2pix} & 2023 & & Diffusion+CLIP & & \url{https://www.timothybrooks.com/instruct-pix2pix} \\
	Null-text~\cite{mokady2023null} & 2023 & & Diffusion+CLIP & & \url{https://null-text-inversion.github.io/} \\
	MasaCtrl~\cite{cao2023masactrl} & 2023 & & Diffusion & & \url{https://github.com/TencentARC/MasaCtrl} \\
	PlugNPlay~\cite{tumanyan2023plug} & 2023 & & Diffusion & & \url{https://pnp-diffusion.github.io/} \\
	ImagenEditor~\cite{wang2023imagen} & 2023 & Iterative editing & Diffusion+CLIP & Mask-based & \url{https://imagen.research.google/editor/} \\
	MoEController~\cite{li2023moecontroller} & 2024 & & Hybrid & LLM-powered & \url{https://oppo-mente-lab.github.io/moe_controller/} \\
	Imagic~\cite{kawar2023imagic} & 2023 & Real images & Diffusion & Caption-based & \url{https://imagic-editing.github.io/} \\
	SmartEdit~\cite{huang2023smartedit} & 2023 & Object Removal or Replacement & Diffusion & MLLM-empowered & \url{https://yuzhou914.github.io/SmartEdit/}\\
	InstructDiffusion~\cite{geng2023instructdiffusion} & 2023 & Multi-turn, Multi-task & Diffusion &  LLM-powered & \url{https://gengzigang.github.io/instructdiffusion.github.io/} \\
	EmuEdit~\cite{sheynin2023emu} & 2023 & Multi-turn, multi-task & Diffusion & Action-based & \url{https://emu-edit.metademolab.com/} \\
	InstructEdit~\cite{wang2023instructedit} & 2023 &  & Diffusion & LLM-powered & \url{https://qianwangx.github.io/InstructEdit/} \\
	PhotoVerse~\cite{chen2023photoverse} & 2023 & & Diffusion &  & \url{https://photoverse2d.github.io/} \\
	PaintbyExample~\cite{yang2023paint} & 2023 & & Diffusion+CLIP & Example-based & \url{https://github.com/Fantasy-Studio/Paint-by-Example} \\
	DiffEdit~\cite{couairon2023diffedit} & 2023 & Mask-free editing & Diffusion & Caption-based & \url{https://github.com/Xiang-cd/DiffEdit-stable-diffusion/} \\
	Inst-inpaint~\cite{yildirim2023inst} & 2023 & Scene editing & Diffusion & Action-based & \url{https://instinpaint.abyildirim.com/} \\
	Prompt2Prompt~\cite{hertz2023prompt} & 2023 & Localised editing & Diffusion & Caption-based & \url{https://prompt-to-prompt.github.io/} \\
	ChatFace~\cite{yue2023chatface} & 2023 & Face editing & Diffusion & Dialog-based & \url{https://dongxuyue.github.io/chatface/} \\
	DE-Net~\cite{tao2023net} & 2023 & Multi-task & GAN & Description-based & \url{https://github.com/tobran/DE-Net} \\
	cManiGAN~\cite{fan2023target} & 2023 & 3D Objects &  GAN & & \url{https://sites.google.com/view/wancyuanfan/projects/cmanigan} \\
	FICE~\cite{pernuvs2023fice} & 2023 & Fashion images & GAN+CLIP & & \url{https://github.com/MartinPernus/FICE} \\
	SpeechPainter~\cite{borsos2022speechpainter} & 2022 & & & & \url{https://google-research.github.io/seanet/speechpainter/examples/} \\
	Text2LIVE~\cite{bar2022text2live} & 2022 & Image/Video editing & GAN+CLIP &  & \url{https://text2live.github.io/} \\
	PTI~\cite{roich2022pivotal} & 2022 & Real images & GAN & Command-based & \url{https://github.com/danielroich/PTI} \\
	ManiTrans~\cite{wang2022manitrans} & 2022 & Multi-object & GAN & Description-based & \url{https://jawang19.github.io/manitrans/} \\
	StyleMC~\cite{kocasari2022stylemc} & 2022 & & GAN+CLIP & Command-based & \url{https://catlab-team.github.io/stylemc/} \\
	BlendedDiffusion~\cite{avrahami2022blended} & 2022 & Mask-conditioned editing & Diffusion & Caption-based & \url{https://github.com/omriav/blended-diffusion} \\
	Text2Mesh~\cite{michel2022text2mesh} & 2022 & 3D Objects & GAN &  & \url{https://threedle.github.io/text2mesh/} \\
	VQGAN-CLIP~\cite{crowson2022vqgan} & 2022 & Domain, Colour, Background & GAN & Caption-based & \url{https://github.com/EleutherAI/vqgan-clip/} \\
	StyleGAN-NADA~\cite{gal2022stylegan} & 2022 & Out-of-domain editing & GAN+CLIP & Command-based  & \url{https://stylegan-nada.github.io/} \\
	DiffusionCLIP~\cite{kim2022diffusionclip} & 2022 & Face editing  &  Diffusion+CLIP & Command-based & \url{https://github.com/gwang-kim/DiffusionCLIP} \\
	GLIDE~\cite{nichol2022glide} & 2022 & Photorealistic & Diffusion+CLIP & Mask-conditioned & \url{https://github.com/openai/glide-text2im} \\
	Talk2Edit~\cite{jiang2021talk} & 2021 & Interactive editing & GAN & Dialog-based & \url{https://www.mmlab-ntu.com/project/talkedit/} \\
	TediGAN~\cite{xia2021tedigan} & 2021 & Continuous editing, Sketches  & GAN & Description-based &  \url{https://github.com/IIGROUP/TediGAN}  \\
	StyleCLIP~\cite{patashnik2021styleclip} & 2021 & Generated and real images & GAN+CLIP & Command-based &\url{https://github.com/orpatashnik/StyleCLIP} \\
	LightManiGAN~\cite{li2020lightweight} & 2020 &  & GAN & & \url{https://github.com/mrlibw/Lightweight-Manipulation} \\
	ManiGAN~\cite{li2020manigan} & 2020 &  & GAN &  & \url{https://github.com/mrlibw/ManiGAN} \\
Open-edit~\cite{liu2020open} & 2020 & Open-domain image & GAN & Open-vocabulary & \url{https://github.com/xh-liu/Open-Edit} \\
GeNeVA~\cite{el2019tell} & 2019 & & GAN & & \url{https://github.com/Maluuba/GeNeVA} \\
LBIE~\cite{chen2018language} & 2018 & Segmentation, Colourization & GAN & Description-based & \url{https://github.com/Jianbo-Lab/LBIE} \\
         \bottomrule
    \end{tabular}
    \end{threeparttable}
    \end{adjustbox}
\end{table*}

\subsection{Published Datasets}
Foundational experiments in instructional image editing controls have relied on key datasets, as outlined in \autoref{tab:datasets}. These datasets support a variety of tasks, including mask-conditioned editing, local editing, and style modification, and are categorised by application domain. Notable datasets are described below.

\begin{table*}[!h]
    \centering
    \caption{Highlighted Datasets} %
    \label{tab:datasets}
    \begin{adjustbox}{max width=\textwidth}
    \begin{threeparttable}
    \begin{tabular}{c|c|c|c|l}
        \toprule
	\textbf{Category} & \textbf{Dataset} &  \textbf{\#Items} & \textbf{Experimented in} & \textbf{URL} \\
	\midrule
    General & Reason-Edit~\cite{huang2023smartedit} & 12.4M+ & \cite{huang2023smartedit} & \url{https://github.com/TencentARC/SmartEdit} 
    \\
	& MagicBrush~\cite{zhang2024magicbrush} & 10K & \cite{huang2023smartedit} &  \url{https://osu-nlp-group.github.io/MagicBrush/} 
    \\
	& InstructPix2Pix~\cite{brooks2023instructpix2pix} & 500K & \cite{huang2023smartedit} & \url{https://github.com/timothybrooks/instruct-pix2pix} 
    \\
	& EditBench~\cite{wang2023imagen} & 240 & \cite{wang2023imagen} & \url{https://imagen.research.google/editor/} 
    \\
    \midrule
    Image Captioning & Conceptual Captions~\cite{sharma2018conceptual} & 3.3M & \cite{liu2020open} & \url{https://ai.google.com/research/ConceptualCaptions/} 
    \\
	& CoSaL~\cite{cosal2015} & 22K+ & \cite{chen2018language} & \url{https://github.com/dragonlee258079/DMT}
	\\
	& ReferIt~\cite{kazemzadeh2014referitgame} & 19K+ & \cite{chen2018language} & \url{https://www.cs.rice.edu/~vo9/referit/}
	\\
	& Oxford-102 Flowers~\cite{nilsback2008automated} & 8K+ & \cite{chen2018language} & \url{https://www.robots.ox.ac.uk/~vgg/data/flowers/102/}
	\\
    & lion-5b \cite{schuhmann2022laion} & 5.85B+ & \cite{li2023moecontroller} & \url{https://laion.ai/blog/laion-5b/}
    \\
    & MS-COCO \cite{lin2014microsoft} & 330K & \cite{gan2024instructcv, wang2023self} & \url{https://cocodataset.org/#home}
    \\
    & DeepFashion \cite{liu2016deepfashion} & 800K & \cite{gan2024instructcv, wang2023self} & \url{https://mmlab.ie.cuhk.edu.hk/projects/DeepFashion.html}
    \\
    & Fashion-IQ \cite{guo2019fashion} & 77K+ & \cite{li2024survey} & \url{https://github.com/XiaoxiaoGuo/fashion-iq}
    \\
    & Fashion200k \cite{han2017automatic} & 200K & \cite{li2024survey} & \url{https://github.com/xthan/fashion-200k}
    \\
    & MIT-States \cite{StatesAndTransformations} & 63K+ & \cite{li2024survey} & \url{https://web.mit.edu/phillipi/Public/states_and_transformations/index.html}
    \\
    & CIRR \cite{Liu_2021_ICCV} & 36K+ & \cite{li2024survey} & \url{https://github.com/Cuberick-Orion/CIRR}
    \\
    \midrule
    ClipArt & CoDraw~\cite{kim2019codraw} & 58K+ & \cite{el2019tell} & \url{https://github.com/facebookresearch/CoDraw} 
    \\
    \midrule
	VQA\tnote{1} & i-CLEVR~\cite{johnson2017inferring} & 70K+ & \cite{el2019tell} & \url{https://github.com/facebookresearch/clevr-iep} 
    \\
    \midrule
    Semantic Segmentation & ADE20K \cite{zhou2019semantic} & 27K+ & \cite{gan2024instructcv} & \url{https://github.com/CSAILVision/ADE20K} 
    \\
    \midrule
    Object Classification & Oxford-III-Pets \cite{parkhi2012cats} & 7K+ & \cite{gan2024instructcv} & \url{https://www.robots.ox.ac.uk/~vgg/data/pets/} 
    \\
    \midrule
    Depth Estimation & NYUv2 \cite{silberman2012indoor} & 408K+ & \cite{gan2024instructcv} & \url{https://cs.nyu.edu/~fergus/datasets/nyu_depth_v2.html} 
    \\
    \midrule
    Aesthetic-Based Editing & Laion-Aesthetics V2 \cite{schuhmann2021laion} & 2.4B+ & \cite{zhang2024hive} & \url{https://laion.ai/blog/laion-aesthetics/} 
    \\
    \midrule
    Dialog-Based Editing & CelebA-Dialog \cite{jiang2021talk} & 202K+ & \cite{jiang2021talk} & \url{https://mmlab.ie.cuhk.edu.hk/projects/CelebA/CelebA_Dialog.html} 
    \\ 
    & Flickr-Faces-HQ (FFHQ) \cite{karras2019style} & 70K & \cite{gan2024instructcv, wang2023self} & \url{https://github.com/NVlabs/ffhq-dataset} 
    \\
    \bottomrule
    \end{tabular}
    \begin{tablenotes}
\item[1] VQA: Visual Question Answering.
\end{tablenotes}
    \end{threeparttable}
    \end{adjustbox}
\end{table*}

\sstitle{General}
\emph{Reason-Edit}~\cite{huang2023smartedit}. For enhancing reasoning skills, a new data production pipeline was established, creating a synthetic editing dataset with around 476 paired samples. This dataset focuses on complex understanding and reasoning scenarios, including details like object location, color, size, and context-specific attributes such as mirror reflections which require a deep understanding of the scene.

\emph{MagicBrush}~\cite{zhang2024magicbrush} contains images with corresponding semantic segmentation masks and natural language instructions for image editing. The dataset is intended for training and evaluating models for automatic image editing.

\emph{InstructPix2Pix}~\cite{brooks2023instructpix2pix} provides pairs of images and corresponding semantic segmentation masks, allowing for training and evaluation of models for image-to-image translation tasks such as semantic segmentation to image generation.

\sstitle{Image Captioning}
\emph{Conceptual Captions}~\cite{sharma2018conceptual} dataset comprises 3 million web-harvested image-caption pairs from various domains and styles, using Alt-text for captions despite their variability and noise. This large and diverse dataset allows a representation learning model to learn an effective space for image manipulation, and trains an image decoder to develop a general prior for open-domain images~\cite{liu2020open}.

\emph{CoSaL}~\cite{chen2018language} focuses on grounding language in images and contains images with captions that describe specific regions or objects within the images.

\emph{ReferIt}~\cite{kazemzadeh2014referitgame} provides images along with referring expressions, which are natural language descriptions that specify a region of interest in an image.

\sstitle{ClipArt} 
\emph{CoDraw}~\cite{kim2019codraw} is a clip art-like dataset featuring images of children playing in a park, with varying poses and expressions, alongside objects like trees, tables, and animals. It contains 58 different object types and comes with conversations between a Teller and a Drawer, who are Amazon Mechanical Turk workers communicating in natural language. The Drawer creates or updates the canvas following the Teller's instructions and may ask for clarification. The dataset comprises 9,993 scenes of different lengths. Each scene begins with an initial canvas that includes only the sky and grass.

\sstitle{VQA}
\emph{i-CLEVR}~\cite{el2019tell} is an iteration of the CLEVR dataset~\cite{johnson2017inferring}, which is used for Visual Question Answering tasks and includes images of objects in various shapes, colours, materials, and sizes, with accompanying complex questions. The i-CLEVR dataset enhances CLEVR by providing a sequence of 5 (image, instruction) pairs, where each instruction guides the addition of an object to an initially empty canvas, specifying shape, colour, and position relative to existing objects. The aim is to make models use context by referring to the most recently added object as ``it''. The i-CLEVR dataset contains 10,000 sequences, totalling 50,000 images and instructions, with a training split of 6,000 sequences and validation and testing splits of 2,000 sequences each.

\sstitle{Semantic Segmentation}
\emph{ADE20K}~\cite{zhou2019semantic} is one of the largest public datasets for semantic segmentation, consisting of diverse scenes in different environments. The dataset includes annotations for object-level segmentation of over 150 object categories.

\sstitle{Object Classification} 
\emph{Oxford-III-Pets}~\cite{parkhi2012cats} contains images of cats belonging to 37 different breeds. Each image is annotated with the corresponding breed label. The dataset is commonly used for evaluating methods for fine-grained object recognition.

\sstitle{Depth Estimation} 
\emph{NYUv2}~\cite{silberman2012indoor} provides RGB-D images captured by a Kinect sensor in indoor scenes. The dataset includes pixel-wise depth maps, allowing for the evaluation of algorithms for depth estimation and scene understanding.

\sstitle{Aesthetic-Based Editing} 
\emph{Laion-Aesthetics V2}~\cite{schuhmann2021laion} contains images with corresponding aesthetics scores. The dataset is designed for applications in image editing, such as automatically enhancing the aesthetic quality of images.

\sstitle{Dialog-Based Editing} \emph{CelebA-Dialog}~\cite{jiang2021talk} consists of images of celebrities along with attribute annotations and natural language dialogues. The dataset is intended for research on editing images based on dialogues.

\emph{Flickr-Faces-HQ (FFHQ)}~\cite{karras2019style} contains high-quality images of human faces from the Flickr website. The dataset is commonly used for tasks such as face recognition and style transfer.

\subsection{Evaluation Metrics}

\autoref{tab:evaluation_metrics} highlights key evaluation metrics for instructional image editing, covering tasks like mask-conditioned, local, multi-object editing, and style modification. 
Key categories of metrics include Perceptual Quality, Structural Integrity, Semantic Alignment, User-Based Metrics, Diversity and Fidelity, Consistency and Cohesion, and Robustness. 

\sstitle{Perceptual Quality}
\textit{LPIPS} (Learned Perceptual Image Patch Similarity) \cite{gal2022stylegan, brooks2023instructpix2pix}: 
This metric measures the perceptual similarity between two images by comparing the activations of a neural network. It captures how similar two images are at a perceptual level. 
\[
\text{LPIPS}(x, x') = \sum_l ||\phi_l(x) - \phi_l(x')||^2
\]
where \( \phi_l(x) \) denotes the feature map of image \(x\) at layer \(l\) of the neural network, and \(x'\) is the compared image. The sum is taken over all layers \(l\) of the neural network.

\textit{SSIM} (Structural Similarity Index) \cite{couairon2023diffedit, nichol2022glide}: 
SSIM quantifies structural similarity between two images by considering luminance, contrast, and structure. It is commonly used to assess image quality, with higher values indicating more similarity.
\[
\text{SSIM}(x, x') = \frac{(2\mu_x\mu_{x'} + C_1)(2\sigma_{xx'} + C_2)}{(\mu_x^2 + \mu_{x'}^2 + C_1)(\sigma_x^2 + \sigma_{x'}^2 + C_2)}
\]
where \( \mu_x \) and \( \sigma_x \) are the mean and standard deviation of image \(x\), respectively, \( \mu_{x'} \) and \( \sigma_{x'} \) are the mean and standard deviation of image \(x'\), and \( \sigma_{xx'} \) is the covariance between the two images. Constants \(C_1\) and \(C_2\) are small values to avoid division by zero.

\sstitle{Structural Integrity}
\textit{PSNR} (Peak Signal-to-Noise Ratio) \cite{yan2024chartreformer, nichol2022glide}: 
PSNR measures the quality of image reconstruction by comparing the peak signal strength to the noise. Higher values suggest better quality.
\[
\text{PSNR} = 10 \log_{10} \left( \frac{\text{MAX}^2}{\text{MSE}} \right)
\]
where \(\text{MAX}\) is the maximum possible pixel value of the image (e.g., 255 for 8-bit images), and MSE is the Mean Squared Error between the original and reconstructed images:
\[
\text{MSE} = \frac{1}{N} \sum_{i=1}^{N} (x_i - \hat{x}_i)^2
\]
where \(x_i\) is the pixel value of the original image and \(\hat{x}_i\) is the pixel value of the reconstructed image.

\textit{mIoU} (Mean Intersection over Union) \cite{couairon2023diffedit, zhao2024instructbrush}: 
This metric quantifies segmentation accuracy by measuring the overlap between the predicted and ground truth segmentation masks.
\[
\text{mIoU} = \frac{1}{N} \sum_{i=1}^{N} \frac{|A_i \cap B_i|}{|A_i \cup B_i|}
\]
where \(A_i\) and \(B_i\) are the predicted and ground truth segmentation masks for the \(i\)-th class, respectively, and \(|A_i \cap B_i|\) is the area of overlap, while \(|A_i \cup B_i|\) is the area covered by either mask.

\sstitle{Semantic Alignment}
\textit{Edit Consistency} \cite{chai2023stablevideo, nguyen2024flexedit}: 
This metric measures the consistency of edits across similar prompts. It quantifies how often the same edit is generated for similar input prompts.
\[
\text{EC} = \frac{1}{N} \sum_{i=1}^{N} 1\{E_i = E_{\text{ref}}\}
\]
where \(E_i\) represents the edit produced for the \(i\)-th input prompt, and \(E_{\text{ref}}\) is the reference edit. The indicator function \(1\{ \cdot \}\) is 1 if the edit matches the reference and 0 otherwise.

\textit{Target Grounding Accuracy} \cite{shagidanov2024grounded, wang2023instructedit}: 
This metric evaluates how accurately the model follows specific instructions in its edits, calculating the ratio of correctly matched target edits to total targets.
\[
\text{TGA} = \frac{\text{Correct Targets}}{\text{Total Targets}}
\]
where \(\text{Correct Targets}\) is the number of target instructions correctly followed, and \(\text{Total Targets}\) is the total number of instructions given.

\sstitle{User-Based Metrics}
\textit{User Study Ratings} \cite{brooks2023instructpix2pix, zhang2024hive}: 
This subjective metric measures user preferences and satisfaction with edited images, typically based on a survey or rating system. Higher ratings indicate better perceived quality.

\textit{Human Visual Turing Test (HVTT)} \cite{jiang2021talk, wang2023instructedit}: 
HVTT assesses if users perceive generated images as real by calculating the ratio of real judgments to total judgments.
\[
\text{HVTT} = \frac{\text{Real Judgements}}{\text{Total Judgements}}
\]
where \(\text{Real Judgements}\) is the number of times a user identifies an image as real, and \(\text{Total Judgements}\) is the total number of judgments made by the user.

\sstitle{Diversity and Fidelity}
\textit{Edit Diversity} \cite{zhao2024instructbrush, han2023instructme}: 
This metric evaluates the variety of generated edits by comparing the distribution of outputs to the mean output distribution.
\[
\text{Diversity} = \frac{1}{N} \sum_{i=1}^{N} D_{KL}(p_i || p_{\text{mean}})
\]
where \(D_{KL}\) is the Kullback-Leibler divergence, \(p_i\) is the probability distribution of the \(i\)-th edit, and \(p_{\text{mean}}\) is the mean distribution of all edits.

\textit{Reconstruction Error} \cite{mokady2023null, nichol2022glide}: 
This metric measures the error between the original image and its reconstruction. Smaller values indicate better reconstruction.
\[
\text{RE} = ||x - \hat{x}||
\]
where \(x\) is the original image and \(\hat{x}\) is the reconstructed image. The error is calculated as the L2 norm of the difference.

\sstitle{Consistency and Cohesion}
\textit{Scene Consistency} \cite{zhang2024tie, joseph2024iterative}: 
Scene Consistency measures the overall visual cohesion of an edited image, ensuring consistency between the various elements in the scene.
\[
\text{SC} = \frac{1}{N} \sum_{i=1}^{N} \text{Sim}(S_i, S_{\text{target}})
\]
where \(\text{Sim}(S_i, S_{\text{target}})\) measures the similarity between the edited scene \(S_i\) and the target scene \(S_{\text{target}}\).

\textit{Shape Consistency} \cite{sabat2024nerf, yang2024objectaware}: 
This metric evaluates how accurately the shape of objects in an image is preserved after editing.
\[
\text{ShapeSim} = \frac{1}{N} \sum_{i=1}^{N} \text{IoU}(S_{\text{edit}}, S_{\text{orig}})
\]
where \(S_{\text{edit}}\) and \(S_{\text{orig}}\) are the edited and original shapes, respectively, and \(\text{IoU}\) is the Intersection over Union between the shapes.

\sstitle{Robustness}
\textit{Noise Robustness} \cite{huang2023noise2music, nichol2022glide}: 
This metric assesses how well the model performs under noisy conditions by comparing the original image to a noisy version.
\[
\text{NR} = \frac{1}{N} \sum_{i=1}^{N} ||x_i - x_{i,\text{noisy}}||
\]
where \(x_i\) is the original image, and \(x_{i,\text{noisy}}\) is the noisy version of the image.

\textit{Perceptual Quality} \cite{brooks2023instructpix2pix, yan2024chartreformer}: 
This is a subjective measure of the quality of an edited image, often derived from human feedback or ratings. Higher ratings indicate better perceived quality.

\begin{table*}[!h]
\centering
\caption{Metrics for Instruction-Based Image Editing Evaluation.}
\label{tab:evaluation_metrics}
\begin{adjustbox}{max width=\textwidth}
\begin{threeparttable}
\begin{tabular}{c|p{4cm}|l|p{8cm}}
\hline
\textbf{Category} & \textbf{Metric} & \textbf{Formula} & \textbf{Description} \\
\hline
\multirow{4}{*}{Perceptual Quality} 
& Learned Perceptual Image Patch Similarity (LPIPS) \cite{gal2022stylegan,brooks2023instructpix2pix} & \( \text{LPIPS}(x, x') = \sum_l ||\phi_l(x) - \phi_l(x')||^2 \) & Measures perceptual similarity between images, with lower scores indicating higher similarity. \\
\cline{2-4}
& Structural Similarity Index (SSIM) \cite{couairon2023diffedit, nichol2022glide} & \( \text{SSIM}(x, x') = \frac{(2\mu_x\mu_{x'} + C_1)(2\sigma_{xx'} + C_2)}{(\mu_x^2 + \mu_{x'}^2 + C_1)(\sigma_x^2 + \sigma_{x'}^2 + C_2)} \) & Measures visual similarity based on luminance, contrast, and structure. \\
\cline{2-4}
& Fréchet Inception Distance (FID) \cite{brooks2023instructpix2pix, kim2022diffusionclip} & \( \text{FID} = ||\mu_r - \mu_g||^2 + \text{Tr}(\Sigma_r + \Sigma_g - 2(\Sigma_r \Sigma_g)^{1/2}) \) & Measures the distance between the real and generated image feature distributions. \\
\cline{2-4}
& Inception Score (IS) \cite{kim2022diffusionclip, han2024stylebooth} & \( \text{IS} = \exp(\mathbb{E}_{x} D_{KL}(p(y|x) || p(y))) \) & Evaluates image quality and diversity based on label distribution consistency. \\
\hline
\multirow{4}{*}{Structural Integrity} 
& Peak Signal-to-Noise Ratio (PSNR) \cite{yan2024chartreformer, nichol2022glide} & \( \text{PSNR} = 10 \log_{10} \left( \frac{\text{MAX}^2}{\text{MSE}} \right) \) & Measures image quality based on pixel-wise errors, with higher values indicating better quality. \\
\cline{2-4}
& Mean Intersection over Union (mIoU) \cite{couairon2023diffedit, zhao2024instructbrush} & \( \text{mIoU} = \frac{1}{N} \sum_{i=1}^{N} \frac{|A_i \cap B_i|}{|A_i \cup B_i|} \) & Assesses segmentation accuracy by comparing predicted and ground truth masks. \\
\cline{2-4}
& Mask Accuracy \cite{wang2023instructedit, couairon2023diffedit} & \( \text{Accuracy} = \frac{\text{TP} + \text{TN}}{\text{TP} + \text{FP} + \text{FN} + \text{TN}} \) & Evaluates the accuracy of generated masks. \\
\cline{2-4}
& Boundary Adherence \cite{zhao2024instructbrush, han2024stylebooth} & \( \text{BA} = \frac{|B_{\text{edit}} \cap B_{\text{target}}|}{|B_{\text{target}}|} \) & Measures how well edits preserve object boundaries. \\
\hline
\multirow{3}{*}{Semantic Alignment} 
& Edit Consistency \cite{chai2023stablevideo, nguyen2024flexedit} & \( \text{EC} = \frac{1}{N} \sum_{i=1}^{N} 1\{E_i = E_{\text{ref}}\} \) & Measures the consistency of edits across similar prompts. \\
\cline{2-4}
& Target Grounding Accuracy \cite{shagidanov2024grounded, wang2023instructedit} & \( \text{TGA} = \frac{\text{Correct Targets}}{\text{Total Targets}} \) & Evaluates how well edits align with specified targets in the prompt. \\
\cline{2-4}
& Embedding Space Similarity \cite{gal2022stylegan, michel2022text2mesh} & \( \text{CosSim}(v_x, v_{x'}) = \frac{v_x \cdot v_{x'}}{||v_x|| \, ||v_{x'}||} \) & Measures similarity between the edited and reference images in feature space. \\
\cline{2-4}
& Decomposed Requirements Following Ratio (DRFR) \cite{qin2024infobench} & \( \text{DRFR} = \frac{1}{N} \sum_{i=1}^{N} \frac{\text{Requirements Followed}_{i}}{\text{Total Requirements}_{i}} \) & Assesses how closely the model follows decomposed instructions. \\
\hline
\multirow{3}{*}{User-Based Metrics} 
& User Study Ratings \cite{brooks2023instructpix2pix, zhang2024hive} & & Captures user feedback through ratings of image quality. \\
\cline{2-4}
& Human Visual Turing Test (HVTT) \cite{jiang2021talk, wang2023instructedit} & \( \text{HVTT} = \frac{\text{Real Judgements}}{\text{Total Judgements}} \) & Measures the ability of users to distinguish between real and generated images. \\
\cline{2-4}
& Click-through Rate (CTR) \cite{gan2024instructcv, yan2024chartreformer} & \( \text{CTR} = \frac{\text{Clicks}}{\text{Total Impressions}} \) & Tracks user engagement by measuring image clicks. \\
\hline
\multirow{4}{*}{Diversity and Fidelity} 
& Edit Diversity \cite{zhao2024instructbrush, han2023instructme} & \( \text{Diversity} = \frac{1}{N} \sum_{i=1}^{N} D_{KL}(p_i || p_{\text{mean}}) \) & Measures the variability of generated images. \\
\cline{2-4}
& GAN Discriminator Score \cite{dalva2024gantastic, patashnik2021styleclip} & \( \text{GDS} = \frac{1}{N} \sum_{i=1}^N D_{\text{GAN}}(x_i) \) & Assesses the authenticity of generated images using a GAN discriminator. \\
\cline{2-4}
& Reconstruction Error \cite{mokady2023null, nichol2022glide} & \( \text{RE} = ||x - \hat{x}|| \) & Measures the error between the original and generated images. \\
\cline{2-4}
& Edit Success Rate \cite{li2024zone, yildirim2023inst} & \( \text{ESR} = \frac{\text{Successful Edits}}{\text{Total Edits}} \) & Quantifies the success of applied edits. \\
\hline
\multirow{4}{*}{Consistency and Cohesion} 
& Scene Consistency \cite{zhang2024tie, joseph2024iterative} & \( \text{SC} = \frac{1}{N} \sum_{i=1}^{N} \text{Sim}(I_{\text{edit}}, I_{\text{orig}}) \) & Measures how edits maintain overall scene structure. \\
\cline{2-4}
& Color Consistency \cite{wang2024texfit, han2024stylebooth} & \( \text{CC} = \frac{1}{N} \sum_{i=1}^{N} \frac{|C_{\text{edit}} \cap C_{\text{orig}}|}{|C_{\text{orig}}|} \) & Measures color preservation between edited and original regions. \\
\cline{2-4}
& Shape Consistency \cite{sabat2024nerf, yang2024objectaware} & \( \text{ShapeSim} = \frac{1}{N} \sum_{i=1}^{N} \text{IoU}(S_{\text{edit}}, S_{\text{orig}}) \) & Quantifies how well shapes are preserved during edits. \\
\cline{2-4}
& Pose Matching Score \cite{bodur2023iedit, xia2021tedigan} & \( \text{PMS} = \frac{1}{N} \sum_{i=1}^{N} \text{Sim}(\theta_{\text{edit}}, \theta_{\text{orig}}) \) & Assesses pose consistency between original and edited images. \\
\hline
\multirow{2}{*}{Robustness} 
& Noise Robustness \cite{huang2023noise2music, nichol2022glide} & \( \text{NR} = \frac{1}{N} \sum_{i=1}^{N} ||x_i - x_{i,\text{noisy}}|| \) & Evaluates model robustness to noise. \\
\cline{2-4}
& Perceptual Quality \cite{brooks2023instructpix2pix, yan2024chartreformer} & \( \text{PQ} = \frac{1}{N} \sum_{i=1}^{N} \text{Score}(x_i) \) & A subjective quality metric based on human judgment. \\
\hline
\end{tabular}
\end{threeparttable}
\end{adjustbox}
\end{table*}

\section{Challenges and Future Directions}
\label{sec:future}

Instruction-guided editing is rapidly advancing with MLLMs and text-to-image models, offering adaptable and ethical solutions that meet user needs and societal standards. Continued research will support integration across diverse applications, though key challenges remain for further development.

\subsection{Safety Considerations}
Ensuring safety in instructional image editing is crucial as these models handle sensitive and diverse content. Key safety considerations include safeguarding user privacy, maintaining content integrity, and ensuring fairness across applications.

\sstitle{Privacy Protection and Anonymization}
Protecting user privacy in image editing is crucial, especially for tasks involving personal features like faces or locations \cite{nichol2022glide}. Embedding techniques such as differential privacy within models reduces data leakage risks by adding noise or transformations, preserving functionality without compromising data. Anonymization processes, like blurring identifiable traits, further protect identities while enabling data use for modeling~\cite{hukkelaas2023does}.

\sstitle{Preventing Harmful Outputs}
Instructional editing models need safeguards to prevent harmful or inappropriate content generation \cite{wang2022informing}. Explicit content filters and human-in-the-loop reviews \cite{zhang2024hive} enhance model safety, especially in sensitive applications. Adversarial filtering and real-time monitoring further ensure alignment with ethical AI guidelines in creative and commercial contexts.

\sstitle{Robustness to Malicious Instructions and Inputs} 
Models must be resilient to adversarial instructions \cite{truong2024attacks}, where users may try to manipulate outputs maliciously. Techniques like adversarial training and input validation strengthen models against such threats, while constraints on transformations prevent processing harmful requests that could violate privacy or produce offensive content~\cite{yang2024structure}.

\sstitle{Bias and Fairness in Generated Content}
Bias in models can disproportionately impact certain groups or cultures, particularly in face or fashion editing~\cite{tanjim2024discovering}. Ensuring diverse, representative training data promotes fairness \cite{d2024improving}. Ongoing monitoring and auditing of outputs help identify and correct biases, preventing unintended stereotyping.

\subsection{Transparency, Accountability, and Ethics}

Transparency in model behavior \cite{beck2016shaping} is essential for user trust, particularly with sensitive edits. Users benefit from understanding model transformations, especially with complex alterations. Logging commands and outcomes, along with review options, supports user control and aligns with accountability standards.

Instruction-based models should include safeguards~\cite{ni2025responsible} for responsible use, especially in open-source and commercial contexts. These can involve access controls, clear documentation of capabilities and limitations, and ethical usage agreements \cite{wang2022ethical}. Such practices enhance accountability, aligning deployment with ethical AI principles.

Advanced instruction-guided media manipulation raises ethical concerns around misinformation and content authenticity. Establishing an ethical framework with automated detection mechanisms to spot manipulations \cite{aneja2023cosmos} is essential. Additionally, watermarking or authenticity certification can verify content integrity, promoting responsible use of these tools.

\subsection{Editing Authorisation}

Instruction-guided image editing models offer convenience but pose risks of unauthorised modifications, potentially harming reputations and spreading misinformation, especially for public figures. Strategies to protect images from diffusion models include adding noise to prevent style mimicry~\cite{shan2023glaze}, using adversarial examples~\cite{liang2023adversarial}, and degrading DreamBooth model quality to prevent copyright infringement~\cite{van2023anti}.

Chen et al.~\cite{chen2023editshield} propose EditShield, a method that distorts latent representations, leading to unrealistic edits. The optimisation of perturbation \( \delta \) maximises the embedding distance between original and altered images:
\[
\delta = \arg \max_{\|\delta\|_p \leq \xi} \text{Dist}(\mathcal{E}(x + \delta), \mathcal{E}(x))
\]
where \( \xi \) is the perturbation budget. This universal perturbation, iterated over image sets \( D \) with gradient updates, makes images safer for online publication~\cite{moosavi2017universal}.

\subsection{Instruction Comprehension and Flexibility}

Current models for instruction-guided editing are generally trained on specific instruction sets, often limited by vocabulary and predefined templates. Future research should focus on developing models capable of understanding a broader range of open-domain instructions. Leveraging \textit{self-supervised learning techniques} \cite{goyal2021self} could enhance MLLMs' comprehension by allowing them to interpret and execute complex, domain-specific instructions without the need for extensive labeled data. This approach could enable more flexible models that generalize across different types of instructions, boosting their applicability and effectiveness.

\subsection{Dynamic and Real-time Editing}

Real-time, interactive editing remains challenging due to computational constraints and the difficulty of aligning dynamic user instructions with image modifications. Advancements could include developing lightweight models with \textit{incremental training} and real-time feedback mechanisms~\cite{kim2021exploiting}. Integrating Reinforcement Learning could further allow models to refine edits through iterative, trial-and-error learning, improving flexibility and responsiveness. Additionally, incorporating user feedback loops and personalization, such as adaptive prompts that evolve with user behavior and history, could make instruction-guided editing more intuitive and tailored to individual editing styles.

\subsection{Multi-Granular and Multi-Modal Consistency}

While existing methods handle simple edits, achieving consistency across complex transformations is challenging. Future models could employ \textit{multi-granular hierarchical learning} \cite{min2020hierarchical} to effectively separate global and local adjustments. This approach would allow for more cohesive edits across varying granularities. Additionally, extending \textit{cross-modal learning} to synchronize edits across visual, auditory, and textual data could support creating more integrated multimedia experiences, enabling seamless transitions across different media types.

\section{Conclusions}
\label{sec:conclusion}

The paper examines the transformative impact of large language models (LLMs) and multimodal LLMs on instructional image editing, highlighting how these technologies are expanding the possibilities of visual content manipulation. LLM-powered editing allows users to make intuitive, precise modifications through natural language, lowering technical barriers and enhancing usability across fields like entertainment, fashion, remote sensing, and medical imaging.
Key to these advancements is the use of diverse datasets and robust evaluation metrics, such as SSIM, LPIPS, and CLIP Cosine Similarity, which help ensure high-quality outcomes by assessing both structural fidelity and semantic alignment. As multimodal LLMs, specialized training methods, and curated datasets continue to evolve, instruction-based editing is set to democratize visual content creation, enabling a new era of creative and accessible visual storytelling across industries.

\input{bib.txt}


\end{document}